\documentclass[11pt,a4paper]{article}
\usepackage{geometry}
\usepackage{setspace}
\pdfoutput=1   
\usepackage{amssymb}
\usepackage{hyperref}
\usepackage{xcolor}
\usepackage{xspace}
\usepackage{amsmath,amsfonts}
\usepackage{xspace}
\usepackage{amsmath,amssymb,amsfonts}

\newcommand{\bs}[1]{\ensuremath{\boldsymbol{#1}}}

\newcommand{\bc}{\ensuremath{\bs c}\xspace}

\newcommand{\bl}{\ensuremath{\bs l}\xspace}

\newcommand{\bn}{\ensuremath{\bs n}\xspace}
\newcommand{\bo}{\ensuremath{\bs o}\xspace}
\newcommand{\bp}{\ensuremath{\bs p}\xspace}

\newcommand{\bt}{\ensuremath{\bs t}\xspace}
\newcommand{\bu}{\ensuremath{\bs u}\xspace}
\newcommand{\bv}{\ensuremath{\bs v}\xspace}
\newcommand{\bw}{\ensuremath{\bs w}\xspace}
\newcommand{\bx}{\ensuremath{\bs x}\xspace}
\newcommand{\by}{\ensuremath{\bs y}\xspace}

\newcommand{\bA}{\ensuremath{\bs A}\xspace}
\newcommand{\bB}{\ensuremath{\bs B}\xspace}

\newcommand{\bI}{\ensuremath{\bs I}\xspace}

\newcommand{\bR}{\ensuremath{\bs R}\xspace}

\newcommand{\bX}{\ensuremath{\bs X}\xspace}
\newcommand{\bY}{\ensuremath{\bs Y}\xspace}
\newcommand{\bZ}{\ensuremath{\bs Z}\xspace}

\renewcommand{\Re}{\ensuremath{\mathbb{R}}\xspace} % latex2e

\newcommand{\pc}{\ensuremath{\boldsymbol{p}_c}\xspace}

\newcommand{\norm}[1]{\ensuremath{\| #1\|}\xspace}

\usepackage{algpseudocode}
\usepackage[linesnumbered,ruled,vlined]{algorithm2e}
\SetKwInput{KwInput}{Input}                % Set the Input
\SetKwInput{KwOutput}{Output}              % set the Output
\usepackage{makecell}
\usepackage{lscape}
\usepackage{multirow}
\newcommand{\bpp}{\ensuremath{\bp_{\text{p}}}\xspace}
\newcommand{\bpc}{\ensuremath{\bp_{\text{c}}}\xspace}
\newcommand{\xc}{\ensuremath{x_{\text{c}}}\xspace}
\newcommand{\yc}{\ensuremath{y_{\text{c}}}\xspace}
\newcommand{\zc}{\ensuremath{z_{\text{c}}}\xspace}
\newcommand{\upone}{\ensuremath{u_{1_\text{p}}}\xspace}
\newcommand{\uptwo}{\ensuremath{u_{2_\text{p}}}\xspace}

\newcommand{\surf}{\ensuremath{\mathcal{S}}\xspace}
\newcommand{\plane}{\ensuremath{\mathcal{P}}\xspace}
\newcommand{\nullsp}{\ensuremath{\mathcal{N}}\xspace}
\newcommand{\tpart}[2]{\ensuremath{\frac{\partial #1}{\partial #2 \hfill}}\xspace}

\newcommand{\rmin}{\ensuremath{r_\text{min}}\xspace}
\usepackage[utf8]{inputenc}
\usepackage{framed}
\usepackage{nomencl}
\makenomenclature
\usepackage{enumitem}
\usepackage{subcaption}
\usepackage[affil-it]{authblk}
\usepackage{tikz}
\usetikzlibrary{shapes.geometric, arrows}
\usetikzlibrary{calc}

\tikzstyle{startstop} = [ellipse, minimum width  = 1 cm, minimum height =1 cm, draw = black,  text centered]
\tikzstyle{data} = [trapezium, trapezium left angle=60, trapezium right angle = 120, minimum width  = 0.1 cm, minimum height =0.1 cm, draw = black]
\tikzstyle{decision} = [diamond, minimum width  = 1 cm, minimum height =1 cm, draw = black,text centered]
\tikzstyle{process} = [rectangle, minimum width  = 1 cm, minimum height =1 cm, draw = black,text centered]
\tikzstyle{arrow} = [thick,->,>=stealth]
\tikzstyle{pagecon}=[circle, minimum width  = 1 cm, minimum height =1 cm, draw = black,text centered]
\allowdisplaybreaks

\usepackage[toc,page]{appendix}

\title{A geometrical approach to determine the proximity of a point to an axisymmetric quadric in space}
\author{Bibekananda Patra} 
\author{Aditya Mahesh Kolte} 
\author{Sandipan Bandyopadhyay\thanks{Corresponding author\\ ~~Email addresses: \texttt{bibeka.patra2@gmail.com} (Bibekananda Patra), \texttt{adityakolte72@gmail.com} (Aditya Mahesh Kolte), \texttt{sandipan@iitm.ac.in} (Sandipan Bandyopadhyay)}} 
\affil{Department of Engineering Design \\ Indian Institute of Technology Madras, Chennai 600036, India}
\date{}
\begin{document}

%Department of Engineering Design \\
%Indian Institute of Technology Madras, Chennai, 600036, India \\
%}

\maketitle

%\affiliation{organization={},%Department and Organization
            %addressline={}, 
          %  city={},
           % country={}}

\begin{abstract}
This paper presents the classification of a general quadric into an axisymmetric quadric (AQ) and the solution to the problem of the proximity of a given point to an AQ. The problem of proximity in~$\Re^3$ is reduced to the same in~$\Re^2$, which is not found in the literature. A new method to solve the problem in~$\Re^2$ is used based on the geometrical properties of the conics, such as sub-normal, length of the semi-major axis, eccentricity, slope and radius. Furthermore, the problem in~$\Re^2$ is categorised into two and three more sub-cases for parabola and ellipse/hyperbola, respectively, depending on the location of the point, which is a novel approach as per the authors' knowledge. The proposed method is suitable for implementation in a common programming language, such as~\verb|C| and proved to be faster than a commercial library, namely,~\verb|Bullet|.     
\end{abstract}

%\begin{keyword}



%\end{frontmatter}


\setlength{\nomitemsep}{-\parsep} % Removes the space between items
%% Nomenclature
\nomenclature[1a]{AQ}{Axisymmetric quadric}
\nomenclature[1b]{$\surf$}{A quadratic surface or a quadric} 
\nomenclature[1c]{$\bx$}{$[x,y,z]^\top$, a point in $\Re^3$}
\nomenclature[1d]{$\bI$}{An identity matrix of appropriate dimension}
\nomenclature[1e]{$\lambda$}{Eigenvalue of a matrix}
\nomenclature[1f]{$\bp_0$}{A given point in $\Re^3$ which does not lie on~$\surf$}
\nomenclature[1g]{$\bpc$}{Centre or vertex of an AQ}
\nomenclature[1h]{$\bv_3$}{Axis of symmetry of an AQ}
\nomenclature[1i]{$\nullsp(\cdot)$}{Nullspace of a matrix}
\nomenclature[1j]{$\norm{\cdot}$}{Euclidean norm of any vector}
\nomenclature[1k]{$\eta_i=0$}{Equation of a conic}
\nomenclature[1l]{$\bpp$}{Point corresponding to~$\bp_0$ represented in a plane}
\nomenclature[1m]{$r_\text{min}$}{Shortest distance between~$\bp_0$ and an AQ}

\begin{framed}
	\printnomenclature[3.5 cm]
\end{framed}

\section{Introduction}
\label{sc:intro}
\begin{figure}[t]
	\centering
	\includegraphics[width=0.6\textwidth]{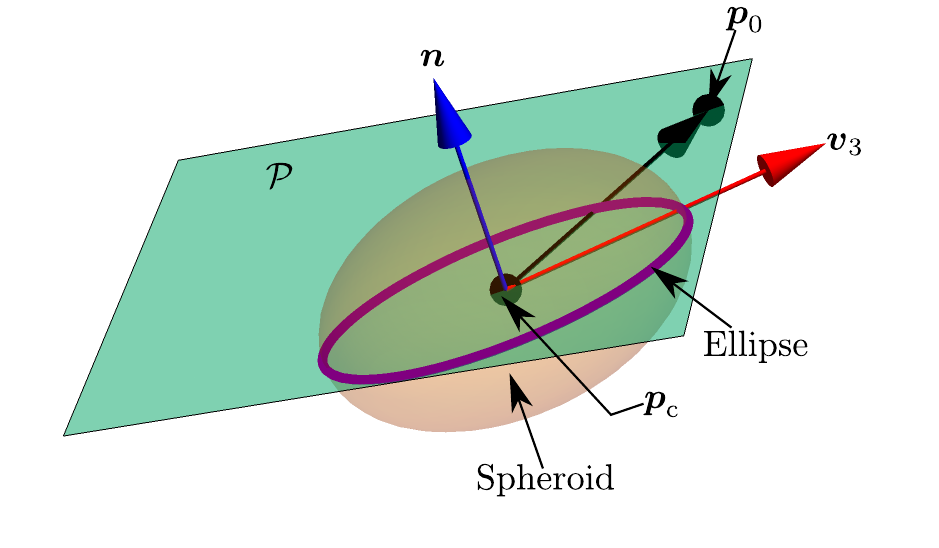}
	\caption{Intersecting plane, which contains the axis~$\bv_3$, and the points~$\bp_0$, and $\bpc$, that intersects the AQ, e.g., a spheroid in this case.}
	\label{fg:intPlaneSphr}
\end{figure}
The shortest distance {\em proximity} of a quadratic surface in~$\Re^3$ (also known as \textit{quadric}) to a given point which does not lie on the given surface has a wide range of applications in engineering and technology, such as robotics, computer graphics, computer vision, computer-aided design, manufacturing, and the global positioning system. In general, quadrics are classified into~$17$ types of surfaces~\cite{Zwillinger2018} (see, pp.~605-606) including imaginary ones. The real surfaces, such as ellipsoids, hyperboloids of one sheet, hyperboloids of two sheets, elliptic paraboloids, elliptic cones, elliptic cylinders, and so on, are discovered to be used in applications for all practical reasons. Many researchers have tried to compute the least distance between a given point and a quadric. The analytical derivation of the proximity of a point to an ellipsoid was derived in~\cite{Lott2014}, and~\cite{Aditya2023}. The most common quadric used in engineering applications is perhaps the ellipsoid, as mentioned in~\cite{Wang2022},~\cite{Bufe2018},~\cite{Song2015},~\cite{Zeng2013}. There are a few research articles which focus on the proximity of a point to an ellipsoid~\cite{Schneider2003} (see, pp.~403-405),~\cite{Uteshev2018}, and~\cite{Bektas2014}. The problem of proximity was solved as the projection of a point onto an ellipsoid in an optimisation framework and a comparison is made among the seven fast numerical algorithms in~\cite{Jia2017}. The shortest distance between a hyperboloid and a point was solved numerically in~\cite{Bektas2017}. The applications related to the proximity of a point and other quadrics are not found in the literature. Moreover, the research on the computation of the proximity of a point to an {\em axisymmetric} quadric (AQ) is not found in the literature according to authors' knowledge. This paper focuses only on the identification of an AQ and the computation of the normal distance from a point to the surface.

The researchers have solved the shortest distance problem either between a point and an ellipsoid or between a point and a hyperboloid. In this paper, the focus is on the axisymmetric surfaces, such as spheroids, hyperboloids, paraboloid, cylinder, cone, and sphere. Using the axis of the symmetry of an AQ, and a given point, the problem of proximity can be reduced from~$\Re^3$ to~$\Re^2$. In other words, a plane can be found which contains the axis of symmetry of an AQ and a given point. The intersection of the plane with the quadric resulting in a conic, as shown in Fig.~\ref{fg:intPlaneSphr}. 
Therefore, the problem of computing the proximity in~$\Re^3$ is equivalent to the problem of normal distance between a point in the plane, denoted by~\plane, corresponding to the point~$\bp_0$ and a conic. Many authors have published articles on the problem of the least distance between a conic and a given point. In most of the literature, the proximity problem of a quadratic curve to a point is used as an orthogonal projection onto the conics for curve fitting applications. In~\cite{Ahn2001}, the conics were fitted by using least-squares orthogonal distances from given points. The computational efficacies of various methods of projecting points onto conics were discussed in~\cite{Chernov2013}. The proximity of a point to a conic was derived in~\cite{Schneider2003} (see, pp.~217-219). The equation of point-to-ellipse distance was analysed in~\cite{Uteshev2018},~\cite{Chou2019}. The distance equations in these literature were derived in closed form. These methods led to a quartic equation for ellipse and hyperbola, while, for parabola, the corresponding equation became a cubic. 

\begin{figure}[!t]
	\centering
	\begin{subfigure}{0.45\textwidth}	
		\centering
		\includegraphics[scale=0.45]{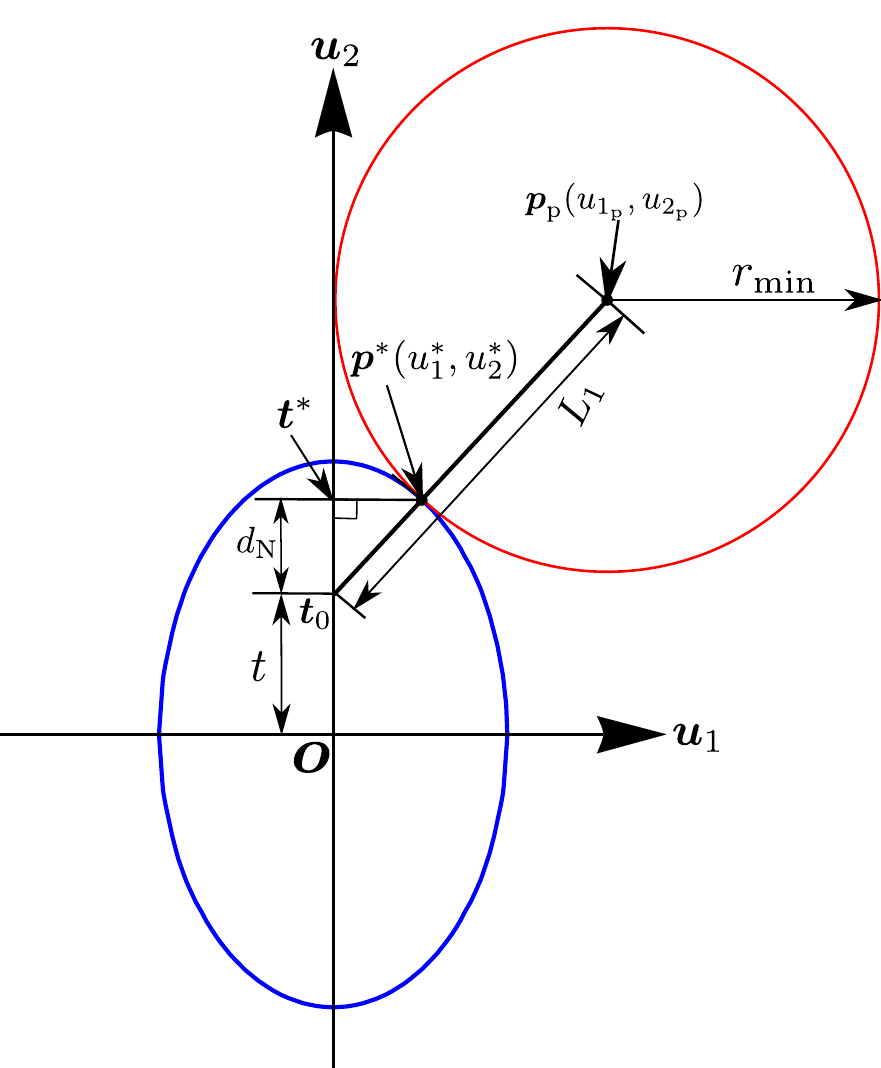}
		\subcaption{An example of the proximity of the point $\bpp$ to an ellipse}
		\label{fg:shortDistanceEllipse}
	\end{subfigure}
	\hspace{1 cm}
	\begin{subfigure}{0.45\textwidth}
		%\vspace{2.0cm}
		\centering
		\includegraphics[scale=0.35]{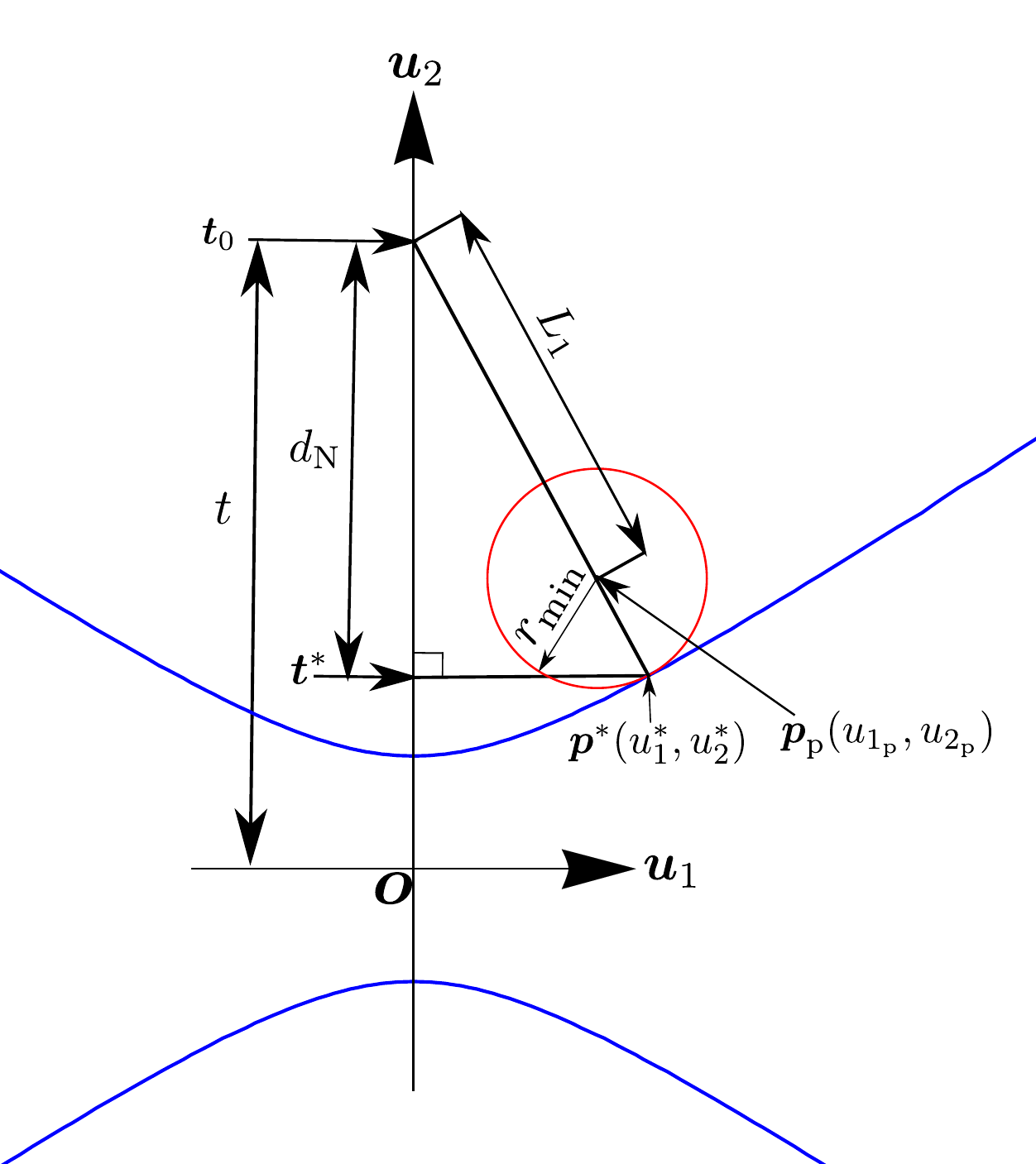}
		\subcaption{An example of the proximity of the point $\bpp$ to a hyperbola}
		\label{fg:shortDistancehyperbola}
	\end{subfigure}
	\caption{Examples of proximity of the point $\bpp$ to an ellipse and a hyperbola in~$\Re^2$}
	\label{fg:shortDistCentralConics}
\end{figure}

\begin{figure}[!t]
	\centering
	\includegraphics[scale = 0.4]{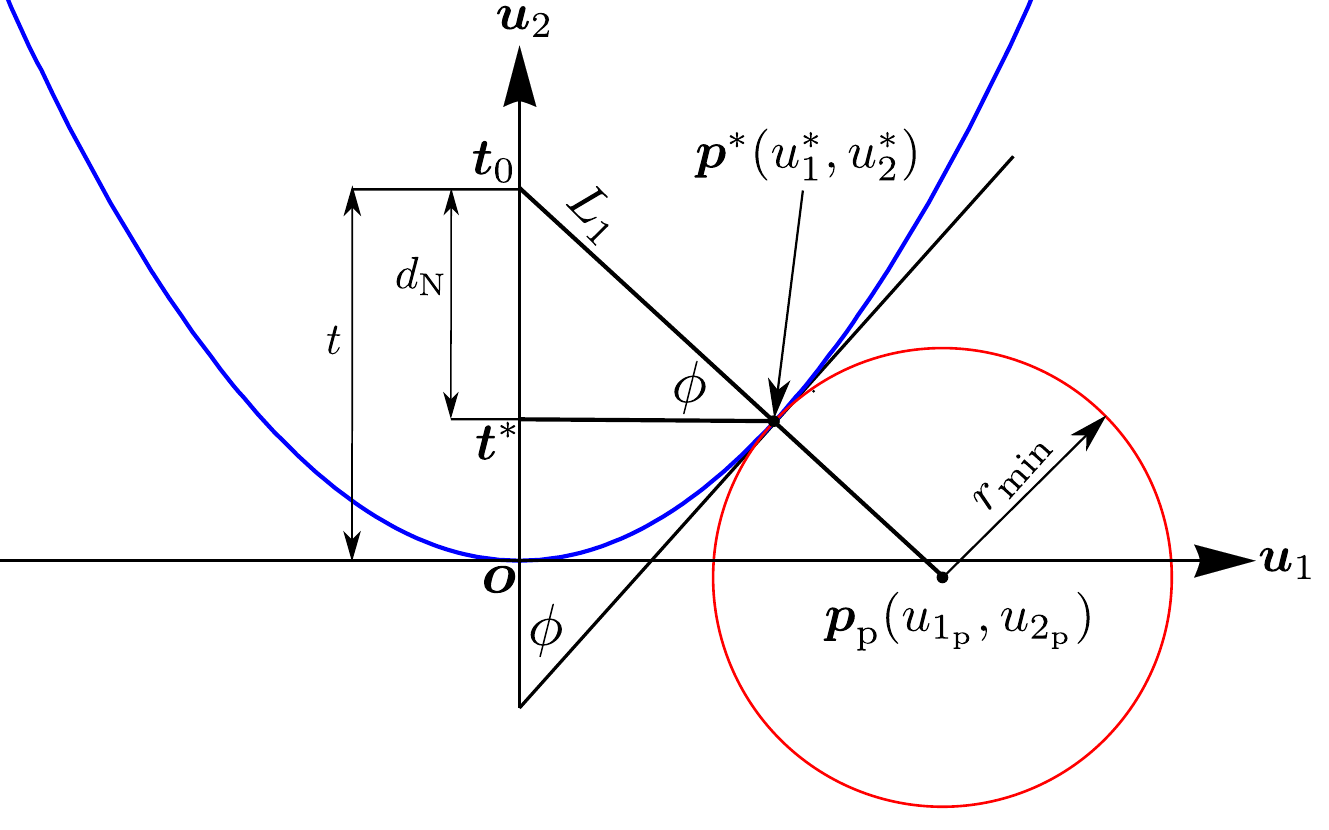}
	\caption{An example of the proximity of the point $\bpp$ to a parabola in~$\Re^2$}
	\label{fg:shortDistParabola}	
\end{figure}

None of the literature uses the basic properties, such as normal, and sub-normal, of these quadratic curves to derive quartic and cubic equations. This method of computing proximity is more appealing in terms of computational efficiencies. In the new approach, the normal passing through a point,~$\bpp$, in $\Re^2$ intersects the major axis of an ellipse or a hyperbola at~$\bt_0$, as shown in Figs.~\ref{fg:shortDistanceEllipse},~\ref{fg:shortDistancehyperbola}. Using the sub-normal of the conics,~$d_\text{N}$, equation of the line~$L_1$ and the equation of conics, a quartic equation in~$t$ is obtained. Similarly, a cubic equation in~$t$ is obtained for the case of a parabola, as shown in Fig.~\ref{fg:shortDistParabola}. The points,~$\bp^*$, corresponding to the real solutions of the quartic and cubic equations are the points where the normal intersects the conics. The minimum Euclidean distance among the distances between the point~$\bpp$, and~$\bp^*$ is the proximity of~$\bpp$ to a conic. Hence, the proposed approach of computing the proximity of a point to a conic mentioned in Sections~\ref{sc:pointToParabola}, and~\ref{sc:pointToEllipHyp} is a novel approach. 
% % % % % % % % % % % % % % % % % % % % % % % % % % % % % % % % % % % %
% % % % % % % % % % % % % % % % % % % % % % % % % % % % % % % % % %
In any approach, the cubic and quartic equation either becomes invalid if the point~$\bpp$ is on the major axis of a parabola, and an ellipse, or a hyperbola. In~\cite{Chernov2013}, the claim was made to achieve ``nearly the same distance" by the authors for the points lie near the centre and on the major axis of an ellipse. Hence, to obtain accurate results for these special cases, these problems are addressed for the first time. The closed-form expressions for the distances for these special cases are mentioned case by case in Sections~\ref{sc:pointToParabola}, and~\ref{sc:pointToEllipHyp}.

The rest of the paper is organised as follows. The procedure for identification of an AQ is described in Section~\ref{sc:MathematicalForm}. Section~\ref{sc:shortDistAAxiSym} includes the identification of the plane that intersects a quadric resulting in a conic, followed by the computation of the proximity of a point to a conic in~$\Re^2$. The numerical examples for all the AQs, along with their computational efficiencies, are described in Section~\ref{sc:examples}. The contributions of the paper are presented in Section~\ref{sc:discussion}. Finally, the paper is concluded in Section~\ref{sc:conclusion}.         
% % % % % % % % % % % % % % % % % % % % %   
\section{Mathematical formulation}
\label{sc:MathematicalForm}
The equation of a general quadric~$\surf$ in $\Re^3$ is written as:
\begin{align}
	&S(\bx):=a x^2 + b y^2 + c z^2 + 2 f x y + 2 g y z + 2 h x z + 2 p x + 2 q y + 2 r z + d = 0, \quad \text{where}\label{eq:generalQuadricEq}\\
	&\bx = [x,y,z]^\top \in \Re^3, \nonumber \\
	&a, b, c, d, f, g, h, p, q, r \in \Re. \nonumber	
\end{align}  
Equation~(\ref{eq:generalQuadricEq}) can be expressed via a {\emph{quadratic form}}:
\begin{equation}
	\by^\top\bA \by = 0, \quad \text{where}~\bA = \begin{bmatrix}
	\bB & \bc \\
	\bc^\top & d
	\end{bmatrix},\bB = \begin{bmatrix}
	a & f & h \\
	f & b & g \\
	h & g & c \\
	\end{bmatrix},~\text{and}~
	\bc = [p,q,r]^\top.
	\label{eq:quadraticForm1}
\end{equation}  
The vector~$\by\in \Re^3$ is defined as~$\by = [\bx^\top,1]^\top$. Based on the values of the coefficients of the LHS of Eq.~(\ref{eq:generalQuadricEq}), the set of quadrics is divided into AQs and non-axisymmetric quadrics. The classification of a general quadric is adopted from~\cite{Zwillinger2018} (see, pp.~219-221) and modified suitably. This paper focuses on the identification of the AQs based on the coefficients of $S(\bx)$, i.e.,~$a,d$, invariants of the characteristic polynomial of~$\bB$, determinant of~$\bA$, and the linear coefficient of the {\em depressed cubic} of the characteristics polynomial of~$\bB$.

The characteristic equation of~$\bB$ is obtained as\footnote{In this article, $\bI$ denotes an {\em identity matrix}, whose dimension is contextual.}:
\begin{equation}
\text{det}(\lambda \bI-\bB) = \lambda^3 - J_1 \lambda^2 + J_2 \lambda - J_3 = 0,
\label{eq:charPolyB}
\end{equation} 
where~$J_j, j=1,2,3$, are the {\em invariants}\footnote{For details, see,~\cite{Procesi2017},~pp.~22-23.} of~$\bB$ obtained as:
\begin{align}
J_1 = &~a + b + c,\nonumber\\
J_2 = &~ab + bc + ca - (f^2+g^2+h^2),\nonumber \\
J_3 = &~a (b c - g^2) - c f^2 + h (2 f g - b h).
\label{eq:invCharB}
\end{align}
The determinant of~$\bA$ can be obtained by performing {\em Gaussian elimination} on the matrix and multiplying diagonals of the lower triangular matrix. The expression for the determinant is as follows:
\begin{align}
	&\det(\bA) = \frac{a (\alpha \beta (p \phi -h \psi)^2 -r^2 - d (h^2 \alpha + \beta \phi^2)+ 2 r (h p \alpha + \beta \phi \psi) + c(d - p^2 \alpha - \beta \psi^2))}{\beta}, \quad \text{where} \\
	&\alpha =1/a, \nonumber\\
	&\phi = g - f h \alpha, \nonumber\\
	&\psi = q - f p \alpha,~\text{and} \nonumber \\
	&\beta = 1/(b - f^2 \alpha). \nonumber
\end{align}

The eigenvalues of~$\bB$ are obtained by solving Eq.~(\ref{eq:charPolyB}) for the variable~$\lambda$. Since~$\bB$ is a real symmetric matrix and~$\surf$ is axisymmetric, at least two roots of Eq.~(\ref{eq:charPolyB}) are the same~\cite{Paul2009}. For this condition, the cubic equation may be solved in the {\em closed form}~\cite{Nickalls2006},~\cite{Nickalls1993}, as described below. 

First, the cubic equation is reduced to its depressed form:
\begin{align}
&\kappa^3 + a_0 \kappa + a_1 = 0, \quad \text{where} \label{eq:depCubic}\\
&\kappa  = \lambda - \frac{J_1}{3}. \nonumber
\end{align}
The coefficients of the depressed cubic are obtained as:
\begin{align}
a_0 = &\frac{1}{3}(ab + bc + ca - (a^2 + b^2 + c^2)) - (f^2+g^2+h^2),\nonumber \\ 
a_1 = &\frac{1}{27}(a(a(3(b + c) - 2 a) + 3 ((b-c)^2 - 2 b c - 3 (f^2 + h^2 -2 g^2))) \nonumber\\
&+ b (3(c(b + c)  -3 (f^2 + g^2 - 2 h^2))-2 b^2) - c (2 c^2 + 9 (g^2 + h^2 - 2 f^2))) - 2 f g h.\nonumber
\end{align}
The condition for at least a pair of roots (say, $\kappa_1=\kappa_2$) to be equal is given by the vanishing of the discriminant~$\Delta$, where,
\begin{equation}
\Delta = 4 a_0^3 + 27 a_1^2.
\label{eq:discriCubicEq}
\end{equation}
The nature of the remaining root, say,~$\kappa_3$, depends on the value of the coefficient of the linear term of Eq.~(\ref{eq:depCubic}), i.e.,~$a_0$. If~$a_0$ vanishes, then the root~$\kappa_3$ equals the other two roots, and it is distinct otherwise. The eigenvalues of~$\bB$ corresponding to the roots of ~Eq.~(\ref{eq:depCubic}) are~\mbox{$\lambda_j = \kappa_j + \frac{J_1}{3},j=1,2,3$}, as shown below:
\begin{subequations}
	\begin{align}
	&\lambda_1 = \lambda_2 = -\left(\frac{J_1}{3} + \frac{3 a_1}{2 a_0}\right), \quad \lambda_1=\lambda_2\neq 0,\\ 
	&\lambda_3 = \begin{cases}
	\lambda_1, & \text{if}~a_0=0,\\
	\frac{3a_1}{a_0} - \frac{J_1}{3}  ,& \text{otherwise}.
	\end{cases}
	\end{align}
	\label{eq:eigenValueB}
\end{subequations}

The discriminant $\Delta$, invariant of~$\bB$, i.e,~$J_3$,~$\det(\bA)$, and the coefficients~$a,d,a_0$ are used for identification of an AQ, which is discussed in Section~\ref{sc:charAxisymSurface}.
\subsection{Identification of an axisymmetric quadric from a generic quadric}
\label{sc:charAxisymSurface}
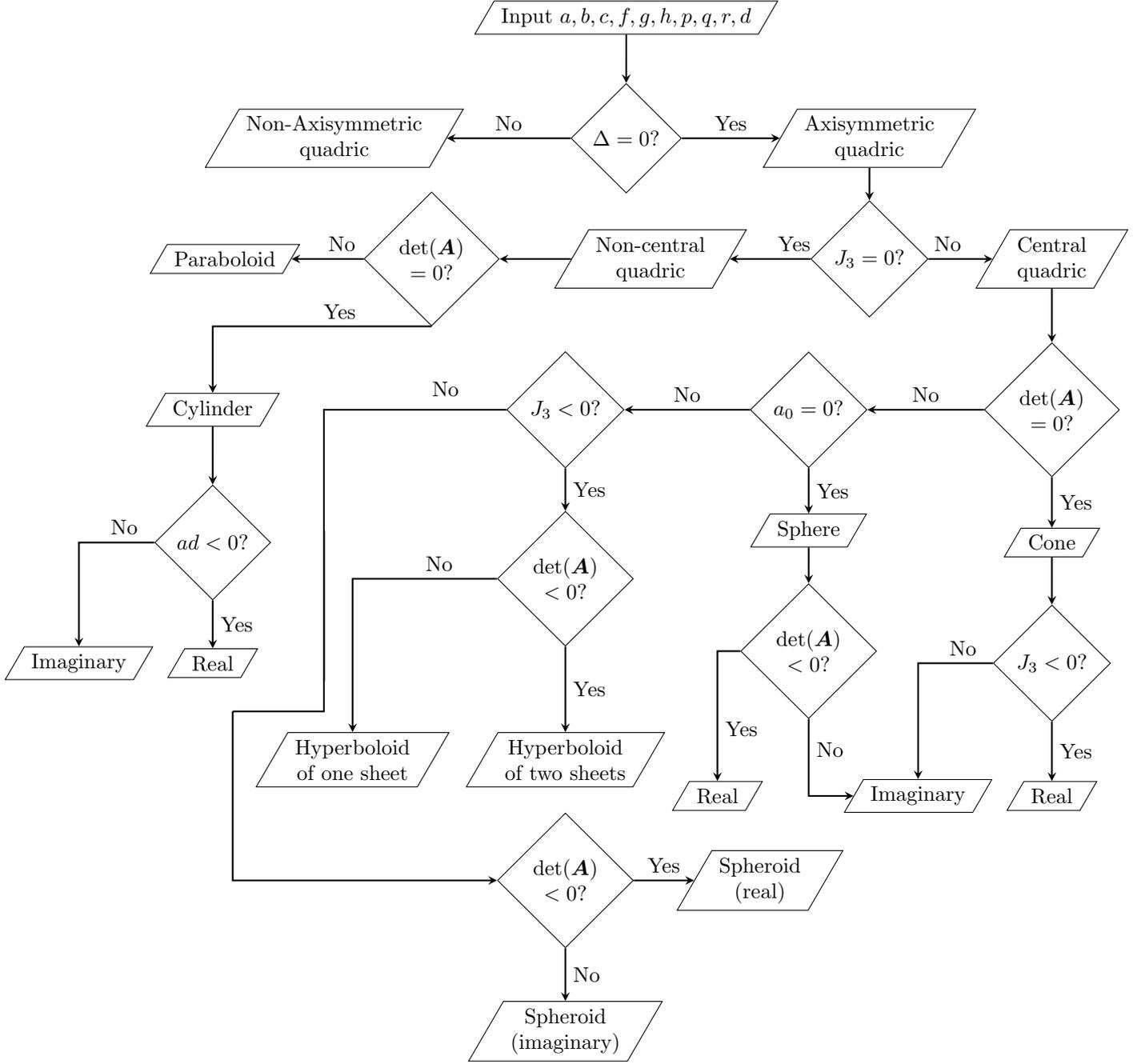
\begin{figure}[!t]
	\centering
	\small
	\begin{tikzpicture}[node distance = 2 cm]
		\node(coefficient)[data]{Input $a,b,c,f,g,h,p,q,r,d$};
		\node(decisionSymmetry)[decision,below of = coefficient, yshift = 0 cm ,align = center]{$\Delta=0$?};
		\draw[arrow](coefficient)--(decisionSymmetry);
		\node(nonaxisymmetry)[data,left of = decisionSymmetry, align=center,xshift = -2.8 cm ]{Non-Axisymmetric \\ quadric};
		\draw[arrow](decisionSymmetry)--node[anchor = south,align=right]{No}(nonaxisymmetry);
		\node(axisymmetry)[data,right of = decisionSymmetry, align=center,xshift = 2 cm ]{Axisymmetric \\ quadric};
		\draw[arrow](decisionSymmetry)--node[anchor = south,align=right]{Yes}(axisymmetry);
		\node(J3zeroDecision)[decision,below of = axisymmetry, yshift = 0 cm]{$J_3=0$?};
		\draw[arrow](axisymmetry)--(J3zeroDecision);
		\node(noncentralQuadric)[data,left of=J3zeroDecision,xshift = -1.4 cm,align =center, xshift = -0.2 cm]{Non-central \\ quadric};
		
		\node(centralQuadric)[data,right of=J3zeroDecision,xshift = 1.0 cm,align =center]{Central \\ quadric};
		\draw[arrow](J3zeroDecision)--node[anchor=south,pos=0.2]{Yes}(noncentralQuadric);
		\draw[arrow](J3zeroDecision)--node[anchor=south,pos=0.3]{No}(centralQuadric);
		\node(I4ZeroDecision)[decision,left of=noncentralQuadric,xshift=-1.6 cm, align = center]{$\det(\bA)$ \\$=0$?};
		\draw[arrow](noncentralQuadric)--(I4ZeroDecision);
		\node(parabola)[data, left of = I4ZeroDecision, xshift = -1.4 cm]{Paraboloid};
		\draw[arrow](I4ZeroDecision)--node[anchor=south,pos=0.3]{No}(parabola);
		\node(cylinder)[data, below of = parabola, yshift = -0.5 cm, xshift = -0.2 cm, align =center]{Cylinder};
		\draw[arrow](I4ZeroDecision.south)-|node[anchor=south,pos=0.21]{Yes}(cylinder.north);
		\node(realCylinCond)[decision, below of = cylinder,yshift = -0.2 cm]{$ad<0$?};
		\draw[arrow](cylinder)--(realCylinCond);
		\node(realCylinder)[data, below of =realCylinCond]{Real};
		\draw[arrow](realCylinCond)--node[anchor=west]{Yes}(realCylinder);
		\node(imagCylinder)[data, left of =realCylinder, xshift = -0.2 cm]{Imaginary};
		\draw[arrow](realCylinCond)-|node[anchor=south,pos=0.2]{No}(imagCylinder);
		\node(I4ZeroDecision2)[decision, below of=centralQuadric,yshift=-0.5 cm, align = center]{$\det(\bA)$ \\$=0$?};
		\draw[arrow](centralQuadric)--(I4ZeroDecision2);
		\node(cone)[data,below of=I4ZeroDecision2,yshift=-0.2 cm,align =center]{Cone};
		\draw[arrow](I4ZeroDecision2)--node[anchor=west]{Yes}(cone);
		\node(conerealcond)[decision,below of=cone,align = center]{$J_3<0?$};
		\draw[arrow](cone)--(conerealcond);
		\node(conereal)[data, below of=conerealcond,yshift = -0.2 cm]{Real};
		\draw[arrow](conerealcond)--node[anchor=west]{Yes}(conereal);
		\node(coneimag)[data, left of=conereal, xshift = -0.2 cm]{Imaginary};
		\draw[arrow](conerealcond)-|node[anchor=south,pos=0.2]{No}(coneimag);
		\node(a1ZeroDecision)[decision, left of = I4ZeroDecision2,xshift=-2 cm]{$a_0=0?$};
		\draw[arrow](I4ZeroDecision2)--node[anchor=south]{No}(a1ZeroDecision);
		\node(sphere)[data, below of = a1ZeroDecision]{Sphere};
		\draw[arrow](a1ZeroDecision)--node[anchor=west]{Yes}(sphere);
		\node(I4negative)[decision, below of = sphere,yshift=0 cm, align = center]{$\det(\bA)$ \\$<0$?};
		\draw[arrow](sphere)--(I4negative);
		\draw[arrow](I4negative)|-node[anchor=west,pos=0.2]{No}(coneimag);
		\node(sphrreal)[data, left of=coneimag, xshift=-1.3 cm]{Real};
		\draw[arrow](I4negative.west)-|node[anchor=west,pos=0.8]{Yes}(sphrreal.north);
		\node(sphrdrealdecision)[decision, left of = a1ZeroDecision,xshift=-2 cm]{$J_3<0?$};
		\draw[arrow](a1ZeroDecision)--node[anchor=south]{No}(sphrdrealdecision);
		\node(I4negative2)[decision, below of=sphrdrealdecision,yshift=-0.8cm, align = center]{$\det(\bA)$ \\$<0$?};
		\draw[arrow](sphrdrealdecision)--node[anchor=west]{Yes}(I4negative2);
		\node(hyptwosheets)[data, below of=I4negative2,align=center,yshift=-1 cm]{Hyperboloid \\ of two sheets};
		\draw[arrow](I4negative2)--node[anchor=west]{Yes}(hyptwosheets);
		\node(hyponesheet)[data, left of=hyptwosheets,align=center,xshift=-1.5 cm]{Hyperboloid \\of one sheet};
		\draw[arrow](I4negative2)-|node[anchor=south,pos=0.2]{No}(hyponesheet);
		\node(I4negative2)[decision, below of=hyptwosheets, align =center]{$\det(\bA)$ \\$<0$?};
		\draw[arrow](sphrdrealdecision.west)-|+(-3,-4.5)--+(-3,-2)|-+(-4.5,-5)node[anchor=south,pos=-0.35, xshift = 2 cm]{No}+(-4.5,-5)|-(I4negative2);
		\node(sprdreal)[data, right of=I4negative2, align=center, xshift =1.2 cm]{Spheroid\\ (real)};
		\node(sprdimag)[data, below of=I4negative2, align=center, yshift =-0.5 cm]{Spheroid\\ (imaginary)};
		\draw[arrow](I4negative2)--node[anchor=south]{Yes}(sprdreal);
		\draw[arrow](I4negative2)--node[anchor=west]{No}(sprdimag);
	\end{tikzpicture}
	\caption{Classification of a quadric into the AQs}
	\label{fg:flowCharSurface}
\end{figure}
A given quadric may be classified as an axisymmetric and a non-axisymmetric based on~$\Delta$. If~$\Delta$ vanishes, the quadric is identified as an axisymmetric otherwise non-axisymmetric, as shown in the flowchart in Fig.~\ref{fg:flowCharSurface}. An AQ is further classified into a {\em central} and {\em non-central quadric} depending upon the value of~$J_3$. Further, a non-central quadric is categorised into either a paraboloid or a cylinder. If the matrix~$\bA$ is a non-singular matrix, the surface is identified as a paraboloid otherwise a cylinder. Based on the signs of the coefficients~$a$, and~$d$, the cylinder is categorised as either real or imaginary.

The central quadric has similar classifications based on the values of~$J_3$,~$\det(\bA)$, and~$a_0$. If the matrix~$\bA$ is singular, the central quadric is identified as a cone and the real nature of the cone is dependent on the sign of~$J_3$. If a central quadric is not a cone, the surface is further characterised into either spherical or non-spherical surfaces. If~$a_0$ of Eq.~(\ref{eq:depCubic}) is zero, the quadric is identified as a spherical surface and further the surface is categorised into either real or imaginary based on the value of~$\det({\bA})$. The non-spherical surface is characterised into hyperboloids and spheroid based on the value of~$J_3$. If both~$J_3$ and~$\det(\bA)$ are negative, the AQ is known as hyperboloid of two sheets. If~$J_3$ is negative but~$\det(\bA)$ is positive, the surface is defined as hyperboloid of one sheet. Similarly, positive value of~$J_3$ and the negative value of~$\det(\bA)$ detect the quadric as a real spheroid. If values of both~$J_3$ and~$\det(\bA)$ are positive, the quadric is known as imaginary spheroid. Once identification of an AQ is done, the next task is to compute the proximity of a point to the said AQ, which is explained in the following section.
% % % % % % % % % % % % % % % % % % % % % % % % % % % % % % %
\section{Computation of the proximity of a given point to an axisymmetric quadric}
\label{sc:shortDistAAxiSym}
There are seven types of AQs, namely, (a) spheroid, (b) hyperboloid of one sheet, (c) hyperboloid of two sheets, (d) cone, (e) paraboloid, (f) cylinder, and (g) sphere. The motivation for computing the proximity of a given point to an AQ is that the problem can be reduced from~$\Re^3$ to~$\Re^2$ and solved using the geometric properties of the planar quadratic curves, i.e., conics. The reason behind doing so is that the solution process is more geometrically intuitive and less computationally expensive than the problem of computing the proximity of a point to an AQ in space. To solve the problem in~$\Re^2$, a plane, described by~\mbox{$\plane$}, that intersects a given AQ and passes through the given point, must be identified. The intersection of the plane with AQs results in either ellipse, hyperbola, parabola, pair of non-coincident vertical parallel lines, pair of intersecting lines, or circle. The determination of~\plane is discussed in Section~\ref{sc:intPlane}.   
% % % % % % % % % % % % % % % % % % % % % % % %
%\begin{figure}[!t]
%	\centering
%	\begin{subfigure}{0.3\textwidth}
%		\includegraphics[scale=0.4]{ellipsoid_edited.pdf}
%		\subcaption{Spheroid}
%	\end{subfigure}
%	\hspace{0.9 cm}
%	\begin{subfigure}{0.3\textwidth}
%		\vspace{-0.5 cm}
%		\includegraphics[scale=0.25]{elliptic_cone_edited.pdf}
%		\subcaption{Cone}
%	\end{subfigure}
%	\hspace{0.1 cm}
%	\begin{subfigure}{0.3\textwidth}
%		\vspace{0.3 cm}
%		\includegraphics[scale=0.25]{elliptic_paraboloid_edited.pdf}
%		\subcaption{Paraboloid}
%	\end{subfigure}
%	\begin{subfigure}{0.3\textwidth}
%		\vspace{1.8 cm}
%		\includegraphics[scale=0.25]{hyperboloid_one_sheet_edited.pdf}
%		\subcaption{Hyperboloid of one sheet}
%	\end{subfigure}
%	\hspace{1.3 cm}
%	\begin{subfigure}{0.3\textwidth}
%		\vspace{0.8 cm}
%		\includegraphics[scale=0.15]{hyperboloid_two_sheets_edited.pdf}
%		\subcaption{Hyperboloid of two sheets}
%	\end{subfigure}
%	\hspace{-0.1 cm}
%	\begin{subfigure}{0.3\textwidth}
%		\vspace{1.3 cm}
%		\includegraphics[scale=0.25]{sphere.pdf}
%		\subcaption{Sphere}
%	\end{subfigure}
%	\caption{Non-extruded axisymmetric quadrics centred at $[0,0,0]^\top$ with $\bZ$-axis as the axis of the symmetry.\bnp{Figures are not required}}
%	\label{fg:axisymmSurf}
%\end{figure}
% % % % % % % % % % % % % % % % % % % % % % % %
\subsection{Determination of intersecting plane~\plane}
\label{sc:intPlane}
The plane~\plane is determined by the axis of symmetry of the given AQ,~$\bv_3$, as shown in Fig.~\ref{fg:intPlaneSphr}, and the point,~$\bp_0$, from which the proximity to the quadric is computed, should be known. The axis~$\bv_3$ is the eigenvector of~$\bB$ corresponding to the eigenvalue~$\lambda_3$~\cite{Paul2009}. To obtain~$\bv_3$, the following system of three linearly dependent equations may be solved.	
\begin{align}
	&\bB_1\bv_3 = \boldsymbol{0},\quad \text{where} \label{eq:eqnforEigvec} \\
	&\bB_1 = \lambda_3\bI-\bB,~\text{and}~\bv_3=[x_3,y_3,z_3]^\top. \nonumber
\end{align}     
For the non-trivial eigenvector~$\bv_3$, the determinant of~$\bB_1$ vanishes and the vector lies in the {\em nullspace} of~$\bB_1$, denoted by~$\nullsp(\bB_1)$. The computation of axis~$\bv_3$ is demonstrated in Appendix~\ref{sc:nullspace}.

\begin{figure}[!t]
	\centering
	\begin{minipage}{0.45\textwidth}
		\centering
		\includegraphics[scale=0.4]{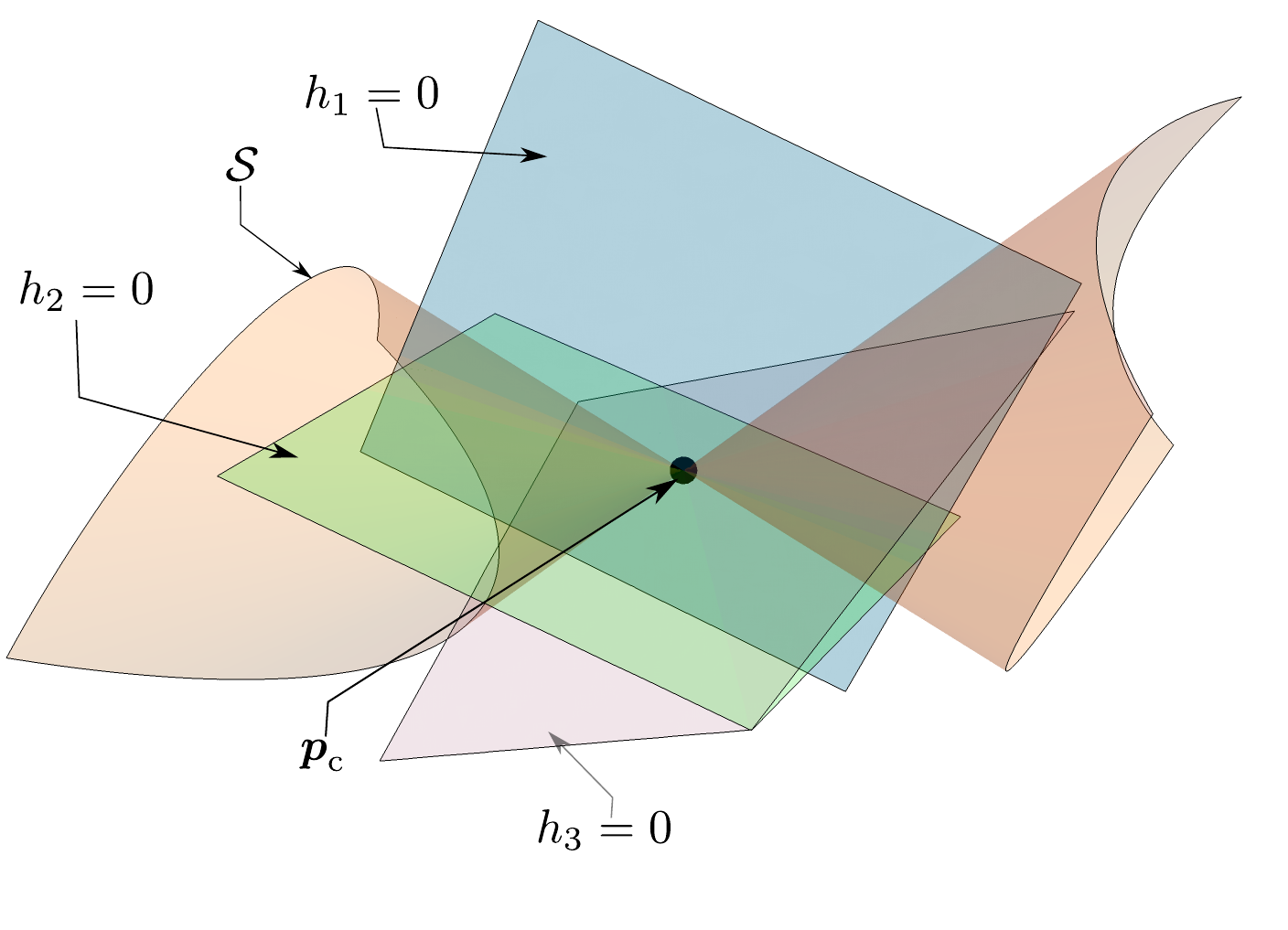}
		\caption{Three planes resulting from Eq.~(\ref{eq:gradients}) intersecting at the centre,~$\bpc$, of a central quadric}
		\label{fg:intPlanesCentSurf}
	\end{minipage}
	\hfill
	\begin{minipage}{0.45\textwidth}
		\centering
		\includegraphics[scale=0.35]{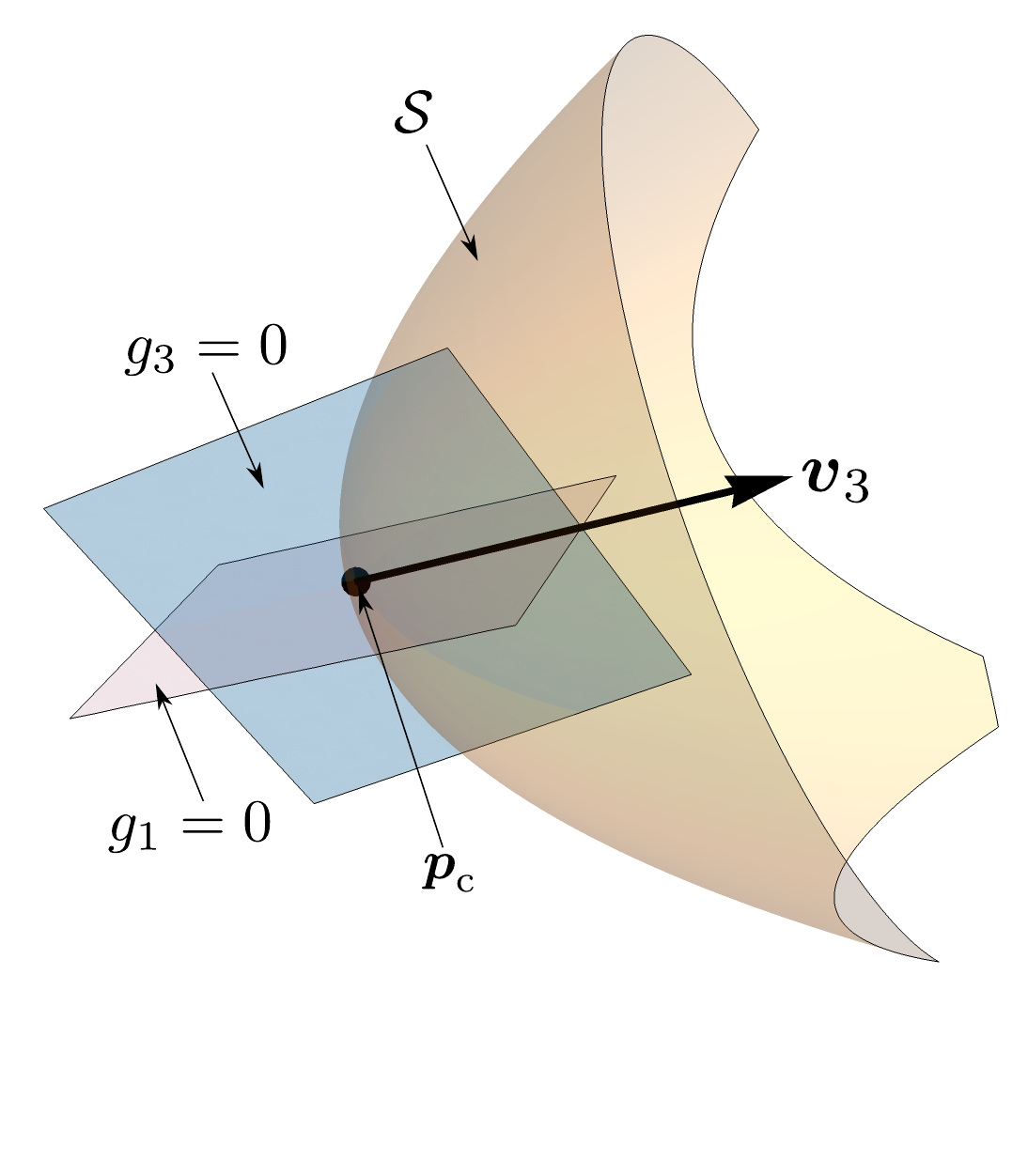}
		\caption{Two planes described by Eq.~(\ref{eq:gradients}) and a paraboloid intersecting at the vertex~$\bpc$}
		\label{fg:intPlanesNonCentSurf}
	\end{minipage}
	
\end{figure}

To obtain~$\plane$, that includes~$\bv_3$ and~$\bp_0$ which is not on the surface, a point on~$\bv_3$, needs to be obtained. The centre or vertex of~$\surf$ can be one of the such points. For the central quadrics, the matrix~$\bB$ is non-singular and it is singular for the non-central quadrics, as shown in flowchart in Fig.~\ref{fg:flowCharSurface}. The centre of the central quadrics,~$\bpc = [\xc,\yc,\zc]^\top$, is obtained by knowing the fact that at~$\bpc$, the gradient of the surface vanishes:
\begin{equation}
	\nabla S(\bpc) := \tpart{S(\bpc)}{\bx} = \boldsymbol{0}.	
	\label{eq:gradients}
\end{equation} 
%\begin{figure}[!t]
%	\centering
%	\includegraphics[scale=0.5]{hyperboloid_plane_intersection.pdf}
%	\caption{Three planes coming from Eq.~(\ref{eq:gradients}) intersecting at the centre,~$\bpc$, of a central quadric}
%	\label{fg:intPlanesCentSurf}
%\end{figure} 
Equation~(\ref{eq:gradients}) consists of the three linearly independent equations,~\mbox{$f_i(\bpc) = 0,i=1,2,3$}. Geometrically, they describe three planes passing through~$\bpc$, as shown in Fig.~\ref{fg:intPlanesCentSurf}. The expressions for~$\{\xc,\yc,\zc\}$ after solving the three equations are as follows:
\begin{subequations}
	\begin{align}
	\xc &= \frac{p (g^2 - b c) + q (c f -g h) + r (b h - f g)}{J_3},\\ 
	\yc &= \frac{p (c f  - g h) + q (h^2 - c a) + r (a g - h f)}{J_3},\\
	\zc &= \frac{p (b h -f g) + q (a g - h f) + r (f^2 - a b)}{J_3}, \quad \text{where}~J_3 \neq 0. 
	\end{align}
	\label{eq:cenQuadSurface}
\end{subequations}

When~$\surf$ is a paraboloid, the gradient of~$\surf$ at~$\bpc$ aligns with the axis of symmetry,~$\bv_3$, hence:
\begin{equation}
	\nabla S(\bpc) \times \bv_3 = \boldsymbol{0}.
	\label{eq:alignCondn}
\end{equation} 
Three scalar equations in Eq.~(\ref{eq:alignCondn}), say,~$g_j = 0,j=1,2,3$, are linear in~$\{\xc,\yc,\zc\}$, and any two of the three equations are linearly independent. Such a pair of linear equations and~$S(\bpc) = 0$, form three independent equations in~$\{\xc,\yc,\zc\}$. Hence, the vertex is found from the intersection of two planes and a paraboloid, as shown in Fig.~\ref{fg:intPlanesNonCentSurf}. Solving any two linear equations, say~$g_1 = 0$, and~$g_3 = 0$, results in the solutions of~$\{\yc,\zc\}$ in terms of~$\xc$ as follows:
\begin{subequations}
	\begin{align}
	&\yc = \frac{x_3 (g r + \xc(g h - cf)- c q ) + y_3 (c p + \xc (c a - h^2)- h r) + z_3(h q + \xc (h f -a g)- g p)}{\Delta_1}, \label{eq:linSolyc}\\
	&\zc = \frac{x_3 (g q + \xc(f g - b h)- b r ) + y_3 (f r + \xc (h f - a g) - g p) + z_3(b p + \xc (a b - f^2) - f q}{\Delta_1}, \label{eq:linSolzc} \\
	&\text{where}~\Delta_1=x_3 (b c - g^2) + y_3 (g h - c f) + z_3 (f g - b h). \nonumber
	\end{align}
	\label{eq:linSolPc}
\end{subequations} 
Substituting the expressions of~$\{\yc,\zc\}$ from Eqs.~(\ref{eq:linSolyc},~\ref{eq:linSolzc}) in~$S(\bpc) = 0$, results in a quadratic equation in~\xc. The leading term of the quadratic equation is always zero because it contains $J_3$ as a factor. Hence, the linear equation is defined as follows:
\begin{equation}
	a_3 \xc + a_4 = 0,
	\label{eq:linearXc}
\end{equation}
and the solution of Eq.~(\ref{eq:linearXc}) is obtained as:
\begin{equation}
	\xc = -\frac{a_4}{a_3}.
	\label{eq:solXc}
\end{equation}
Substitution of~$\xc$ from Eq.~(\ref{eq:solXc}) in Eq.~(\ref{eq:linSolPc}) yields in the vertex~\bpc of the paraboloid. However, Eq.~(\ref{eq:linSolPc}) is invalid if the axis~$\bv_3$ aligns with either of the principal axes of the global frame of reference. For example, if~$\bv_3$ is along the~$\bZ$-axis, the values of~$c,f,g$, and $h$ are zero resulting the denominator of~\yc and~\zc in Eqs.~(\ref{eq:linSolyc},~\ref{eq:linSolzc}) to zero. By setting~$c,f,g$, and $h$ to zero, the equation of paraboloid is obtained as:
\begin{equation}
	S_\text{p}(\bpc) := b (\xc^2 + \yc^2) + 2 p \xc + 2 q \yc + 2 r \zc + d = 0.
	\label{eq:parabZalignEq}
\end{equation}
By replacing~$S_\text{p}(\bpc)$ in place of~$S(\bpc)$, and substituting~$\bv_3 = [0,0,1]^\top$, Eq.~(\ref{eq:alignCondn}) reduces to the following two linear equations in~$\{x,y\}$:
\begin{align}
	&b \xc + p = 0, \label{eq:parbZalignLinEq1}\\
	&b \yc + q = 0. \label{eq:parbZalignLinEq2}
\end{align}
Substituting the solutions of Eqs.~(\ref{eq:parbZalignLinEq1},~\ref{eq:parbZalignLinEq2}) in Eq.~(\ref{eq:parabZalignEq}), the vertex of the paraboloid is found as:
\begin{equation}
	\bpc = -\frac{1}{b}\left[p,q,-\frac{p^2 + q^2- b d}{2 r}\right]^\top.
	\label{centParbZAligned}
\end{equation} 
In Eq.~(\ref{centParbZAligned}),~$b$ and~$r$ can not be zero for a valid paraboloid.
Similarly, the expressions for the vertex of a paraboloid for which~$\bv_3$ is aligned along the~$\bY$-axis is mentioned as:
\begin{equation}
\bpc = -\frac{1}{c}\left[p,-\frac{p^2 + r^2-cd}{2q},r\right]^\top.\label{eq:vertParabYaligned}
\end{equation}
The vertex of a paraboloid, which axis~$\bv_3$ is aligned along the~$\bX$-axis, is derived as follows:
\begin{equation}
	\bpc = -\frac{1}{b}\left[-\frac{q^2 + r^2-bd}{2p},q,r\right]^\top.\label{eq:vertParabXaligned}
\end{equation}
\begin{figure}[!t]
	\centering
		\includegraphics[scale=0.3]{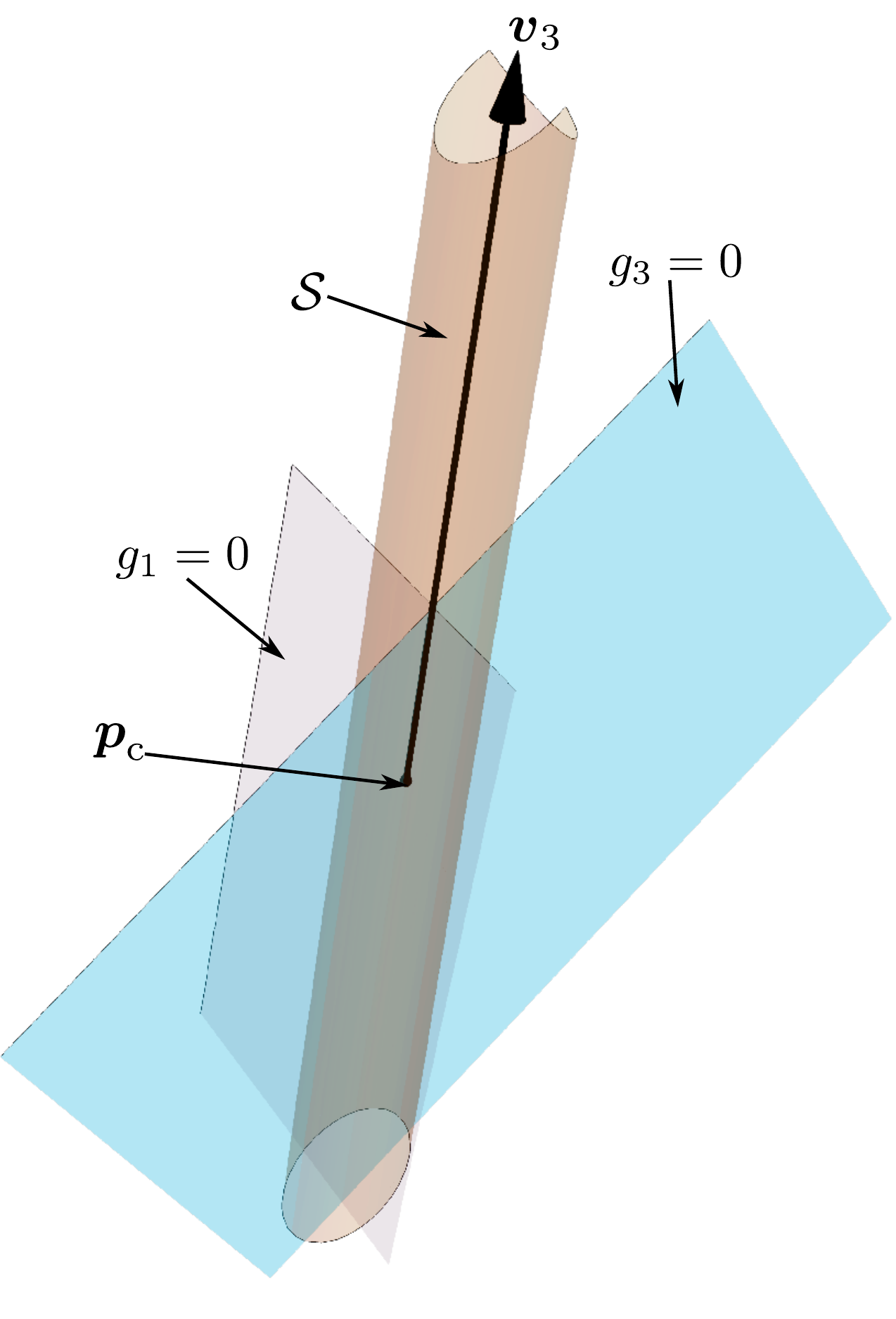}
		\caption{Vector~$\bv_3$ along the intersection of two planes and a point~$\bpc$ on the axis}
\end{figure}

For the case of cylinder, a point on the axis of symmetry~$\bv_3$ is obtained from two intersecting planes obtained from Eq.~(\ref{eq:alignCondn}). To solve two linear equations~$g_1=0$, and $g_3=0$ in two variables, one of the variables in~$\{\xc,\yc,\zc\}$ is chosen arbitrarily, e.g., in this case,~$\xc =1$, hence, Eqs.~(\ref{eq:linSolyc},~\ref{eq:linSolzc}) become:

\begin{subequations}
	\begin{align}
		\yc = \frac{x_3 (g (r + h) - c (f + q) ) + y_3 (c (p + a) - h (h + r)) + z_3(h (q + f) - g (a + p))}{\Delta_1}, \label{eq:linSolycCylinder}\\
		\zc = \frac{x_3 (g (q + f)  - b (h + r) ) + y_3 (f (r + h) -g (a + p)) + z_3(b (p + a) - f (f + q))}{\Delta_1}. \label{eq:linSolzcCylinder}
	\end{align}
	\label{eq:linSolPcCylinder}
\end{subequations} 
Therefore, in this case, a point on~$\bv_3$ is obtained as $\bp_c = [1, \yc , \zc ]\top$. However, Eqs.~(\ref{eq:linSolycCylinder},~\ref{eq:linSolzcCylinder}) are invalid if the axis $\bv_3$ aligns with either of the principal axes of the global frame of reference. Hence, the computation of~$\bpc$ is the same as demonstrated for the case of paraboloid. Therefore, the point~$\bpc$ for the cylinder for which the axis of symmetry is along~$\bX$,~$\bY$, and~$\bZ$-axis is obtained as~$[1,-q/b,-r/b]^\top$,~$[-p/c,1,-r/c]^\top$, and~$[-p/b,-q/b,1]^\top$, respectively.

The plane~\plane is found using the centre or vertex of the surface~$\bpc$, a given point~$\bp_0$, and axis of symmetry~$\bv_3$, as shown in Fig.~\ref{fg:intPlaneSphr}. 
The vector normal to~\plane, denoted by~$\bn$, is obtained as: 
\begin{align}
&\bn = \bv_3 \times \bl, \quad \text{where} 
\label{eq:planeNormal} \\
&\bl=\bp_0-\bpc. \label{eq:defl}
\end{align}
% % % % % % % % % % % % % % % % % % % % % % % %
Once~\plane is determined for a given~\surf and~$\bp_0$, the problem of proximity in~$\Re^3$ can be reduced the same in~$\Re^2$. This can be done by determining the three orthogonal axes of the local frame of reference located at~$\bpc$. The direction of one of the axes of the local frame along the vector normal to the plane,~$\bn$, is computed as~$\bu_3 = \bn/\norm{\bn}$. The direction of the other two axes are defined as follows~$\bu_2 = \bv_3/\norm{\bv_3}$, and~$\bu_1  = \bu_2 \times \bu_3$. The quadric~\surf and~$\bp_0$ are expressed in the orthogonal frame of reference~$\bo$-$\bu_1\bu_2\bu_3$. The surface becomes a conic in plane~$u_3=0$, e.g., ellipse, as defined in Eq.~(\ref{eq:verticalEllipse}).
\begin{equation}
|\lambda_1| u_1^2 + |\lambda_3| u_2^2 - |\gamma| = 0.
\label{eq:verticalEllipse}
\end{equation}
The constant term,~$\gamma$, in Eq.~(\ref{eq:verticalEllipse}) of a conic is obtained in the following manner. First, the quadric~\surf is represented in the local frame of reference by using an affine transformation mentioned below:    
\begin{align}
	&\bx = \bpc + \bR \bx', \quad \text{where}
	\label{eq:affineTransfLocal} \\
	&\bR = [\bu_1 | \bu_2 | \bu_3]. \nonumber
\end{align}
The point~$\bx'$ is a point in the local frame of reference. Thereafter, by substituting Eq.~(\ref{eq:affineTransfLocal}) in Eq.~(\ref{eq:generalQuadricEq}) and setting~$\bx' = \boldsymbol{0}$, the expression for~$\gamma$ is determined to be as follows:
\begin{equation}
	\gamma = S(\bpc).
	\label{eq:constConics}
\end{equation}
The standard form of conics are discussed case by case below.
% % % % % % % % % % % % % % % % % % % % % % % % % % % %
\subsubsection{Intersection of~$\plane$ and a spheroid} 
\label{sc:intPlaneSprd}
%\begin{figure}[!h]
%	\begin{subfigure}{0.45\textwidth}
%		\centering
%		\includegraphics[scale=0.4]{prolateSpheroid.pdf}
%		\subcaption{Vertical ellipse}
%		\label{fg:vertEllipse}
%	\end{subfigure}
%	\begin{subfigure}{0.45\textwidth}
%		\centering
%		\includegraphics[scale=0.3]{oblateSpheroid.pdf}
%		\subcaption{Horizontal ellipse}
%		\label{fg:horinEllipse}
%	\end{subfigure}
%	\caption{Ellipses obtained from the intersection of plane and spheroids \bnp{Figures are not required.}}	
%\end{figure}
There are two types of a spheroid, namely, {\em prolate} and {\em oblate} spheroid, which are converted into vertical and horizontal ellipses, respectively, after the intersection of the plane. The equation of both vertical ellipse~($|\lambda_1| > |\lambda_3|$) and horizontal ellipse~($|\lambda_1| < |\lambda_3|$) are the same and is defined in Eq.~(\ref{eq:verticalEllipse}).

% % % % % % % % % % % % % % % % % % % % % % % % % % % % % % 
\subsubsection{Intersection of $\plane$ with hyperboloid of one sheet and hyperboloid of two sheets}
\label{sc:intPlanehyp}
%\begin{figure}
%	\begin{subfigure}{0.45\textwidth}
%		\centering
%		\includegraphics[scale=0.3]{hyperbolaXintercept.pdf}
%		\subcaption{Hyperbola with major axis along the~$\bu_1$-axis}
%		\label{fg:hypXint}
%	\end{subfigure}
%	\hspace{0.5 cm}
%	\begin{subfigure}{0.45\textwidth}
%		\centering
%		\includegraphics[scale=0.16]{hyperbolaYintercept.pdf}
%		\subcaption{Hyperbola with major axis along the~$\bu_2$-axis}
%		\label{fg:hypYint}
%	\end{subfigure}
%	\caption{Hyperbol\ae~obtained from the intersection of the plane and hyperboloid}
%\end{figure}
The intersection of~$\plane$ with a hyperboloid of one sheet results in a hyperbola with the major axis along the~$\bu_1$-axis. The equation for the hyperbola is defined as:
\begin{equation}
	|\lambda_1| u_1^2 - |\lambda_3| u_2^2 - |\gamma| = 0.
	\label{eq:hyperbolaXint}
\end{equation}
Similarly, the intersection of~$\plane$ with a hyperboloid of two sheets results in a hyperbola with the major axis along the~$\bu_2$-axis and the equation for the same is defined as: 
\begin{equation}
|\lambda_3| u_2^2 - |\lambda_1| u_1^2 - |\gamma| = 0.
\label{eq:hyperbolaYint}
\end{equation}
% % % % % % % % % % % % % % % % % % % % % % %
\subsubsection{Intersection of~$\plane$ with a cone}
\label{sc:intPlaneCone}
%\begin{figure}[!h]
%	\centering
%	\includegraphics[scale=0.4]{intStlines.pdf}
%	\caption{Pair of intersection of lines obtained from the intersection of a plane and a cone\bnp{Figure is not required}}
%	\label{fg:pairIntLine}
%\end{figure}
The intersection of~$\plane$ and a cone results in a pair of intersecting straight lines passing through the origin~$\bo$. The equation for the pair of intersecting lines passing through the origin is defined as:   
\begin{equation}
	\sqrt{|\lambda_1|} u_1 \pm \sqrt{|\lambda_3|} u_2 = 0.
	\label{eq:pairOntLines}
\end{equation}
%%%%%%%%%%%%%%%%%%%%%%%%%%%%%%%%%%%%%%%%%%%%%%%%%%%%%%%%%%%%%%%
\subsubsection{Intersection of~$\plane$ with a sphere}
\label{sc:intPlaneSphr}
As a sphere is symmetric about all the three axes, and its intersection with~$\plane$ is a circle. The equation of the circle is defined as: 
\begin{equation}
	u_1^2 + u_2^2 - \left|\frac{\gamma}{\lambda_1}\right| = 0, \quad \lambda_1 \neq 0.
	\label{eq:circEq}
\end{equation}
% % % % % % % % % % % % % % % % % % % % % % % % % % % % %
\subsubsection{Intersection of~$\plane$ with a paraboloid}
\label{sc:intPlaneParab}
%\begin{figure}
%	\begin{minipage}{0.45\textwidth}
%		\centering
%		\vspace{4 cm}
%		\includegraphics[scale=0.3]{parabola.pdf}
%		\caption{Parabola obtained from the intersection of a plane and a paraboloid\bnp{Figure is not required.}}
%		\label{fg:parabola}
%	\end{minipage}
%	\begin{minipage}{0.45\textwidth}
%		\centering
%		\includegraphics[scale=0.3]{circle.pdf}
%		\caption{Circle obtained from the intersection of a plane and a sphere\bnp{Figure is not required.}}
%		\label{fg:circle}
%	\end{minipage}
%\end{figure}
The intersection of~$\plane$ with a paraboloid yields a parabola whose axis of symmetry is along the~$\bu_2$-axis. The equation of the parabola is as follows:
\begin{align}
	&\lambda_1 u_1^2 + \gamma_\text{l} u_2 = 0, \quad \text{where} \label{eq:parabola} \\
	&\gamma_\text{l} = 2\bu_2 \cdot \left(\bc + \bB \bpc\right). \nonumber
\end{align}
%~\bnp{A question was asked on the derivation of~$\gamma'$. Needs a discussion!}
% % % % % % % % % % % % % % % % % % % % % % % % % % % % % % % %
\subsubsection{Intersection of~$\plane$ with a cylinder}
\label{sc:intPlaneCylinder}
The intersection of~$\plane$ with a cylinder yields a pair of non-coincident vertical parallel lines. The equation for a pair of a parallel lines is obtained as:
\begin{equation}
	\lambda_1 u_1^2 -|\gamma| = 0.
	\label{eq:cylinder}
\end{equation}
%%%%%%%%%%%%%%%%%%%%%%%%%%%%%%%%%%%%%%%%%%%%%%%%%%%%%%%
\subsection{Computation of~$\bp_\text{p}$ on~$\plane$}
\begin{figure}[!t]
	\centering
	\includegraphics[scale=0.5]{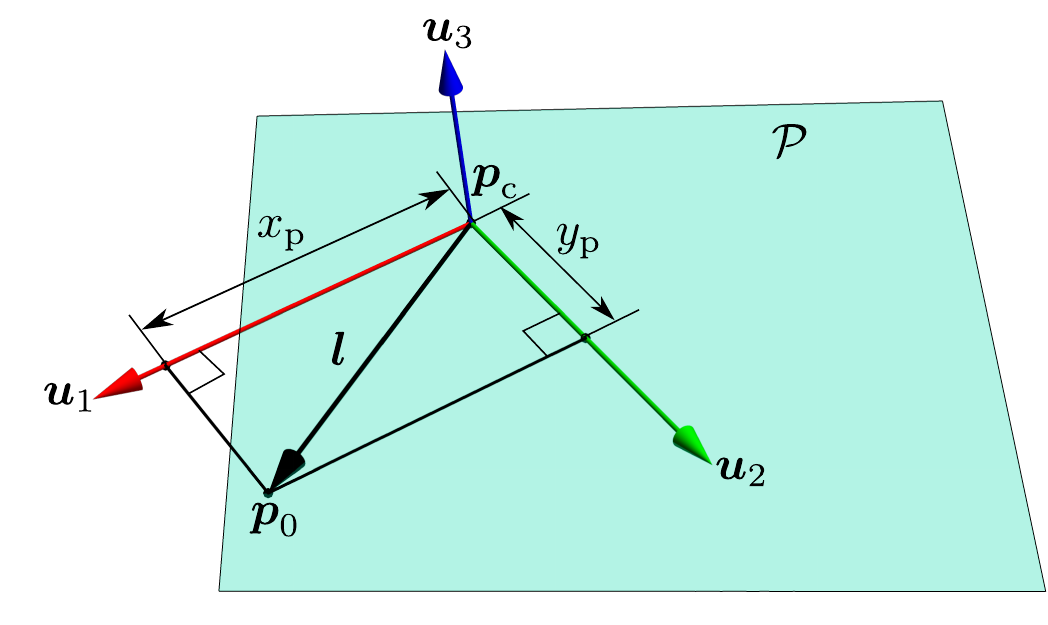}
	\caption{Expressing~$\bl$ (see Eq.~(\ref{eq:defl})) in the local frame of reference~}
	\label{fg:pointR3}
\end{figure}

Once, the standard form of the conics are defined, the point~$\bp_0$ needs to be located in the~$u_1 u_2$-plane by projecting~$\bl$ onto~$\bu_1$ and~$\bu_2$-axes, as shown in Fig.~\ref{fg:pointR3}. Hence, the point,~$\bpp$, in the plane is defined as follows: 
\begin{align}
	&\bpp = [x_\text{p}, y_\text{p}]^\top, \quad \text{where}, \label{eq:ptOnPlane}\\
	&x_p = \bu_1 \cdot \bl,~\text{and}~y_\text{p} = \bu_2 \cdot \bl.\nonumber 
\end{align}
The computation of proximity of the conics and $\bpp$ are explained in the following section.   
\subsection{Proximity of~$\bpp$ and conics in $\Re^2$}
\label{sc:shrtDistPtConics}
To compute the proximity of~$\bpp$ and a conic, the standard form of conics are derived in Sections~\ref{sc:intPlaneSprd}-\ref{sc:intPlaneSphr} and summarised below.
\begin{enumerate}
	\item Ellipse:
	\begin{equation}\eta_1:=\frac{u_1^2}{m^2} + \frac{u_2^2}{n^2}-1 =0,\quad  \text{where}~m = \sqrt{\left|\frac{\gamma}{\lambda_1}\right|}, n = \sqrt{\left|\frac{\gamma}{\lambda_3}\right|}. 
	\label{eq:stdFrmVEllip}
	\end{equation}
	\item Hyperbola with major axis along the $\bu_1$-axis:\begin{equation}
		\eta_2:=\frac{u_1^2}{m^2} - \frac{u_2^2}{n^2} - 1 =0,\quad \text{where}~m = \sqrt{\left|\frac{\gamma}{\lambda_1}\right|},n = \sqrt{\left|\frac{\gamma}{\lambda_3}\right|}. 
		\label{eq:stdFrmHypu1int}
	\end{equation}
	\item Hyperbola with major axis along the $\bu_2$-axis: 
	\begin{equation}
		\eta_3:=\frac{u_2^2}{n^2} - \frac{u_1^2}{m^2} - 1 =0, \quad \text{where $m$ and $n$ are the same as in Eq.~(\ref{eq:stdFrmHypu1int})}.
		\label{eq:stdFrmHypu2int}
	\end{equation}
	\item Pair of intersecting lines:
	\begin{equation}
		\eta_4:=\sqrt{|\lambda_1|} u_1 \pm \sqrt{|\lambda_3|} u_2= 0.
		\label{eq:stdFrmIntLines}
	\end{equation}
	\item Circle:
		\begin{equation}
			\eta_5:=u_1^2 + u_2^2 - r_0^2 = 0, \quad \text{where}~r_0=\sqrt{\left|\frac{\gamma}{\lambda_1}\right|}.
			\label{eq:stdFrmCircle} 
		\end{equation}
	\item Parabola: 
		\begin{equation}
			\eta_6:=u_1^2 - 4 \gamma u_2 = 0, \quad \text{where}~\gamma = -\frac{\gamma_\text{l}}{4\lambda_1}.\label{eq:stdFrmParab}
		\end{equation}	
	\item  Pair of non-coincident vertical parallel:
		\begin{equation}
			\eta_7:= u_1^2 - m^2 = 0,\quad \text{where}~m = \sqrt{\frac{\gamma}{\lambda_1}} .
			\label{eq:stdFrmparallelLines}
		\end{equation}
\end{enumerate}
Further, the above conics are characterised based on the value of their eccentricity,~$e$. The ratio,~$\mu$ is defined as~$m/n$, where~$m$ and~$n$ are the length of the semi-minor and semi-major axis, respectively. The value or the expression of~$e$ in terms of~$\mu$ for each case is listed in Table~\ref{tb:charConicSec}.
\begin{table}[!htb]
	\caption{Characterisation of conics based on their values of eccentricity~($e$), and~$\mu = \frac{m}{n}$}
	\centering
	\begin{tabular}{|c|l|l|}
		\hline
		Sl. no. & \makecell{Conics} & \makecell{Eccentricity ($e$)} \\
		\hline
		1 & Circle & $e = 0$ \\
		\hline
		2 & Ellipse & $e = \sqrt{1 -\mu^2}, e \in (0,1)$\\
		\hline
		3 & Parabola & $e=1$ \\
		\hline
		4 & Hyperbola& $e = \sqrt{1 +\mu^2}, e>1$\\
		\hline
		5 & Pair of intersecting lines  & $e = \infty$ \\
		\hline
		6 & Pair of non-coincident vertical parallel lines & $e = \infty$ \\
		\hline
	\end{tabular}
	\label{tb:charConicSec}
\end{table}
The proximity of~$\bpp$ to a conic is described case by case below, provided the point does not lie on the conics, i.e.,~$\eta_i(\bpp)\neq0, i=1,\dots,7$. 
\subsubsection{Proximity of~$\bpp$ to a circle $(e=0)$}
\begin{figure}[!t]
	\centering
	\begin{minipage}{0.4\textwidth}
		\centering
		\includegraphics[scale=0.6]{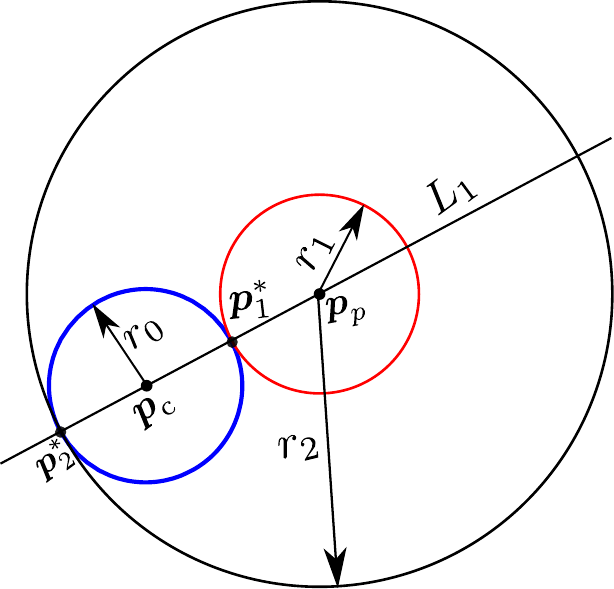}
		\caption{The circles with radii~$r_1$, and~$r_2$ that are tangent to the circle of radius~$r_0$ centred at~\pc. The inner circle has the radius~$r_\text{min} = \min (r_1,r_2)$.}
		\label{fg:circleTanCcircle}
	\end{minipage}  
	\hfill
	\begin{minipage}{0.4\textwidth}
		\centering
		\includegraphics[scale=0.25]{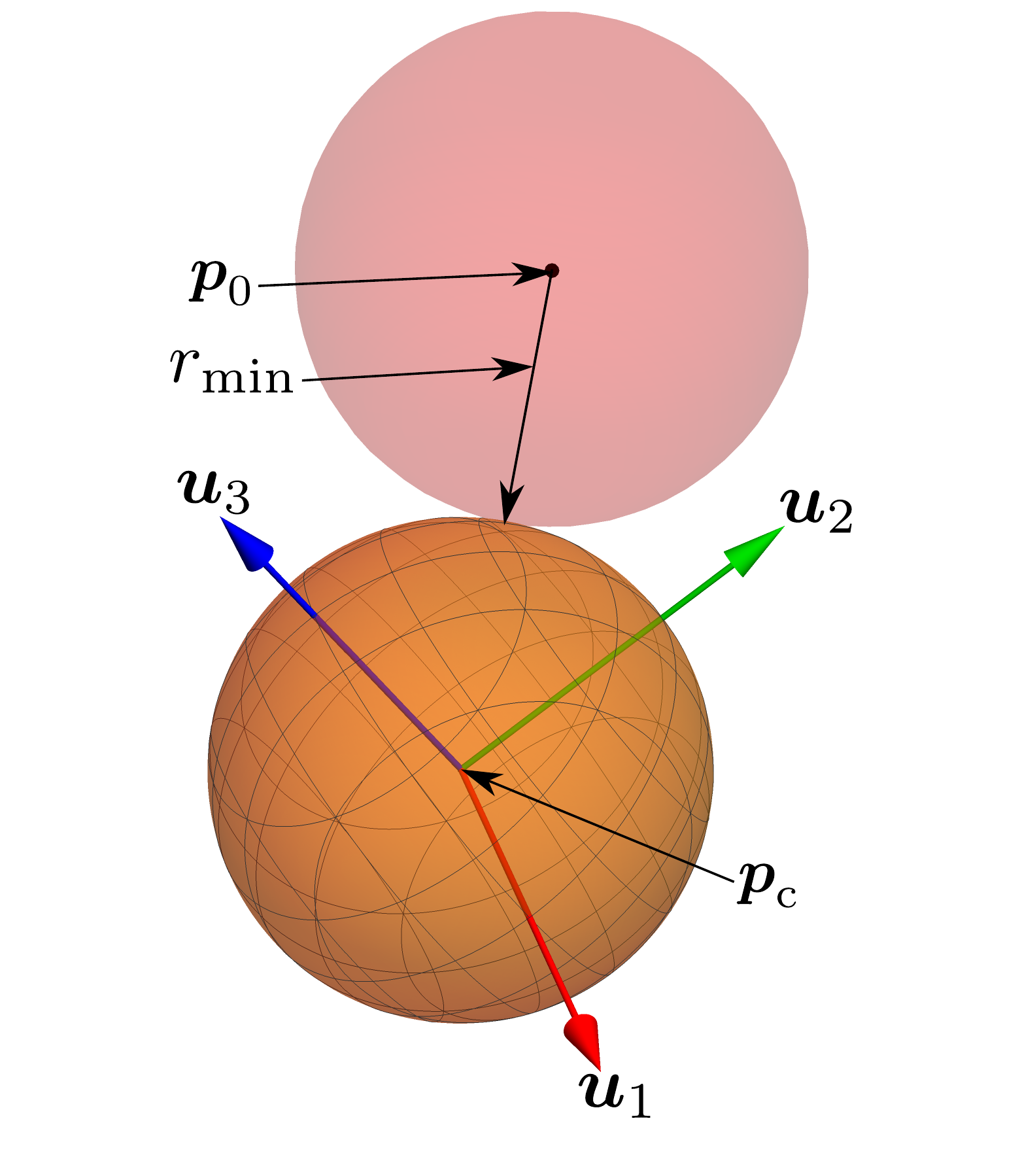}
		\caption{Distance~\rmin is the proximity of the point~$\bp_0$ to the sphere centred at~$\bpc$, which is the same as a sphere with radius~\rmin (see Table~\ref{tb:numresultAQ}) that is tangent to the sphere centred at~$\bp_0$.}
		\label{fg:sphrShortDist}
	\end{minipage}
\end{figure}
To compute the proximity of~\bpp to a circle, assume that a line,~$L_1$, passes through the origin~\pc and the point~\bpp and this line intersects the circle at the two points~$\bp^*_j,j=1,2$, as shown in Fig.~\ref{fg:circleTanCcircle}. This line is the normal to the circle at the two intersection points passing through~$\bpp$. Hence, the distances from~$\bpp$ to the point of intersections are as follows:
\begin{align}
	&r_1 = \left|\sqrt{\upone^2 + \uptwo^2} - r_0\right|, \\ 
	&r_2 = r_0 + \sqrt{\upone^2 + \uptwo^2}, 
	\label{eq:shortDistCircle}
\end{align} 
where~$r_0$ is the radius of the circle, as mentioned in Eq.~(\ref{eq:stdFrmCircle}).
The minimum value between~$r_1$, and~$r_2$ is the proximity~$\bpp$ to the circle, as shown Fig.~\ref{fg:circleTanCcircle}, which is the same as the distance between the point~$\bp_0$, and the sphere demonstrated in Fig.~\ref{fg:sphrShortDist}.
\subsubsection{Proximity of~$\bpp$ to a parabola $(e=1)$}
\label{sc:pointToParabola}
\begin{figure}[!t]
	\centering
\includegraphics[scale=0.4]{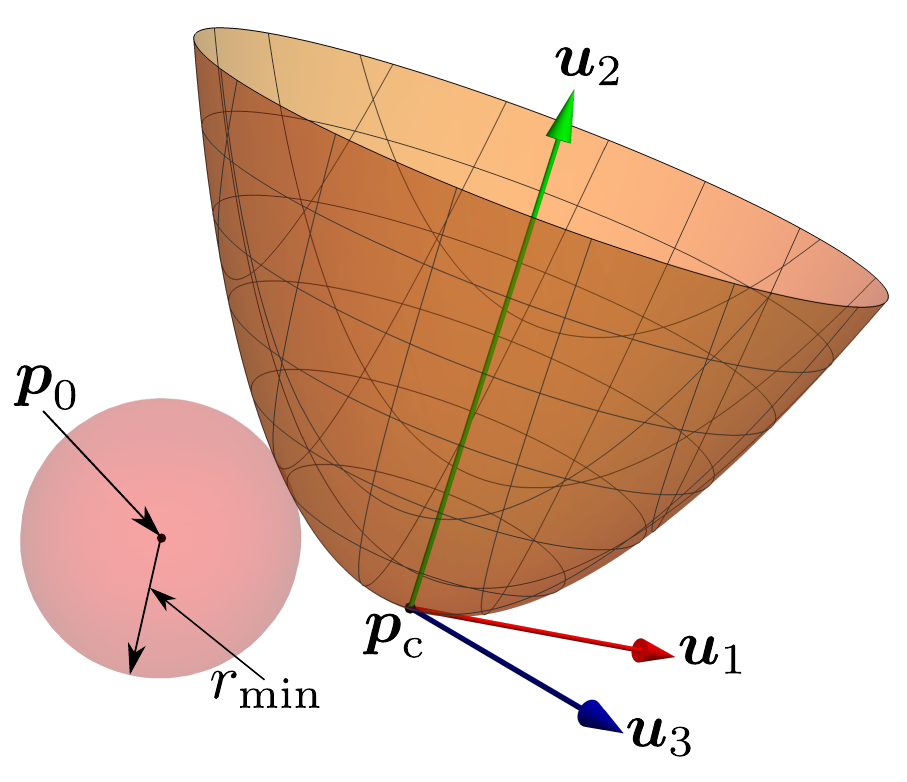}
	\caption{Proximity of~$\bp_0$ to the paraboloid; sphere with a radius~\rmin (see Table~\ref{tb:numresultAQ}) is tangent to the parabola.}
	\label{fg:parabolaShortDist}
\end{figure}
The proximity of~$\bpp$ to a parabola is the projection of~\bpp onto the said parabola at the point~$\bp^*$, as shown in Fig.~\ref{fg:shortDistParabola}. As mentioned in Section~\ref{sc:shrtDistPtConics}, the point~$\bpp$ should not lie on the parabola. Based on the location of $\bpp$, the proximity is computed for two cases, i.e., (a)~\bpp lies on the~$\bu_2$-axis, and (b)~$\bpp$ does not lie on the~$\bu_2$-axis, which are analysed below. Further, the point lies either inside or outside a parabola which is characterised by~$\eta_4(\bpp)<0$ or~$\eta_4(\bpp)>0$, respectively. 
\begin{enumerate}[label=(\alph*)]
	\item \textbf{\bpp lies on the $\bu_2$-axis}: The point,~\bpp, lies either inside or outside the parabola, which are dealt as follows:
	\begin{enumerate}
		\item[(a.1)]\textbf{\bpp lies outside parabola}: If the point~$\bpp$ is on the~$\bu_2$-axis and lie on the outside the parabola, the~$u_1$-coordinate,~$\upone$, is zero and the slope of~$L_1$ in Fig.~\ref{fg:shortDistParabola} is undefined in this case. Therefore, the proximity from~$\bpp$ to the parabola is~$r_\text{min} = |\uptwo|$, as shown in Fig.~\ref{fg:ptOnU2OutsideParab}.
		\begin{figure}[!t]
			\centering
			\begin{subfigure}{0.45\textwidth}
				\centering
				\includegraphics[width=0.7\textwidth]{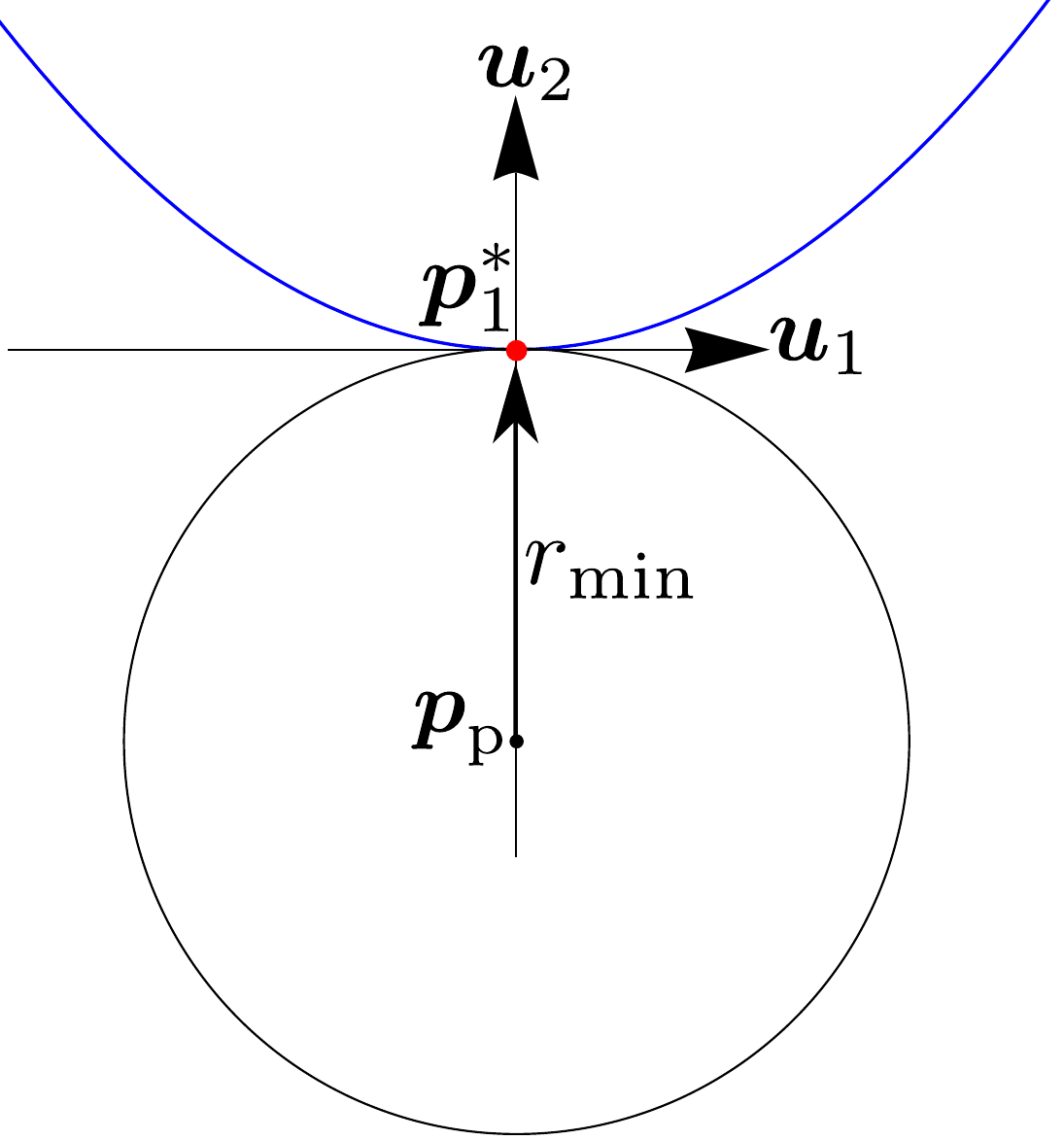}
				\subcaption{$\bpp$ lies outside the parabola; a case of one unique solution leading to only distance~\rmin}
				\label{fg:ptOnU2OutsideParab}
			\end{subfigure}
			\hfill
			\begin{subfigure}{0.45\textwidth}
				\centering
				\includegraphics[width=\textwidth]{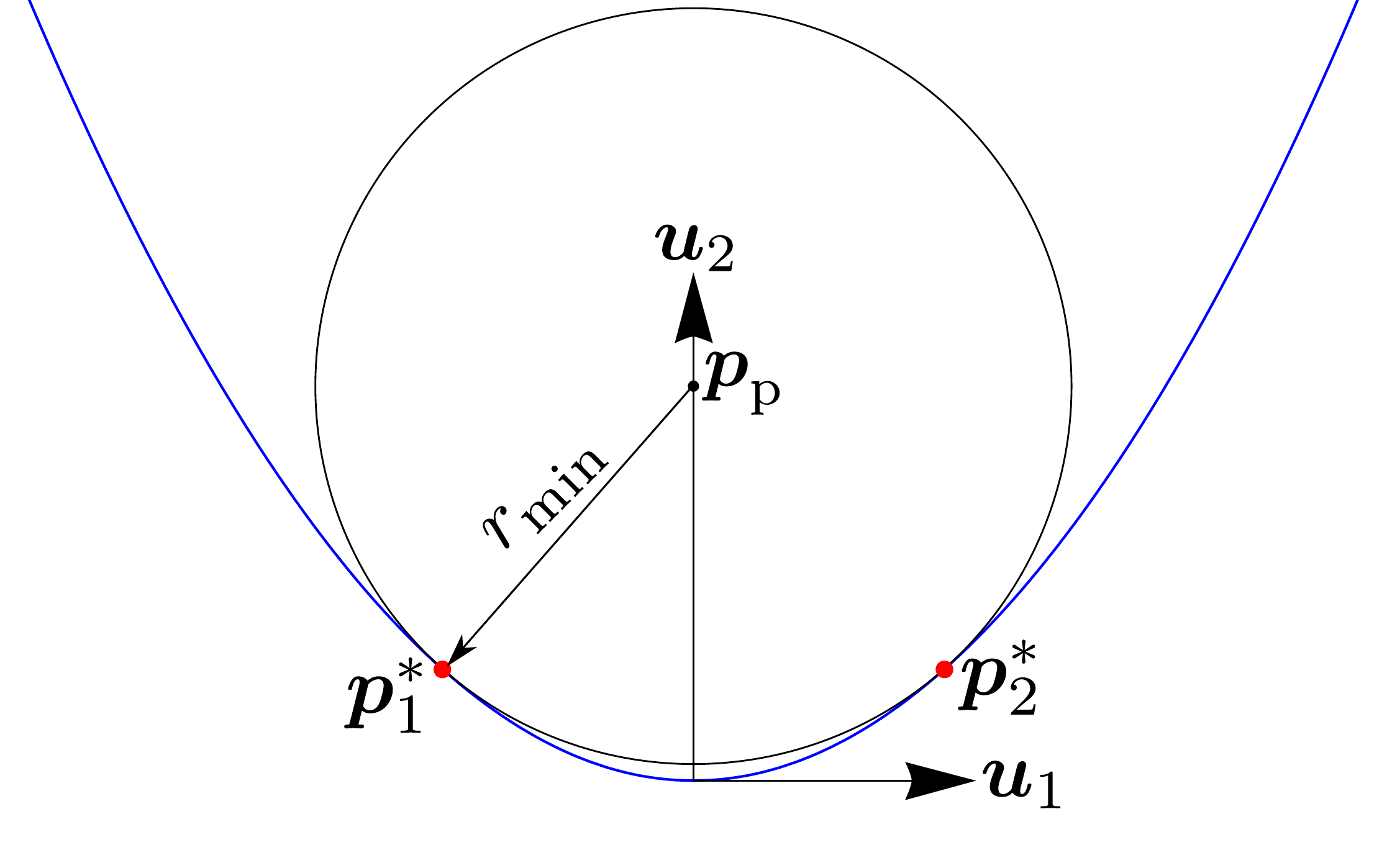}
				\subcaption{$\bpp$ lies inside the parabola; a case of two distinct solutions leading to the unique distance ~\rmin}
				\label{fg:ptOnU2InsideParab}
			\end{subfigure}
			\caption{Representation of point of tangency or tangencies, when the point $\bpp$ is on the $\bu_2$-axis}
			\label{fg:ptOnAxisParab}
		\end{figure}
		\item[(a.2)]\textbf{\bpp lies inside parabola}: If~$\bpp$ lies inside the parabola, the proximity is computed in the following manner. The line,~$L_1$, passing through the points~$\bpp$ and~$\bp^*$ intersects the~$\bu_2$-axis at~$\bt_0 = [0, t]^\top$, as shown Fig.~\ref{fg:shortDistParabola}. The equation of the line in the slope-intercept form is defined as:
		\begin{equation}
		u_2 = \frac{\uptwo-t}{\upone} u_1 + t.
		\label{eq:slopeIntLine}
		\end{equation}  
		The length of the sub-normal of the parabola is~$d_\text{N} = \sqrt{(\bt_0-\bt^*)\cdot(\bt_0-\bt^*)}$. From Fig.~\ref{fg:shortDistParabola}, the $d_\text{N}$ can be written as:
		\begin{equation}
			d_\text{N}= u_1 \tan\phi,
			\label{eq:subNormalParabola}
		\end{equation} 
		where $\phi$ is the angle between the~$\bu_2$-axis and tangent to the parabola at~$\bp^*$, as shown in Fig.~\ref{fg:shortDistParabola}. Hence, the slope of the tangent is defined as:
		\begin{equation}
			\tan \phi = \frac{\text{d}u_1}{\text{d} u_2}.
			\label{eq:slopeParabola} 
		\end{equation}
		By taking the derivative of Eq~(\ref{eq:stdFrmParab}) w.r.t.~$u_2$, Eq.~(\ref{eq:slopeParabola}) is expressed as:
		\begin{equation}
			\tan \phi = \frac{2 \gamma}{u_1}.
			\label{eq:slopeParabola2}
		\end{equation}
		Using Eq.~(\ref{eq:slopeParabola2}) in Eq.~(\ref{eq:subNormalParabola}), the length of the sub-normal is obtained to be~$2 \gamma$. Hence,~$u_2$-coordinate of~$\bp^*$ is~$t-2\gamma$. If the point~\bpp is on the~$\bu_2$-axis, and lies inside the parabola, the point~$\bt_0$ in Fig.~\ref{fg:shortDistParabola} becomes~$[0, \uptwo]^\top$. Therefore,~$u_2$-coordinate of~$\bp^*$ is obtained as:~$\uptwo-2\gamma$. Substituting~\mbox{$u_2 = \uptwo-2\gamma$} in Eq.~(\ref{eq:stdFrmParab}), and solving the quadratic equation for~$u_1$, the solutions are yielded as follows:
		\begin{equation}
		u^*_1 = \pm2 \sqrt{\gamma(\uptwo-2\gamma)}.
		\label{eq:solu1ParabEq}
		\end{equation}
		The point~$\bpp$ is on the~$\bu_2$-axis and lies inside the parabola,~$\uptwo \geq 2\gamma$, hence, Eq.~(\ref{eq:solu1ParabEq}) has either two distinct or two repeated real values. Figure~\ref{fg:ptOnU2InsideParab} depicts two points,~$\bp^*_j,j=1,2$, corresponding to distinct real solutions. The Euclidean distance from~$\bpp$ to either~$\bp^*_1$ or~$\bp^*_2$ is the desired proximity,~\rmin.
	\end{enumerate}
	\item \textbf{\bpp does not lie on the $\bu_2$-axis}: If~$\bpp$ does not lie on the~$\bu_2$-axis, $L_1$ in Eq.~(\ref{eq:slopeIntLine}) is expressed as:  
	\begin{equation}
	L_1 := \upone(u_2-t)-u_1(\uptwo-t) = 0
	\label{eq:stLineU2intercept}
	\end{equation}
	The~$u_1$-coordinate of~$\bp^*$ can be found by substituting~\mbox{$u_2 = t-2\gamma$} in Eq.~(\ref{eq:stLineU2intercept}) and solving for~$u_1$, yielding the expression for the point~$\bp^*$ as:
	\begin{equation}
	\bp^* = \left[\frac{2 \gamma \upone}{t - \uptwo}, t -2 \gamma\right]^\top.
	\label{eq:ptOfTanParabola}
	\end{equation}  
	Since~\bpp is not on the~$\bu_2$-axis,~$t$ is not equal to~\uptwo. Substitution of~$\bp^*$ from Eq.~(\ref{eq:ptOfTanParabola}) in Eq.~(\ref{eq:stdFrmParab}) results in following equation:
	\begin{equation}
	\frac{4 \gamma((t - \uptwo)^2 (2 \gamma - t) + \gamma  \upone^2)}{(t-\uptwo)^2} = 0.	
	\label{eq:cubicRatTparab}
	\end{equation} 
	The numerator of Eq.~(\ref{eq:cubicRatTparab}) is expanded to a cubic equation in~$t$ as shown below:
	\begin{equation}
	t^3  - 2 (\uptwo + \gamma) t^2 +  \uptwo (\uptwo + 4 \gamma) t - \gamma (\upone^2 + 2 \uptwo^2) = 0.
	\label{eq:cubicPolyT}
	\end{equation}
	Equation~(\ref{eq:cubicPolyT}) is reduced to a depressed cubic in the variable~$\rho$ as:
	\begin{equation}
	\rho = t - \frac{2}{3}(\uptwo + \gamma).
	\label{eq:changeOfVar}
	\end{equation}
	The depressed cubic equation and its coefficients are:
	\begin{align}
	& \rho^3 + b_0 \rho + b_1 = 0, \quad \text{where}, \label{eq:depCubicParabola}\\
	&b_0 =-\frac{1}{3}(\uptwo-2\gamma)^2, \nonumber\\
	&b_1 =  \frac{2}{27}(\uptwo-2\gamma)^3-\upone^2\gamma.\label{eq:coeffDepCubicParabCoeff2} \nonumber
	\end{align} 
	The minimum and maximum number of real roots of Eq.~(\ref{eq:depCubicParabola}) are one and three, respectively, depending on the values of $b_0$, and the discriminant which is defined as:
	\begin{equation}
	\delta = 4 b_0^3 + 27 b_1^2.
	\end{equation}
	The classification of real roots of Eq.~(\ref{eq:depCubicParabola}) and their geometric significance are discussed below. The expressions for the roots are mentioned in Appendix~\ref{ap:cubeRoot}.	
	\begin{enumerate}
		\item[(b.1)] \textbf{$b_0<0 \land \delta <0$}: Three distinct real roots and the expressions for the roots~$\rho_j,j=1,2,3$, are obtained from Eq.~(\ref{eq:threeDistRootCube}). Using~$\rho_j$, the solutions to Eq.~(\ref{eq:cubicPolyT}), i.e.~$t_j$, are obtained from Eq.~(\ref{eq:changeOfVar}), followed by the points~$\bp^*_j$ from Eq.~(\ref{eq:ptOfTanParabola}), as shown in Fig.~\ref{fg:threePointsParabola}. 
		\begin{figure}[!t]
			\centering
			\begin{subfigure}{0.45\textwidth}
				\centering
				\includegraphics[scale=0.4]{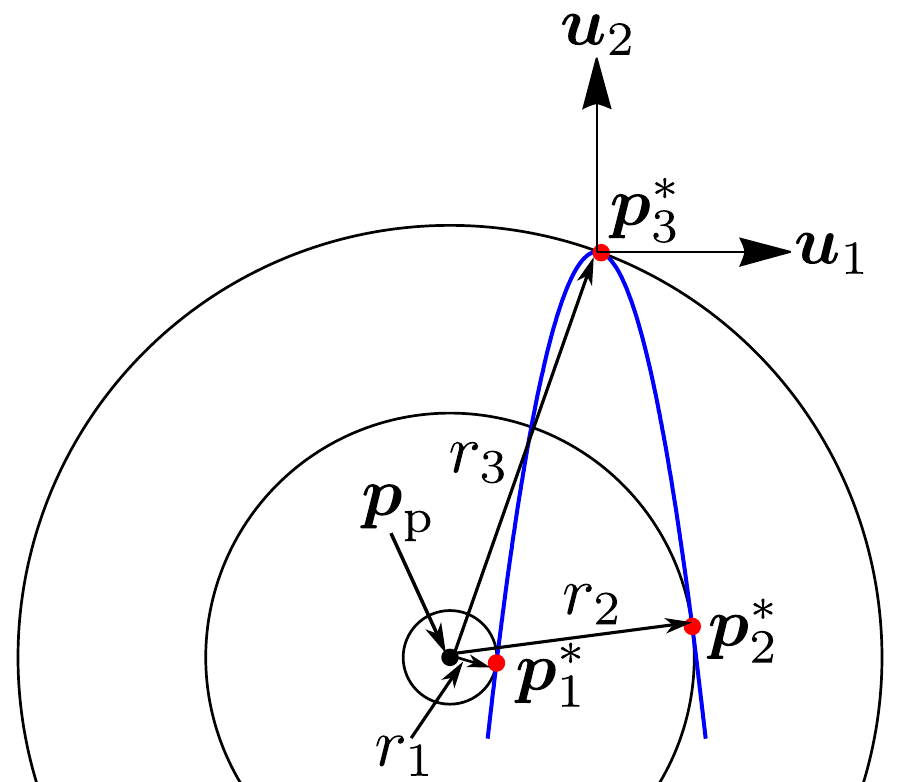}
				\subcaption{$\bpp$ outside the parabola; a case of three unique real solutions leading to three distances~$r_j,j=1,2,3$}
			\end{subfigure}
			\hfill
			\begin{subfigure}{0.45\textwidth}
				\centering
				%\vspace{-0.6 cm}
				\includegraphics[scale=0.5]{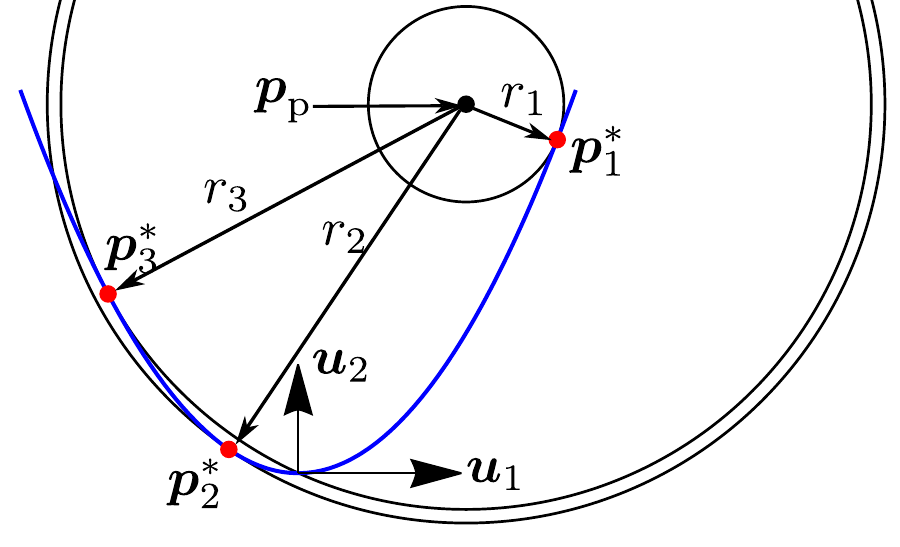}
				\subcaption{$\bpp$ inside the parabola; a case of three unique real solutions leading to three distances~$r_j,j=1,2,3$}
			\end{subfigure}
			\caption{Points, $\bp^*_j,j=1,2,3$, corresponding to the three distinct real roots of Eq.~(\ref{eq:cubicPolyT}) for a parabola, when~\bpp is not on the~$\bu_2$-axis}
			\label{fg:threePointsParabola}
		\end{figure}
		\item[(b.2)] \textbf{$b_0<0 \land \delta = 0$}: In this case, Eq.~(\ref{eq:depCubicParabola}) has one simple root and a pair of double roots. The expressions in Eq.~(\ref{eq:threeRealRootCube}) are used for computing the real roots. The points~$\bp^*_j$ corresponding to the roots~$\rho_j$ are obtained using Eq.~(\ref{eq:cubicPolyT}), followed by Eq.~(\ref{eq:ptOfTanParabola}). The points~$\bp^*_j$ are shown in Fig.~\ref{fg:threePointParabolaDel0}.  
		\begin{figure}[!t]
			\centering
			\begin{subfigure}{0.45\textwidth}
				\centering
				\includegraphics[scale=0.6]{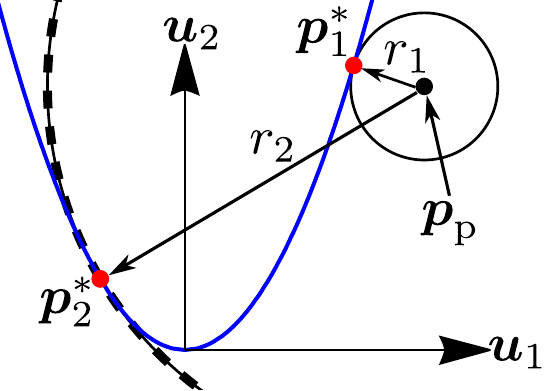}
				\subcaption{$\bpp$ outside the parabola; a case of one distinct and two repeated real roots leading to one distinct distance~$r_1$  and two repeated distances~$r_2=r_3$, respectively}
			\end{subfigure}
			\hfill
			\begin{subfigure}{0.45\textwidth}
				\centering
				%\vspace{0.5 cm}
				\includegraphics[scale=0.55]{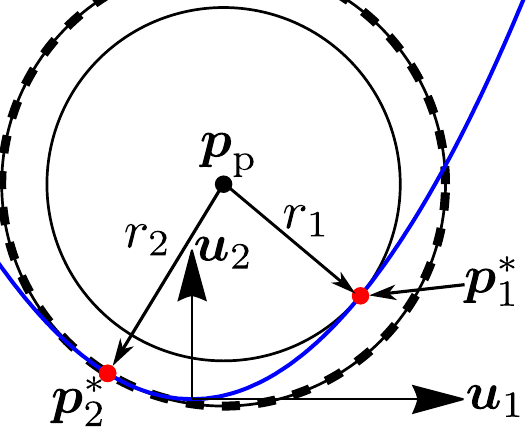}
				\subcaption{$\bpp$ inside the parabola; a case of one distinct and two repeated real roots leading to one distinct distance~$r_1$  and two repeated distances~$r_2=r_3$, respectively}
			\end{subfigure}
			\caption{Points, $\bp^*_1$, and $\bp^*_2=\bp^*_3$, corresponding to three real roots of Eq.~(\ref{eq:cubicPolyT}), when~\bpp is not on the~$\bu_2$-axis}
			\label{fg:threePointParabolaDel0}
		\end{figure}
%		If the point,~$\bpp$, is on the~$\bu_2$-axis, there are two repeated and one distinct roots of Eq.~(\ref{eq:cubicPolyT}). The values of the repeated roots are the same as~$\uptwo$, hence, for these cases, the~$\bu_1$-coordinate of Eq.~(\ref{eq:ptOfTanParabola}) are undefined. Therefore, the point,~$\bp^*_1$, corresponding the only distinct real root is the only solution, i.e.,~$\bp^*_1 = [0,0]^\top$, as shown in Fig.~\ref{fg:ptOnAxisParab}.
		
		\item[(b.3)]\textbf{$b_0<0 \land \delta >0$}: In this case, there is one real root and the expression for the real root is mentioned in Eq.~(\ref{eq:oneRealRootCube}). The solution of Eq.~(\ref{eq:cubicPolyT}) and corresponding point~$\bp^*_1$, as shown in Fig.~\ref{fg:onePointParabolaDel0}, are computed using~$\rho_1$. 
		\begin{figure}[!t]
			\centering
			\begin{subfigure}{0.45\textwidth}
				\centering
				\includegraphics[scale=0.5]{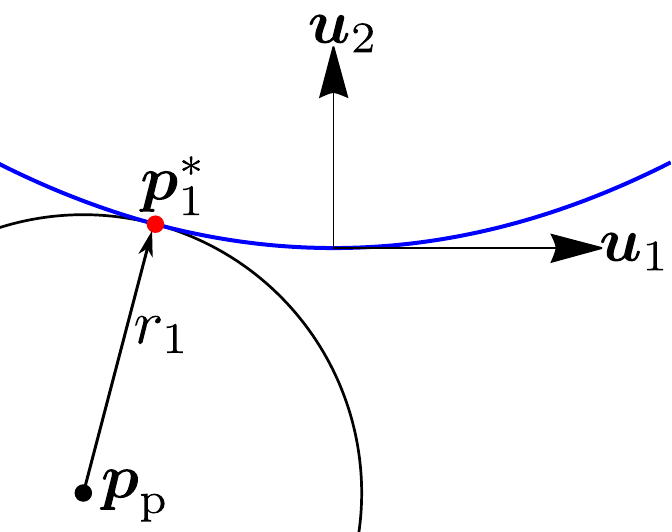}
				\subcaption{$\bpp$ outside the parabola; a case of only one real solution leading to one distance~$\rmin = r_1$}
			\end{subfigure}
			\hfill
			\begin{subfigure}{0.45\textwidth}
				\centering
				%\vspace{.5 cm}
				\includegraphics[scale=0.6]{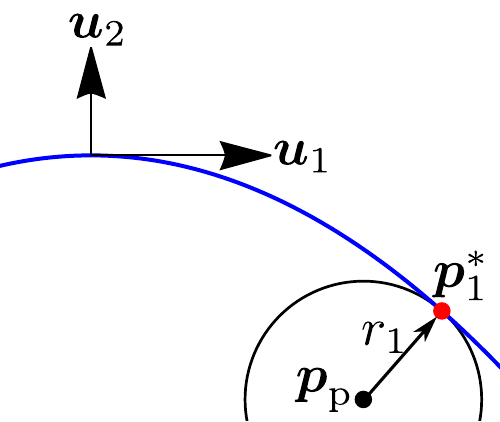}
				\subcaption{$\bpp$ inside the parabola; a case of only one real solution leading to one distance~$\rmin = r_1$}
			\end{subfigure}
			\caption{Representation of point, $\bp^*_1$, corresponding to one real root of Eq.~(\ref{eq:cubicPolyT}) for a parabola corresponding to the condition $b_0<0 \land \delta >0$, when~\bpp is not on the~$\bu_2$-axis}
			\label{fg:onePointParabolaDel0}
		\end{figure}
		\item[(b.4)]\textbf{$b_0 = 0 \land \delta >0$}: If the value of $\bu_2$-coordinate of $\bpp$ is twice of the semi-latus rectum, Eq.~(\ref{eq:depCubicParabola}) becomes $\rho^3 + b_2 = 0$, and the only real root is $\rho_1$ mentioned in Eq.~(\ref{eq:oneRealRootCubic}).
		The point, $\bp^*_1$, corresponding to $\rho_1$ is shown in Fig.~\ref{fg:onePointParabolaDel01}.
		\begin{figure}[!t]
			\centering
			\begin{subfigure}{0.45\textwidth}
				\centering
				\vspace{0.5 cm}
				\includegraphics[scale=0.55]{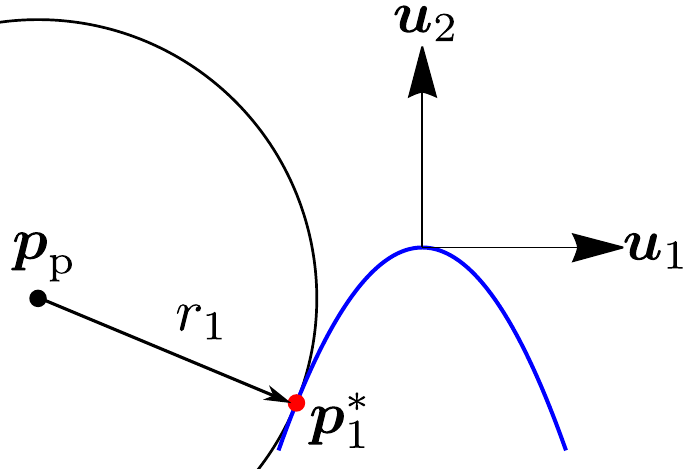}
				\subcaption{$\bpp$ outside the parabola; a case of only one real solution leading to one distance~$\rmin = r_1$}
			\end{subfigure}
			\hfill
			\begin{subfigure}{0.45\textwidth}
				\centering
				\includegraphics[scale=0.55]{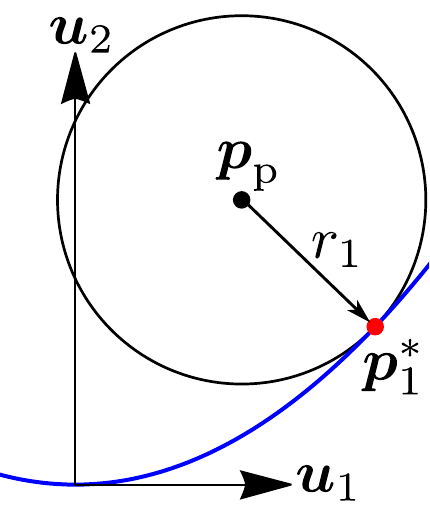}
				\subcaption{$\bpp$ inside the parabola; a case of only one real solution leading to one distance~$\rmin = r_1$}
			\end{subfigure}
			\caption{Representation of point, $\bp^*_1$, corresponding to one real root of Eq.~(\ref{eq:cubicPolyT}) for a parabola corresponding to the condition~$b_0 = 0 \land \delta >0$, when~\bpp is not on the~$\bu_2$-axis}
			\label{fg:onePointParabolaDel01}
		\end{figure}
	\end{enumerate}       
	The distances from the point $\bpp$ to $\bp^*_j,j=1,2,3,$ are computed as:
	\begin{equation}
	r_j = \sqrt{(\bp^*_j-\bpp)\cdot(\bp^*_j-\bpp)},\quad \text{where}~j = 1~\text{or}~j = 1, 2, 3.
	\label{eq:shrtDistParabola}
	\end{equation} 
	The value, which is minimum among $r_j$, is the desired proximity~\rmin of~\bpp to a parabola, as shown in Fig.~\ref{fg:shortDistParabola}, which is the same as the distance from $\bp_0$ to a paraboloid depicted in Fig.~\ref{fg:parabolaShortDist}. 	
	
\end{enumerate}
% % % % % % % % % % % % % % % % % % % % % % % % % % % % % % % % % % % % % % %
\subsubsection{Proximity of a point to ellipse $(e\in(0,1))$ and proximity of a point to hyperbola $(e>1)$}
\label{sc:pointToEllipHyp}
In this section, the ellipse and hyperbola whose, major axis is along the~$\bu_2$-axis, are considered for the analysis. Therefore, the principal axes of the ellipse and hyperbola with major axis along the~$\bu_1$-axis need to be rotated by an angle~$\pi/2$ about the out of plane axis in the CW manner. The change of coordinate frame results in an interchange of coefficients of the quadratic terms of Eq.~(\ref{eq:stdFrmVEllip}) leads to an rotated ellipse as follows:
\begin{align}
&\overline{\eta_1}:=\frac{u_1^2}{m^2} + \frac{u_2^2}{n^2} - 1 = 0,\quad n>m,\quad \text{where} \label{eq:horizontalEllipse} \\
&m = \sqrt{\left|\frac{\gamma}{\lambda_3}\right|},~\text{and}~n = \sqrt{\left|\frac{\gamma}{\lambda_1}\right|}.  \nonumber
\end{align} 
Similarly, the interchange of variables of the quadratic terms of~$\eta_2 = 0$ (see Eq.~(\ref{eq:stdFrmHypu1int})) leads to the equation ($\eta_3 = 0$ mentioned in Eq.~(\ref{eq:stdFrmHypu2int})) of a rotated hyperbola associated with the hyperboloid of two sheets. Due to the change of co-ordinates, the point~\bpp corresponding to these ellipse and hyperbola is re-defined as follows:
\begin{equation}
	\bpp = [-y_\text{p}, x_\text{p}]^\top.
	\label{eq:pointPlaneEllipHyp}
\end{equation}
The equation of ellipse and hyperbola defined in Eqs.~(\ref{eq:stdFrmVEllip},~\ref{eq:stdFrmHypu2int}), respectively, may be rewritten in terms of~$e_1 = e^2 - 1$, and by multiplying~$m^2$ on both sides and replacing~$\mu^2 = -e_1$, as follows:
\begin{align}
&\text{Ellipse:}\quad u_1^2 - e_1 u_2^2+ n^2 e_1  = 0,\quad \text{where} ~0<e<1,\label{eq:stEllipseEqEccen}\\
&\text{Hyperbola:}\quad u_1^2 + e_1 u_2^2 - n^2 e_1  = 0,\quad \text{where} ~e>1.\label{eq:stHypEqEccen1}
\end{align}
It is sufficient to choose any of the two Eqs.~(\ref{eq:stEllipseEqEccen},~\ref{eq:stHypEqEccen1}) for computing proximity because both of the equations act in the same way based on the values of~$e$. Therefore, Eq.~(\ref{eq:stEllipseEqEccen}) is chosen for further analysis. Using the properties of an ellipse and a hyperbola, e.g., normal, and sub-normal, and the position of~$\bpp$, the procedure for computing proximity of~$\bpp$ to a curve is explained below. 

The equation of the normal of the vertical ellipse, mentioned in Eq.~(\ref{eq:stdFrmVEllip}), that intersecting the~$\bu_2$-axis at~$\bt_0=[0,t]^\top$, at the point~$\bp^*$, as shown in Fig.~\ref{fg:shortDistanceEllipse}, is defined as:
\begin{align}
	&\frac{u^*_2}{n^2}(u_1-u^*_1) - \frac{u^*_1}{m^2}(u_2-u_2^*) = 0, \nonumber\\
	\implies &\frac{u^*_2}{n^2}(-u^*_1) = \frac{u^*_1}{m^2}(t-u_2^*), \nonumber\\
	\implies &t = u^*_2 - \mu^2u^*_2.
	\label{eq:normalEqEllipse}
\end{align} 
In Eq.~(\ref{eq:normalEqEllipse}), the term~$\mu^2u^*_2$ is the length of the sub-normal,~$d_\text{N}$ (see Fig.~\ref{fg:shortDistanceEllipse}). Equation~(\ref{eq:normalEqEllipse}) can be written as:\\
\begin{equation}
	t = u^*_2 e^2.
	\label{eq:normalEqEllipseEccen}
\end{equation} 
Similarly, the normal equation of a hyperbola mentioned in Eq.~(\ref{eq:stdFrmHypu2int}), which intersects the~$\bu_2$-axis at~$\bt_0=[0,t]^\top$, at the point~$\bp^*$, as shown in Fig.~\ref{fg:shortDistancehyperbola}, is defined as:
\begin{align}
&\frac{u^*_2}{n^2}(u_1-u^*_1) + \frac{u^*_1}{m^2}(t-u_2^*) = 0, \nonumber\\
\implies &\frac{u^*_2}{n^2}(u^*_1) = \frac{u^*_1}{m^2}(t-u_2^*), \nonumber\\
\implies &t = u^*_2 + \mu^2u^*_2.
\label{eq:normalEqhyperbola}
\end{align}  
From Eq.~(\ref{eq:normalEqhyperbola}), the length of the sub-normal of a hyperbola,~$d_\text{N} = \mu^2u^*_2$ (refer Fig.~\ref{fg:shortDistancehyperbola}), which is the same as an ellipse, and the equation is reduced to Eq.~(\ref{eq:normalEqEllipseEccen}), where~$e = \sqrt{1+\mu^2}$. From Eqs.~(\ref{eq:normalEqEllipseEccen},~\ref{eq:normalEqhyperbola}),~$u_2^*$ is expressed as~$t/e^2$ and the corresponding value of~$u_1$-coordinate of~$\bp*$,~$u^*_1$, is obtained based on the location of~$\bpp$.

Like the parabola in Section~\ref{sc:pointToParabola}, there are two cases to analyse, i.e., (a)~$\bpp$ lies on the major and minor axes, and (b)~$\bpp$ does not lie on the major and minor axes. Further, the point~\bpp may lie either inside~($\overline{\eta_1}(\bpp)~\text{and}~\eta_3(\bpp)<0$) or outside~($\overline{\eta_1}(\bpp)~\text{and}~\eta_3(\bpp)>0$) an ellipse and a hyperbola.
\begin{enumerate}[label=(\alph*)]
	\item \textbf{$\bpp$ lies on the major axis}: If the point~$\bpp$ lies on the major axis, i.e.,~$\bu_2$-axis, the length of the sub-normal $d_\text{N}$ is not zero, and~\mbox{$u^*_2 =\uptwo/e^2$} and the expression for the~$u_1$-coordinate of~$\bp^*$ is determined based on the position of~\bpp w.r.t. an ellipse and a hyperbola, as listed below.
	\begin{enumerate}
		\item[(a.1)]\textbf{$\bpp$ lies inside an ellipse and a hyperbola}: Substituting the expression of~$u^*_2$ in place of~$u_2$ in Eq.~(\ref{eq:stEllipseEqEccen}), the equation becomes:
		\begin{equation}
			\frac{e^4 u_1^2 + e_1(n^2e^4-\uptwo^2)}{e^4} = 0, \quad \text{where}~e \neq 0.
			\label{eq:quadEqPtOnAxisInEllip} 
		\end{equation} 
		Equation~(\ref{eq:quadEqPtOnAxisInEllip}) is further simplified to a quadratic equation as:
		\begin{equation}
			u_1^2 + e_1(n^2-u^{*^2}_2) = 0.
			\label{eq:quadEqPtOnAxisInEllip2}
		\end{equation}
		The solution of~$u_1$ are distinct real roots because for an ellipse,~\mbox{$e_1<0$}, and~$n>u^*_2$, and for a hyperbola,~$e_1>0$, and~$n<u^*_2$. The~$u_1$-coordinate of~\mbox{$\bp^*_j,j=1,2$}, are corresponding to the distinct real roots of Eq.~(\ref{eq:quadEqPtOnAxisInEllip2}).
		\begin{figure}[!t]
			\centering
			\begin{subfigure}{0.45\textwidth}
				\centering
				\includegraphics[scale=0.3]{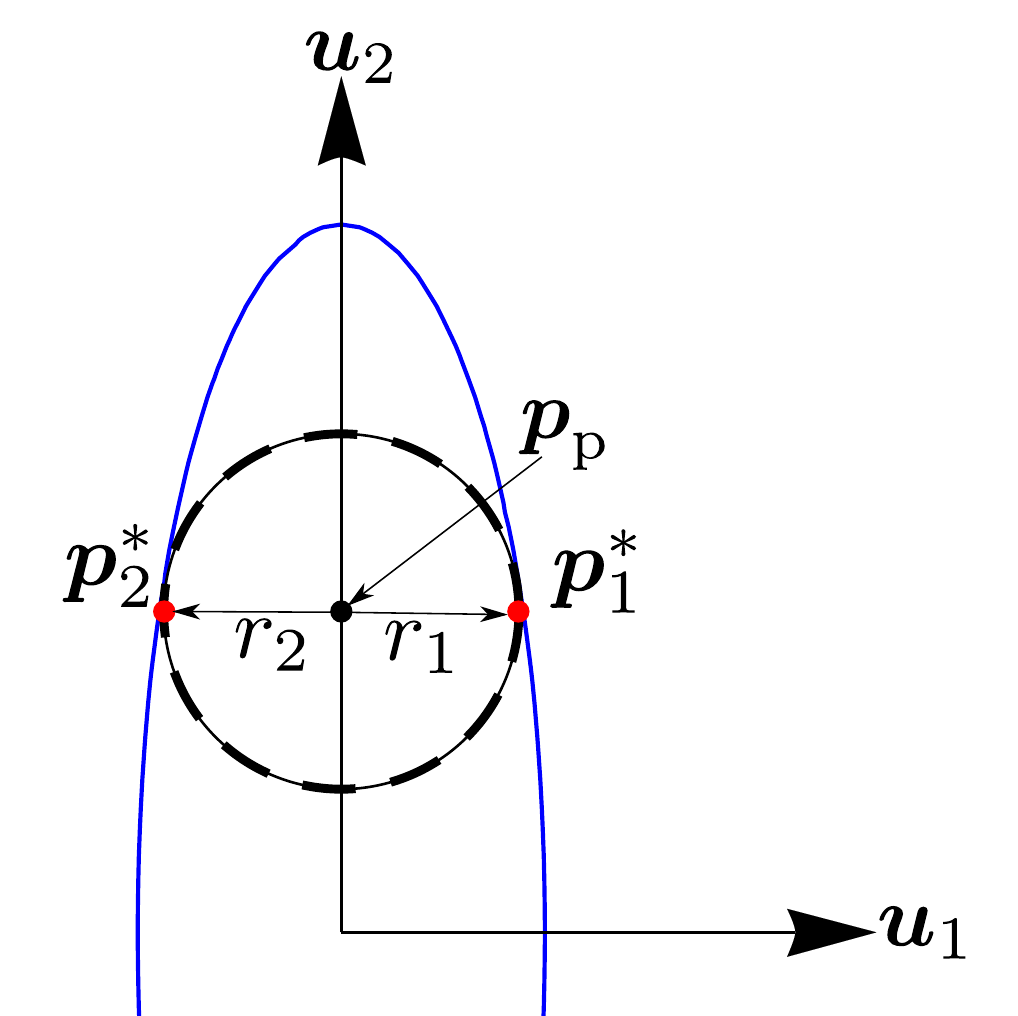}
				\subcaption{$\bpp$ lies inside the ellipse; a case of two distinct solutions leading to two distinct distances~$r_1,r_2$}
				\label{fg:elliptworootonaxisinside}
			\end{subfigure}
			\hfill
			\begin{subfigure}{0.45\textwidth}
				\centering
				\includegraphics[scale=0.2]{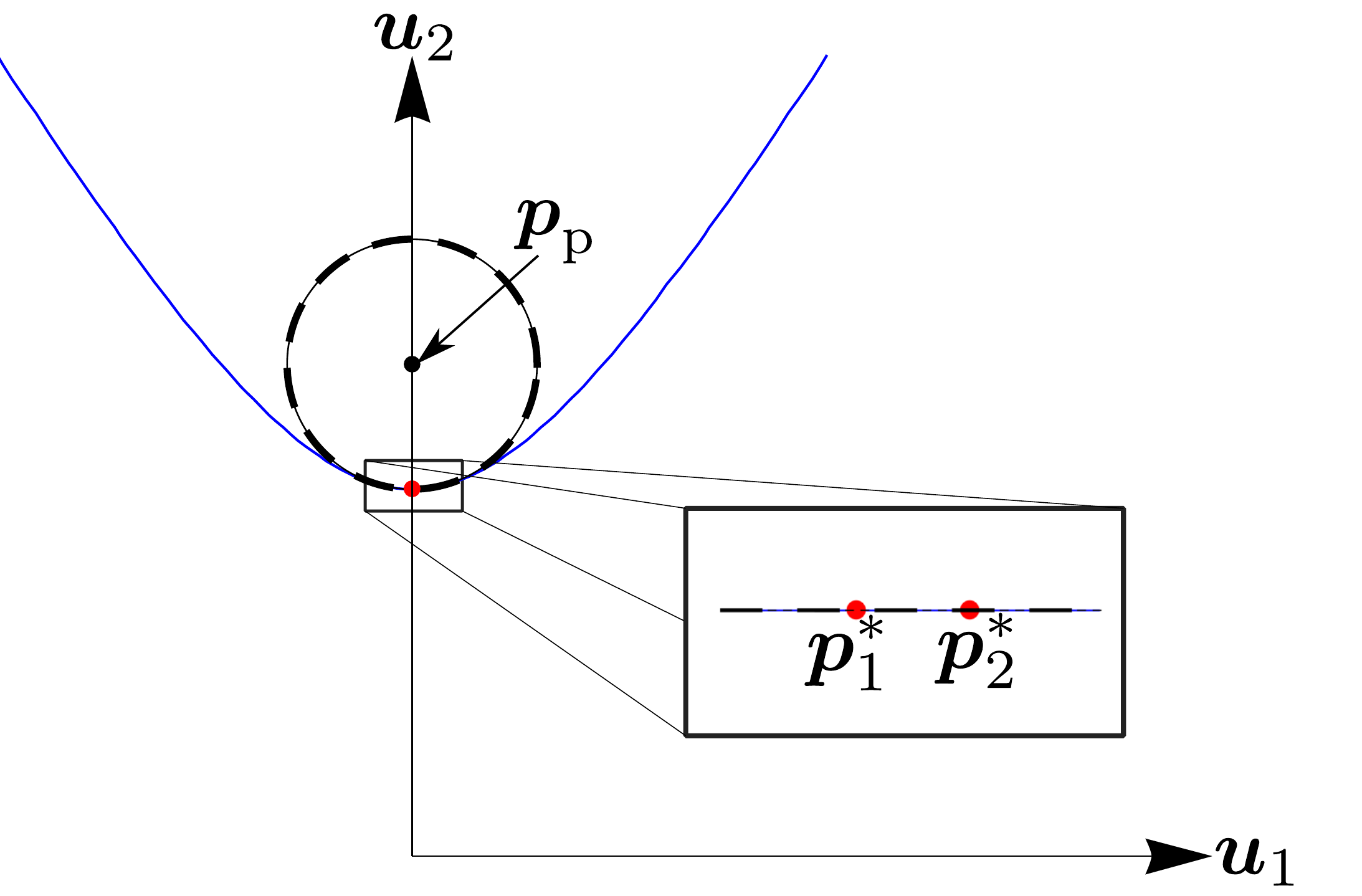}
				\subcaption{$\bpp$ lies inside the Hyperbola; a case of two distinct solutions leading to two distinct distances~$r_1,r_2$}
				\label{fg:hyptworootonaxisinside}
			\end{subfigure}
			\caption{Points,~$p^*_j,j=1,2$, corresponding to the distinct real roots of Eq.~(\ref{eq:quadEqPtOnAxisInEllip2}) corresponding to an ellipse and a hyperbola, when~\bpp is on the~$\bu_2$-axis}
			\label{fg:ptInsideEllipOnU2axis}
		\end{figure}
		The proximity~\rmin is the minimum Euclidean distance among the two distances between points~$\bpp$ and~$\bp^*_1$, and between the points~$\bpp$ and~$\bp^*_2$, as shown in Figs.~\ref{fg:elliptworootonaxisinside},~\ref{fg:hyptworootonaxisinside}.
		\item[(a.2)]\textbf{\bpp lies outside an ellipse and a hyperbola}: If the point~\bpp lies outside an ellipse and a hyperbola, there are two complex roots for Eq.~(\ref{eq:quadEqPtOnAxisInEllip2}) because the discriminant of the equation is always less than zero. The points~$\bp^*_j,j=1,2$, are on the major axis and defined as:~$\bp^*_1 = [0,\uptwo]^\top$, and~$\bp^*_2 = [0,-\uptwo]^\top$, respectively. The corresponding distances from~$\bpp$ are~$r_1 = \left|n-\uptwo\right|$, and~$r_2 = \left|n+\uptwo\right|$, as shown in Figs.~\ref{fg:elliptworootonaxisoutside},~\ref{fg:hyptworootonaxisoutside}. The  proximity~\rmin is the minimum between~$r_1$, and~$r_2$.
		\begin{figure}[!t]
			\centering
			\begin{subfigure}{0.45\textwidth}
				%\vspace{1.2 cm}
				\centering
				\includegraphics[scale=0.3]{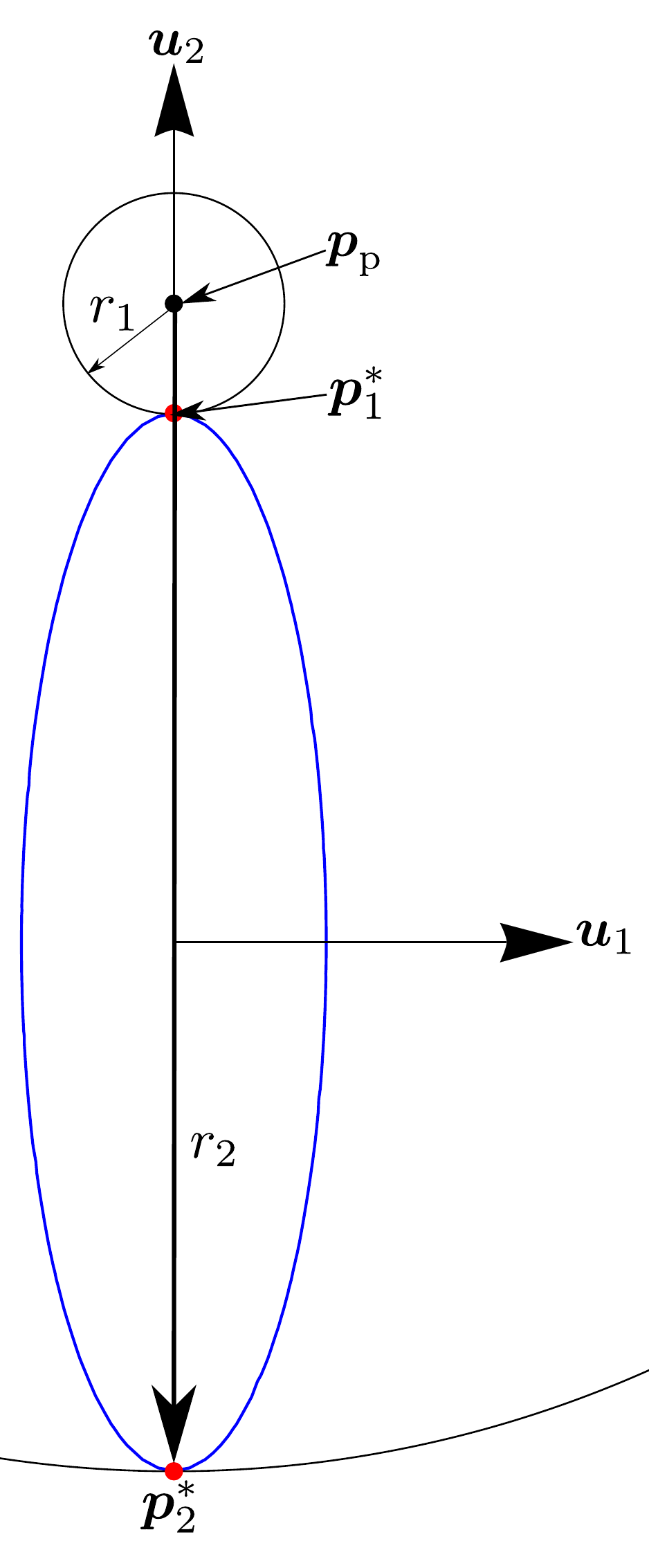}
				\subcaption{$\bpp$ lies outside the ellipse; a case of two distinct solutions leading to two distances~$r_1,r_2$}
				\label{fg:elliptworootonaxisoutside}
			\end{subfigure}
			\hfill
			\begin{subfigure}{0.45\textwidth}
				\centering
				\includegraphics[scale=0.2]{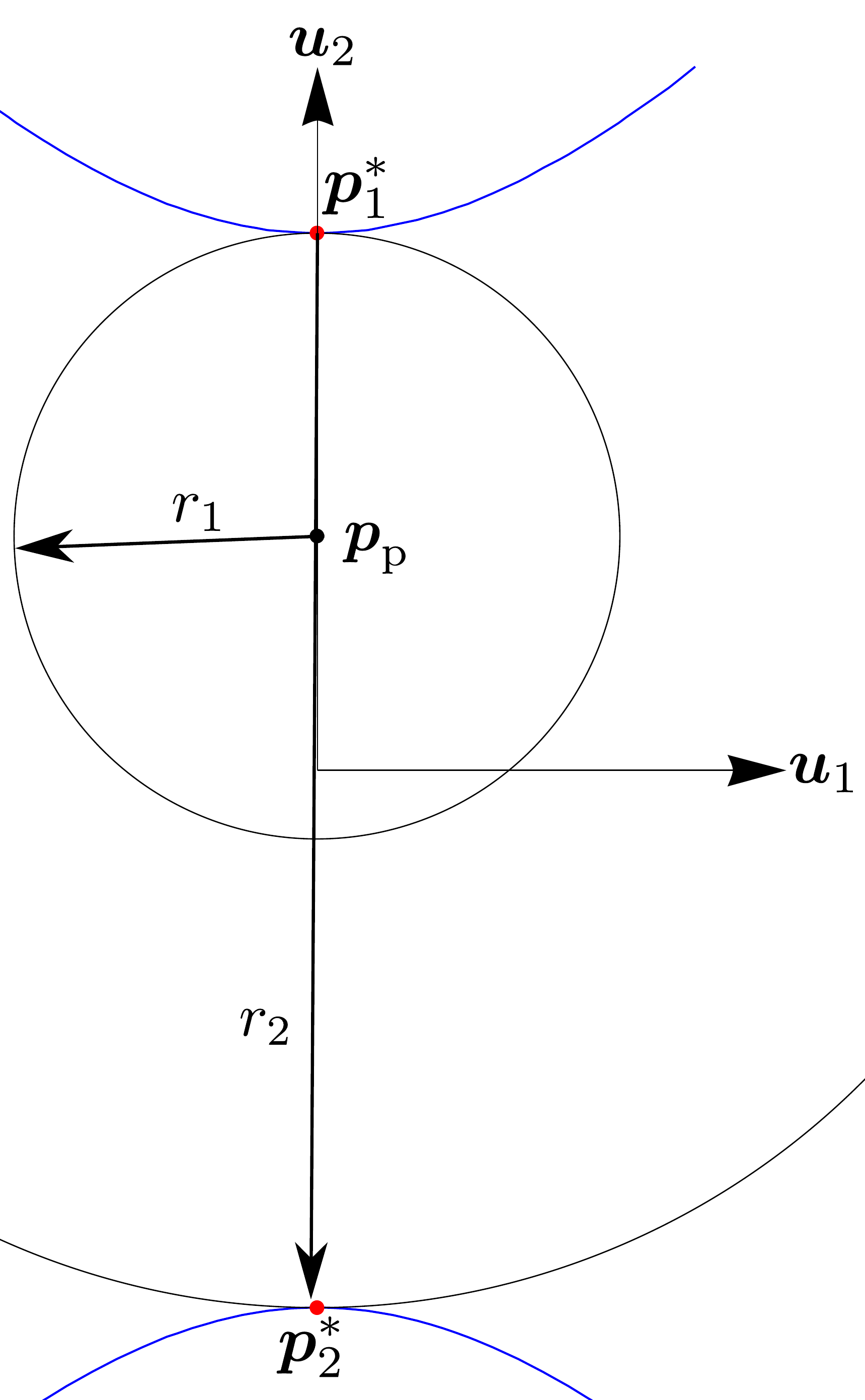}
				\subcaption{$\bpp$ lies outside of the hyperbola; a case of two distinct solutions leading to two distances~$r_1,r_2$}
				\label{fg:hyptworootonaxisoutside}
			\end{subfigure}
			\caption{Points,~$\bp^*_j,j=1,2$, corresponding to the real roots of Eq.~(\ref{eq:quadEqPtOnAxisInEllip2}) for an ellipse and a hyperbola, when~\bpp is on the~$\bu_2$-axis}
			\label{fg:ptOutsideEllipOnU2axis}
		\end{figure}
	\end{enumerate}
	\item \textbf{$\bpp$ lies on the minor axis}: In this case, the point~$\bpp$ lies either outside or inside an ellipse. However, the procedure for computing~$\bp^*_j$ is the same for both the cases. Hence, the case in which~$\bpp$ lies outside an ellipse is mentioned below. Similarly, for a hyperbola, the case where~$\bpp$ always lies outside the curve is discussed below. 
	\begin{enumerate}
		\item[(b.1)]\textbf{$\bpp$ lies outside an ellipse}: If the point~$\bpp$ lies outside an ellipse, the~$u_2$-coordinate of~$\bp^*$ is zero. Hence, the points~$\bp^*_j,j=1,2$, are defined as follows:~\mbox{$\bp^*_1 = [m,0]^\top$}, and~$\bp^*_1 = [-m,0]^\top$. Therefore, the distances from~$\bpp$ to~$\bp^*_1$, and~$\bp^*_2$, are~\mbox{$r_1 = \left|\upone-m\right|$}, and~$r_2 = \left|\upone+m\right|$, respectively, as shown in Fig.~\ref{fg:ptOnXaxisOutEllip}. The minimum value among~$r_1$ and~$r_2$ is the proximity~\rmin. 
		\begin{figure}[!t]
			\centering
			\begin{subfigure}{0.45\textwidth}
				\includegraphics[scale=0.25]{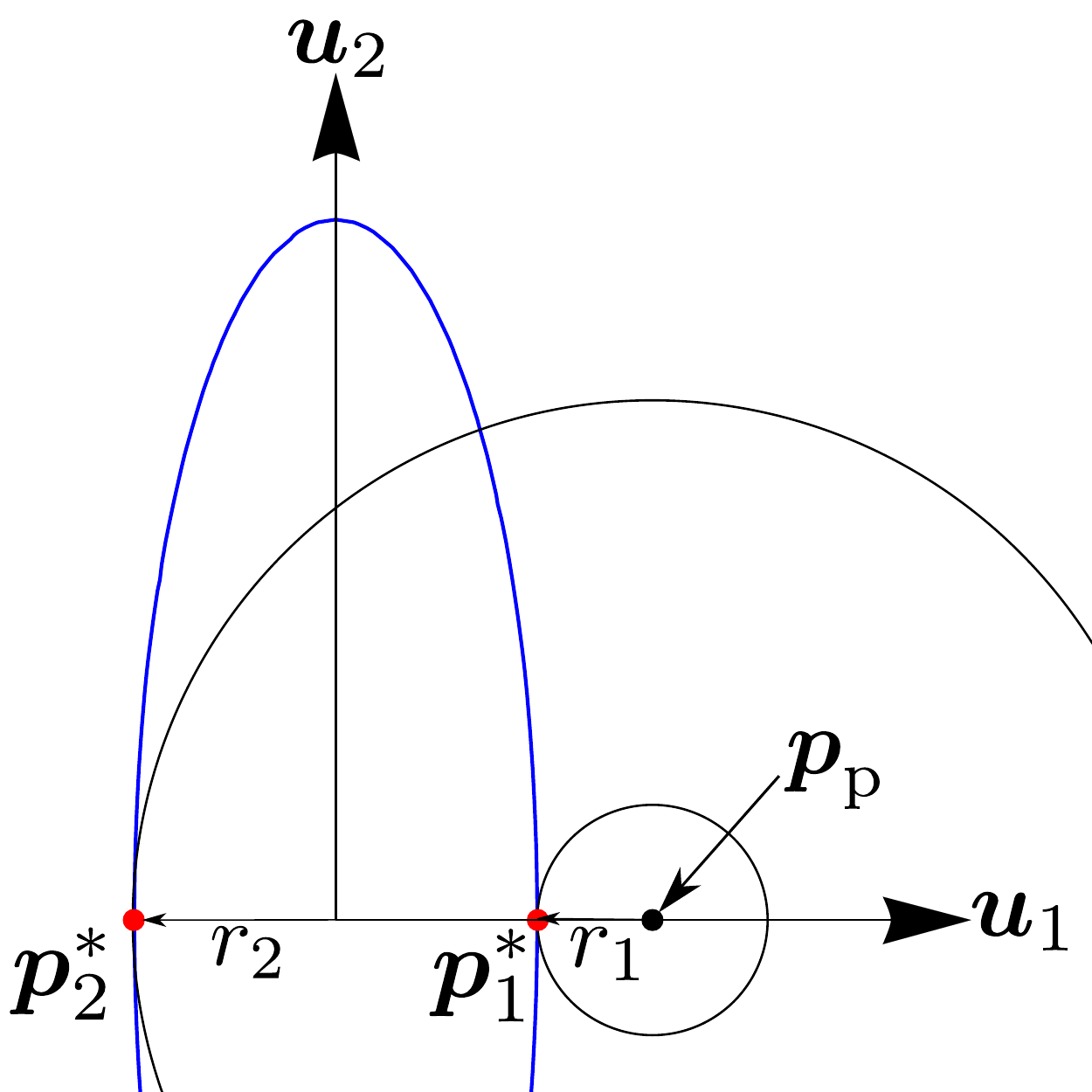}
				\subcaption{$\bpp$ lies outside the ellipse; two distances~$r_1,r_2$ are computed based on the length of the semi-major axis~$m$}
				\label{fg:ptOnXaxisOutEllip}
			\end{subfigure}
			\hfill
			\begin{subfigure}{0.45\textwidth}
				%\vspace{1.5 cm}
				\includegraphics[scale=0.3]{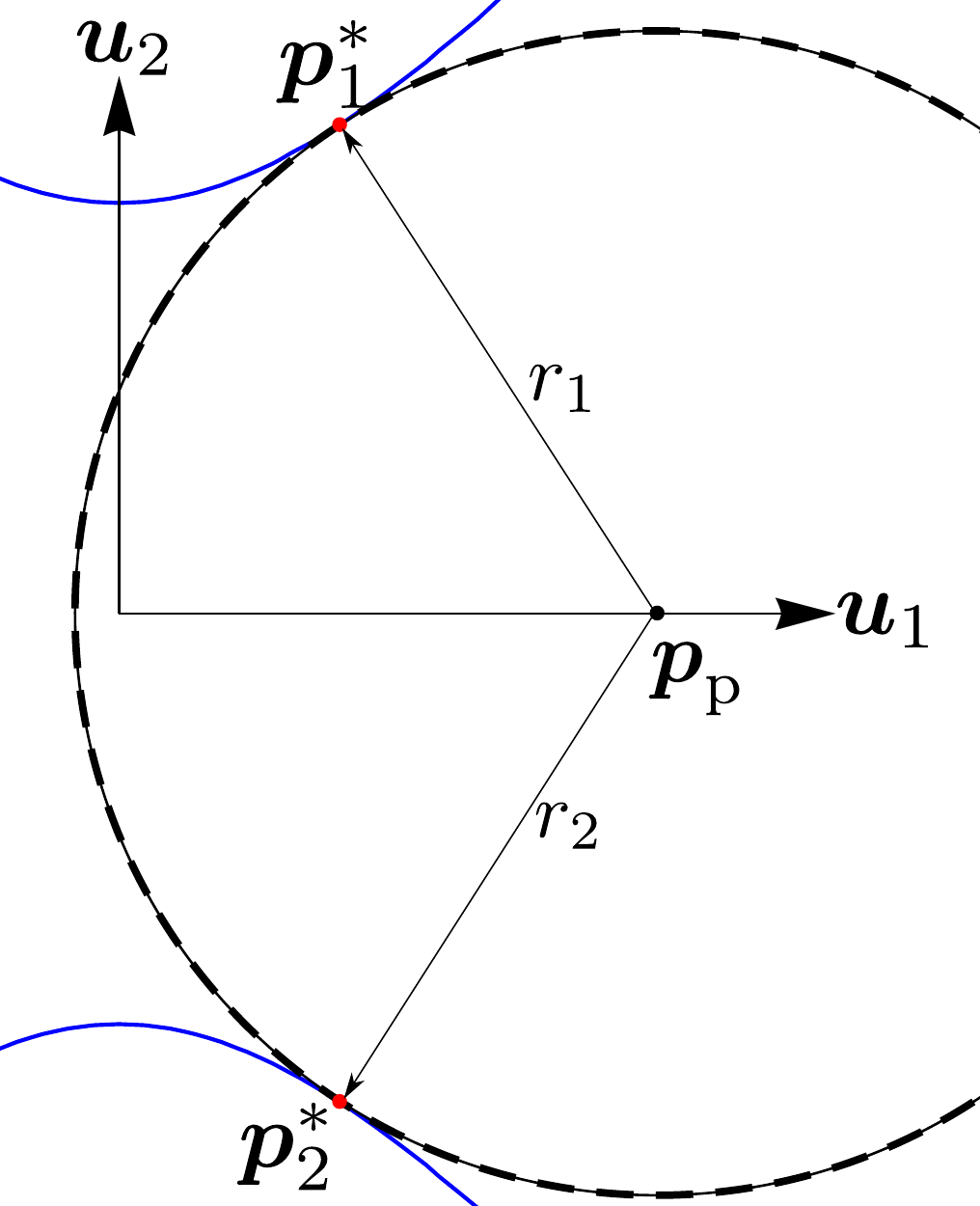}
				\subcaption{$\bpp$ lies outside the hyperbola; a case of two distinct solutions corresponding to Eq.~(\ref{eq:quadHypPtOutHyp}) leading to distances~$r_1,r_2$}
				\label{fg:ptOnXaxisOutHyp}
			\end{subfigure}
			\caption{Points~$\bp^*_j, j=1,2$, for the point~$\bpp$ on the~$\bu_1$-axis of an ellipse and a hyperbola}
			\label{fg:ptonXaxisEllipHyp}
		\end{figure}
		\item[(b.2)]\textbf{$\bpp$ lies outside a hyperbola}: In this case,~\mbox{$d_\text{N} \neq 0$}, hence, the~$u_2$-coordinate of~$\bpp$ is~$u^*_2 = t/e^2$. Substituting~$u^*_2$ in place of~$u_2$ of the equation of the line~$L_1$ mentioned in Eq.~(\ref{eq:stLineU2intercept}) results into the following equation:
		\begin{equation}
			\upone\left(\frac{t}{e^2}-t\right)+u_1t=0.
			\label{eq:lineHypOutXaxis}
		\end{equation}   
		Solving Eq.~(\ref{eq:lineHypOutXaxis}) for~$u_1$ results in the~$u_1$-coordinate of~$\bp^*$ as follows:
		\begin{equation}
			\bp^* = \frac{1}{e^2}\left[e_1 \upone ,t\right]^\top.
			\label{eq:tanPointHypXaxis}
		\end{equation}
		A quadratic equation in the variable~$t$ in its rational form is obtained after substituting~$\bp^*$ in Eq.~(\ref{eq:stEllipseEqEccen}) as:
		\begin{align}
			&\frac{e_1\left(t^2-e_2-e_1\upone^2\right)}{e^4} = 0,\quad \text{where}
			\label{eq:quadHypPtOutHyp}\\
			&e \neq 0,~\text{and} \nonumber\\
			&e_2=n^2e^4. \nonumber
		\end{align}
		Solving Eq.~(\ref{eq:quadHypPtOutHyp}) for~$t$, the solutions are obtained to be:
		\begin{equation}
			t = \pm \sqrt{e_2 + e_1 \upone^2}.
			\label{eq:rootsQuadHyp}
		\end{equation}
		Equation~(\ref{eq:quadHypPtOutHyp}) has two distinct real roots because the discriminant of the equation is always positive. The points~$\bp^*_j$, as shown in Fig.~\ref{fg:ptOnXaxisOutHyp}, is computed for the corresponding distinct real roots,~$t_j$. The minimum among the distances from~$\bpp$ to~$\bp^*_1$ and from~$\bpp$ to~$\bp^*_2$ is the proximity~\rmin.  
	\end{enumerate}  
	\item \textbf{$\bpp$ does not lie on the axes}: In this case,~\mbox{$d_\text{N} \neq 0$}, hence, substituting~$u^*_2 = t/e^2$ in place of~$u_2$ in Eq.~(\ref{eq:stLineU2intercept}) results in the~$u_1$-coordinate of $\bp^*$. The point $\bp^*$ is obtained as:
	\begin{equation}
	\bp^* = \frac{1}{e^2}\left[\frac{\upone t e_1}{ t-\uptwo},t\right]^\top.
	\label{eq:ptoFtanEllipHyp}
	\end{equation}
	Substituting~$\bp^*$ in Eq.~(\ref{eq:stEllipseEqEccen}) results an equation in the following rational form:
	\begin{align}
	\frac{e_1 (t^4 - 2 \uptwo  t^3 - (e_2 + e_1 \upone^2 - \uptwo^2) t^2  + 2 e_2 \uptwo t - e_2 \uptwo^2 )}{e^4 (t-\uptwo)^2} = 0, \quad \text{where}~e \neq 0.
	\label{eq:univaTEllipHyp}
	\end{align}
	The numerator of Eq.~(\ref{eq:univaTEllipHyp}) is reduced to a quartic equation in $t$ as follows: 
	\begin{equation}
	t^4 - 2 \uptwo t^3 + (\uptwo^2- e_2 - \upone^2e_1) t^2 + 2 e_2 \uptwo t - e_2 \uptwo^2  = 0.
	\label{eq:quarticTellipHyp}
	\end{equation} 
	The characterisation of the real roots\footnote{The expressions of the roots of a quartic equation are mentioned in Appendix~\ref{sc:rootquartic}.} is done by reducing Eq.~(\ref{eq:quarticTellipHyp}) to its {\em depressed quartic} by dropping the cubic term after substituting $t = \varLambda + \frac{\uptwo}{2}$ in the equation as follows:
	\begin{align}
	&\varLambda^4 +  c_0 \varLambda^2 + c_1 \varLambda + c_2 = 0,\quad \text{where}
	\label{eq:depQuartic}\\
	&c_0 = -(e_1 \upone^2 + e_2 + \frac{\uptwo^2}{2}),\label{eq:coeffCubeOfQuartic}\\
	&c_1 = (e_2 - e_1 \upone^2) \uptwo, \text{and}\nonumber \\
	&c_2 = \frac{\uptwo^2}{16} (\uptwo^2 - 4 e_2 - 4 e_1 \upone^2). \nonumber
	\end{align}
	The parameters~\cite{Arnon1988} required to characterise the real roots of Eq.~(\ref{eq:depQuartic}) along with $c_0$ are:
	\begin{align}
	&\delta_1 = -16 e_2^2 e_1 \upone^2 \uptwo^2 \left(e_2^2 + 27  e_1 \upone^2 \uptwo^2 + e_3\left(e_2\left(\sqrt{3} + \frac{e_3}{e_2}\right)^2 + (3-2\sqrt{3})e_3\right)\right), \label{eq:quarticRootCharCond1}\\
	&\delta_2 =  2 (e_2^3 + e_2^2 (3 e_1 \upone^2 - 2 \uptwo^2) + e_3^2 e_1 \upone^2  + e_2 ((\sqrt{3}e_1\upone^2-\uptwo^2)^2 + e_1 \upone^2 \uptwo^2 (14+2\sqrt{3}))),\label{eq:quarticRootCharCond2} \\
	&\text{where}~e_3=e_1\upone^2-\uptwo^2. \nonumber
	\end{align}
	 From Eq.~(\ref{eq:coeffCubeOfQuartic}), the value of~$c_0$ is always less than zero. As the point~$\bpp$ does not lie on the axes, the value of $\delta_2$ in Eq.~(\ref{eq:quarticRootCharCond2}) can not be zero. Hence, the classification of the real roots~\cite{Arnon1988} are based on the value of~$\delta_1$, which are enumerated below.
	\begin{enumerate}
		\item[(c.1)] $\delta_1 > 0$: In this case, there are four distinct real solutions of Eq.~(\ref{eq:quarticTellipHyp}). Substituting the roots $t_j,j=1,\dots,4$, in Eq.~(\ref{eq:ptoFtanEllipHyp}) results in four intersection points~$\bp^*_j$ on the ellipse, and hyperbola, as shown in Fig.~\ref{fg:ellipse4Pt}.
		\begin{figure}[!t]
			\centering
			\begin{subfigure}{0.45\textwidth}
				\centering
				\includegraphics[scale=0.25]{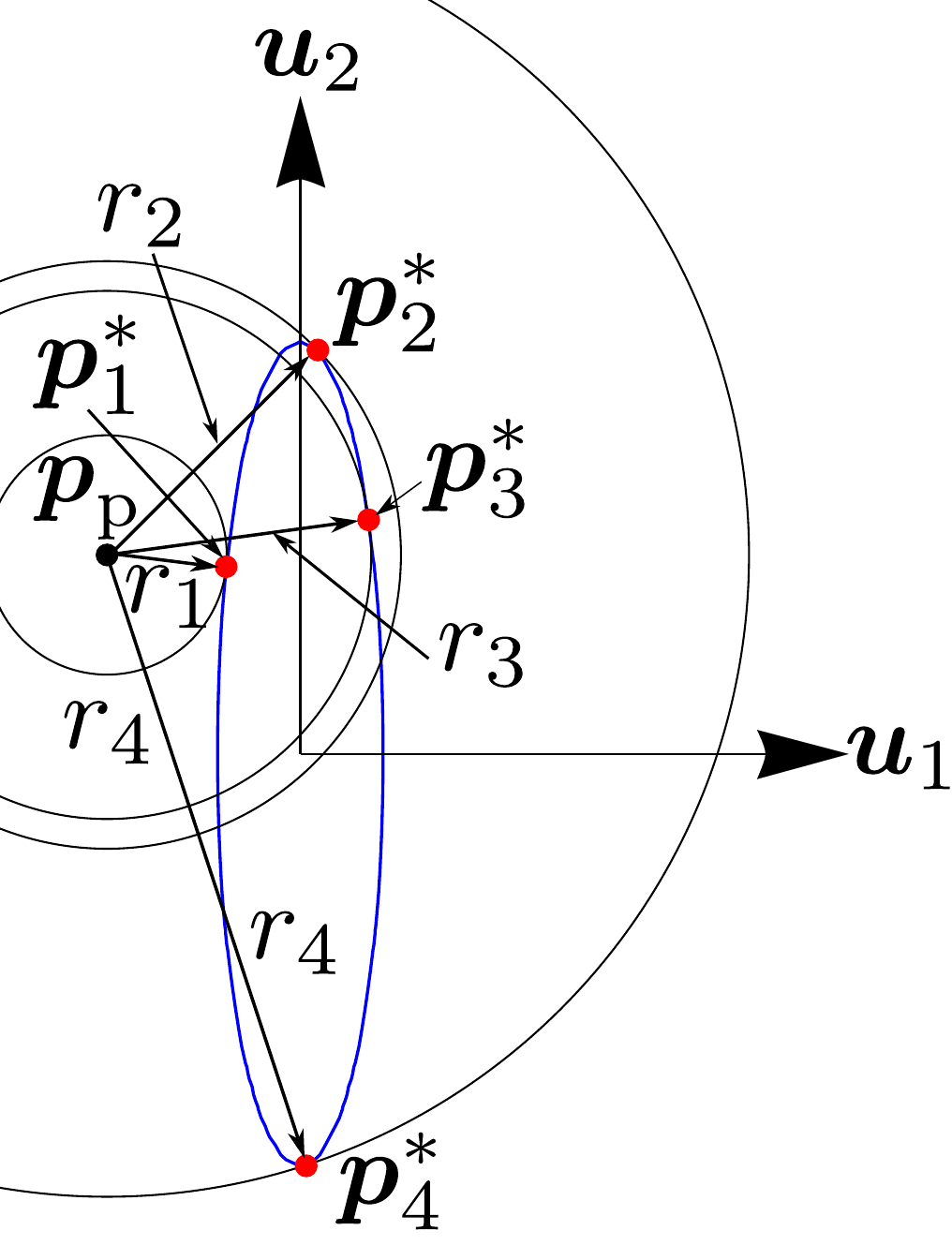}
				\subcaption{\bpp is outside the ellipse; a case of four distinct real solutions leading to distances~$r_j,j=1,\dots,4$}
			\end{subfigure}
			\hfill
			\begin{subfigure}{0.45\textwidth}
				%\vspace{0.5 cm}
				\centering
				\includegraphics[scale=0.3]{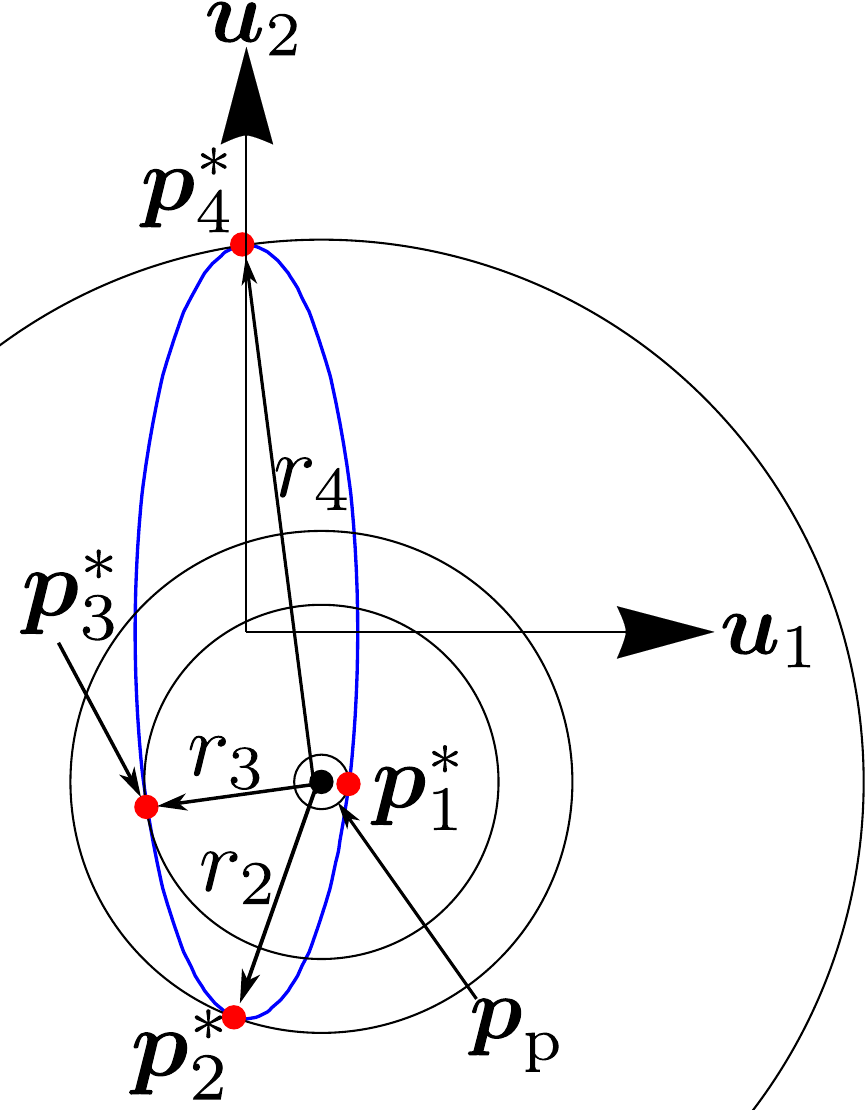}
				\subcaption{\bpp is inside the ellipse; a case of four distinct real solutions leading to distances~$r_j,j=1,\dots,4$}
			\end{subfigure}
			\hfill
			\begin{subfigure}{0.45\textwidth}
				%\vspace{0.5 cm}
				\centering
				\includegraphics[scale=0.6]{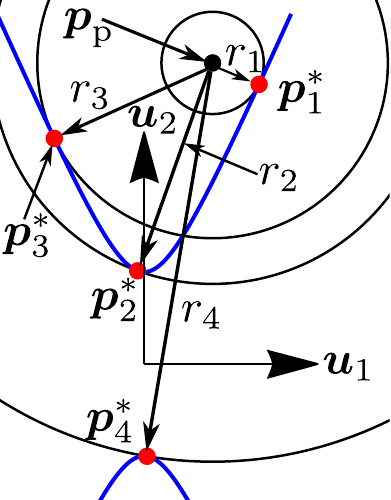}
				\subcaption{\bpp is at inside of the hyperbola; another case of four distinct real solutions leading to distances~$r_j,j=1,\dots,4$}
			\end{subfigure}
			\hfill
			\begin{subfigure}{0.45\textwidth}
				%\vspace{0.5 cm}
				\centering
				\includegraphics[scale=0.4]{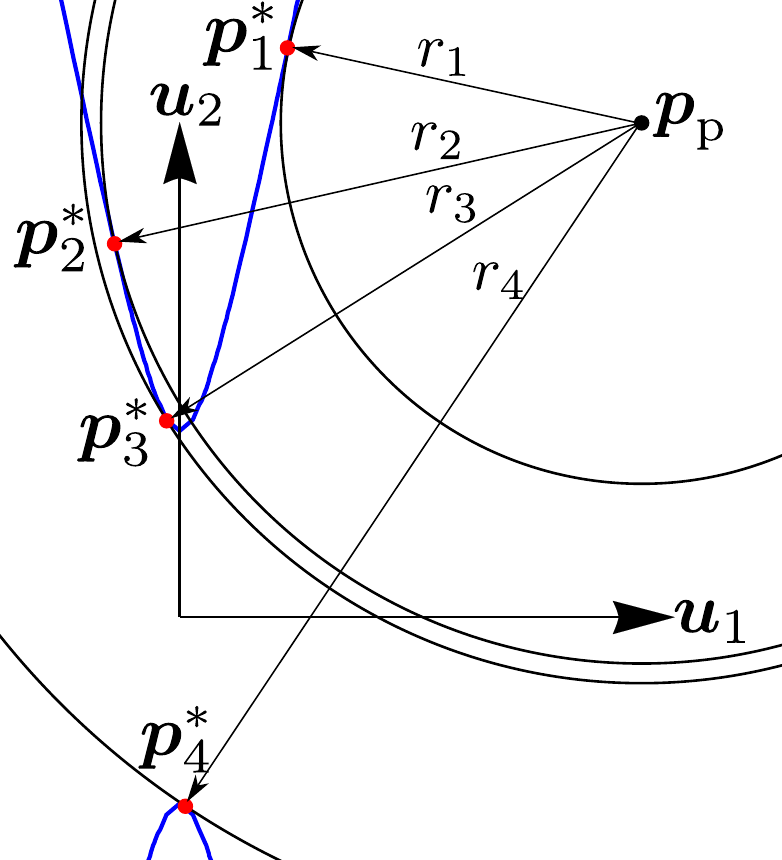}
				\subcaption{\bpp is at outside of the hyperbola; another case of four distinct real solutions leading to distances~$r_j,j=1,\dots,4$}
			\end{subfigure}
			\caption{Representation of point, $\bp^*_j, j=1,\dots,4$, corresponding to four distinct real roots of Eq.~(\ref{eq:quarticTellipHyp}) for an ellipse and a hyperbola, when~\bpp does not lie on both the axes}
			\label{fg:ellipse4Pt}
		\end{figure}
		\begin{figure}[!t]
			\centering
			\begin{subfigure}{0.45\textwidth}
				%\vspace{2.5 cm}
				\centering
				\includegraphics[scale=0.4]{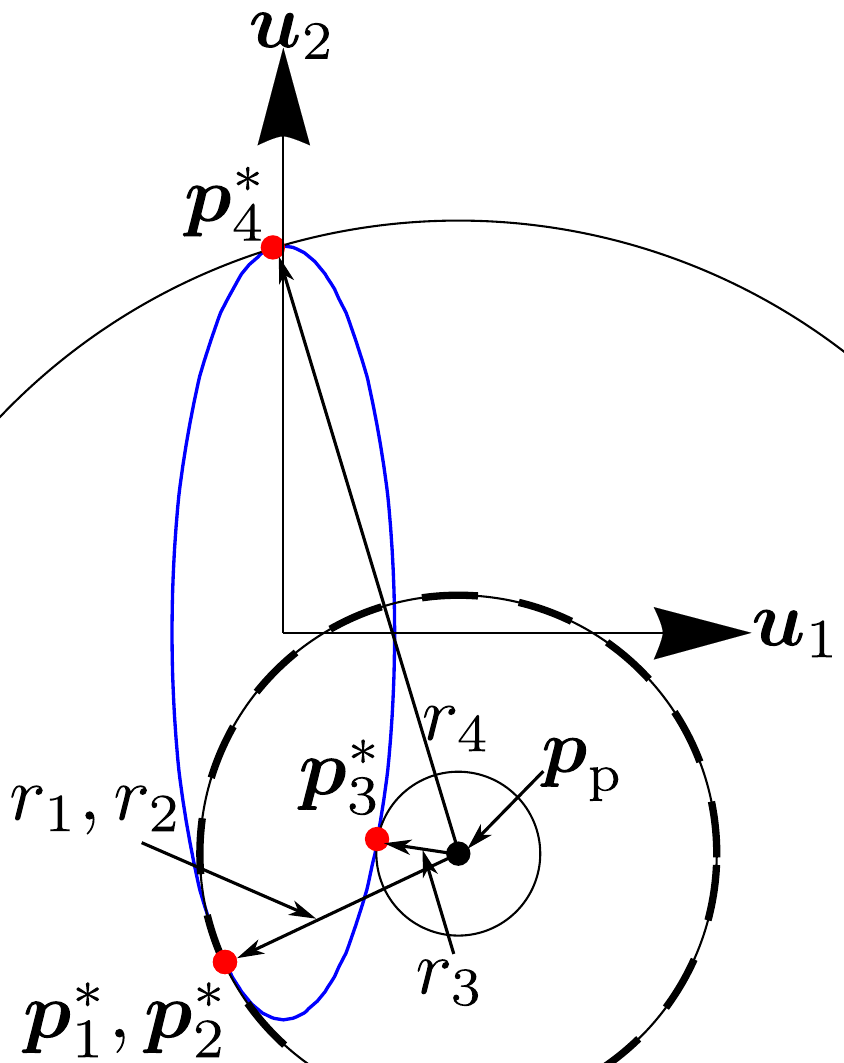}
				\subcaption{\bpp lies outside the ellipse; a case of two distinct and two repeated real solutions leading to two distinct distances~$r_3,r_4$ and two same distances~$r_1=r_2$, respectively}
			\end{subfigure}
			\hfill
			\begin{subfigure}{0.45\textwidth}
				\centering
				\includegraphics[scale=0.52]{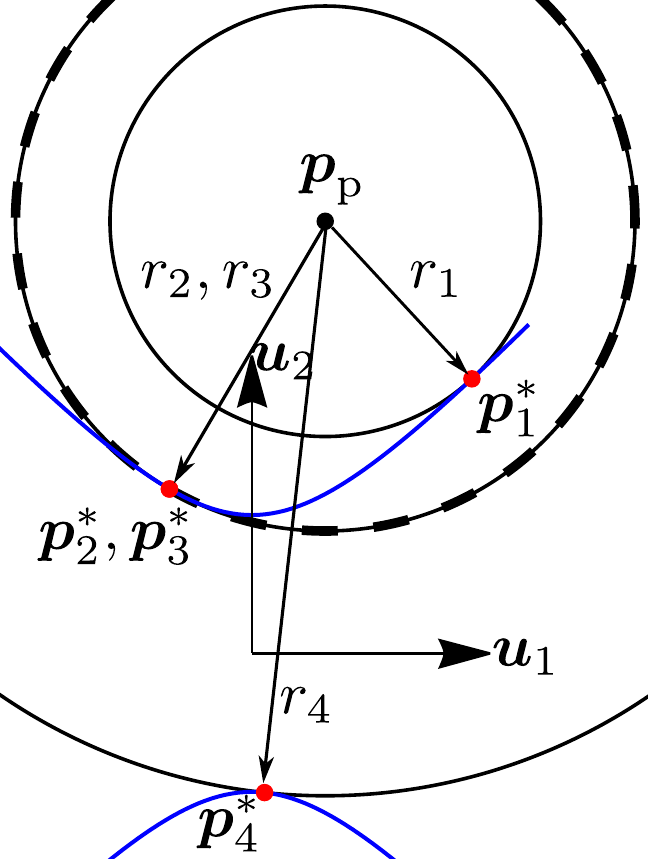}
				\subcaption{$\bpp$ lies outside of hyperbola; a case of two distinct and two repeated real solutions leading to two distinct distances~$r_1,r_4$ and two same distances~$r_2=r_3$, respectively}
			\end{subfigure}
			\caption{Representation of point, $\bp^*_j, j=1,\dots,4$, corresponding to the two repeated real roots and two distinct real roots of Eq.~(\ref{eq:quarticTellipHyp}) for an ellipse and a hyperbola, when $\bpp$ is not on the axes}
			\label{fg:twoReaptTwoDistReRoots}
		\end{figure}
		\item[(c.2)] $\delta_1=0$: In this case, Eq.~(\ref{eq:quarticTellipHyp}) has four real roots, out of which two of the roots are distinct and two roots are repeated. The intersection points $\bp^*_j, j=1,\dots,4$, are computed from Eq.~(\ref{eq:ptoFtanEllipHyp}), as shown in Fig.~\ref{fg:twoReaptTwoDistReRoots}.  
		
		\item[(c.3)] $\delta_1<0$: In this case, Eq.~(\ref{eq:quarticTellipHyp}) has two distinct real and two complex roots. The points~$\bp^*_j,j=1,2$, corresponding to the distinct real roots are shown in Fig.~\ref{fg:twoDistRootsQuartEq}.  
		\begin{figure}[!t]
		\centering
			\begin{subfigure}{0.45\textwidth}
				\centering	
				\includegraphics[width=0.5\textwidth]{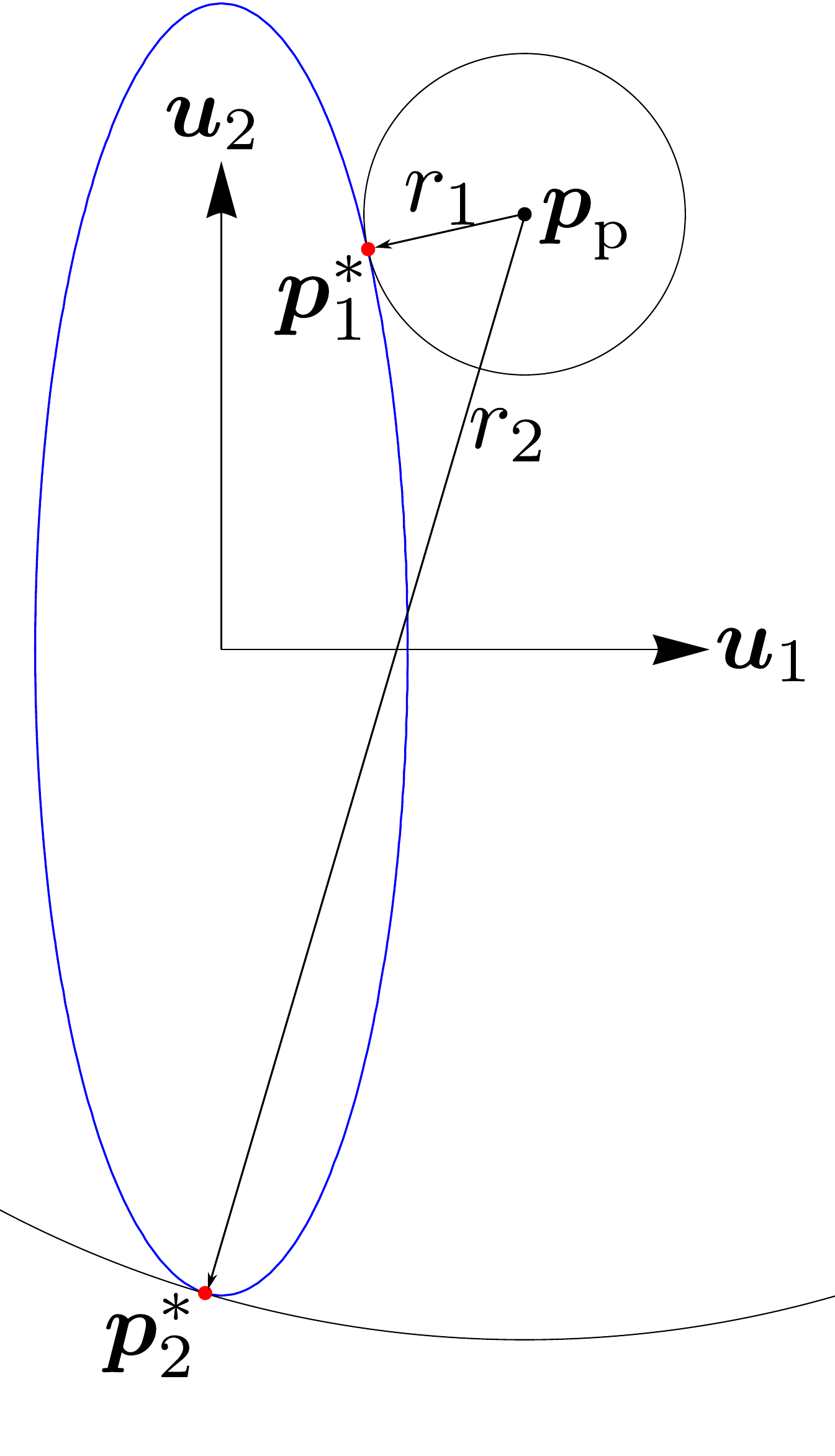}
				\subcaption{$\bpp$ lies outside the ellipse; a case of two distinct real solutions leading to two distances~$r_1,r_2$}
			\end{subfigure}
			\hfill
			\begin{subfigure}{0.45\textwidth}
				%\vspace{0.5 cm}
				\centering
				\includegraphics[width=0.7\textwidth]{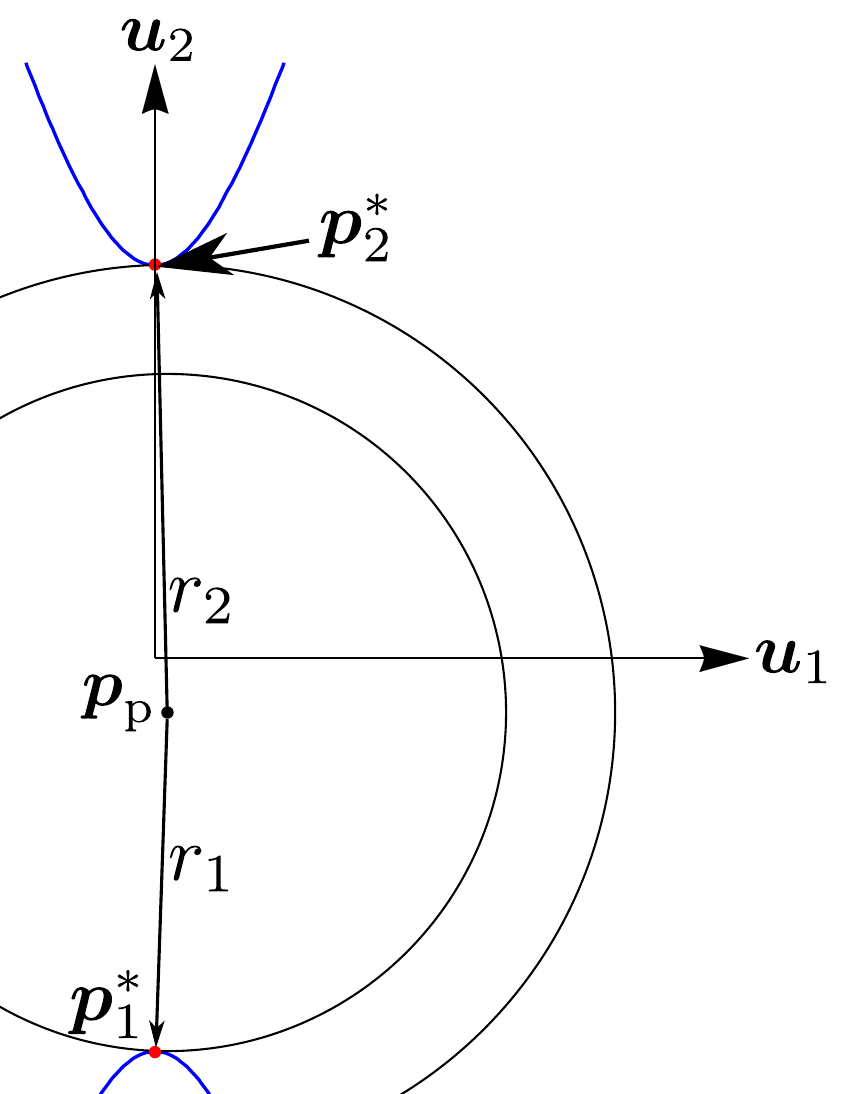}
				\subcaption{$\bpp$ lies outside the hyperbola; a case of two distinct real solutions leading to two distances~$r_1,r_2$}
			\end{subfigure}
			\hfill
			\begin{subfigure}{0.45\textwidth}
				%\vspace{1.8 cm}
				\centering
				\includegraphics[width=0.7\textwidth]{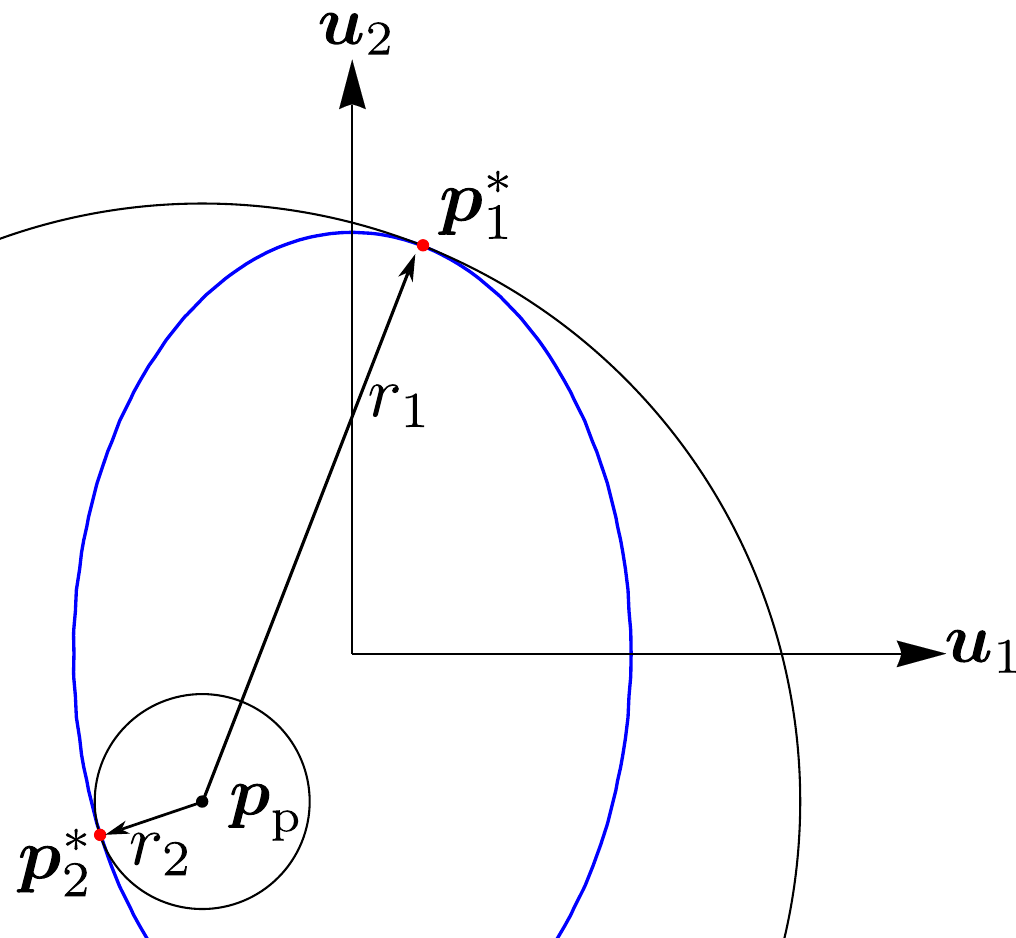}
				\subcaption{$\bpp$ lies inside the ellipse; a case of two distinct real solutions leading to two distances~$r_1,r_2$}
			\end{subfigure}
			\hfill
			\begin{subfigure}{0.45\textwidth}
				%\vspace{0.8 cm}
				\centering
				\includegraphics[width=0.6\textwidth]{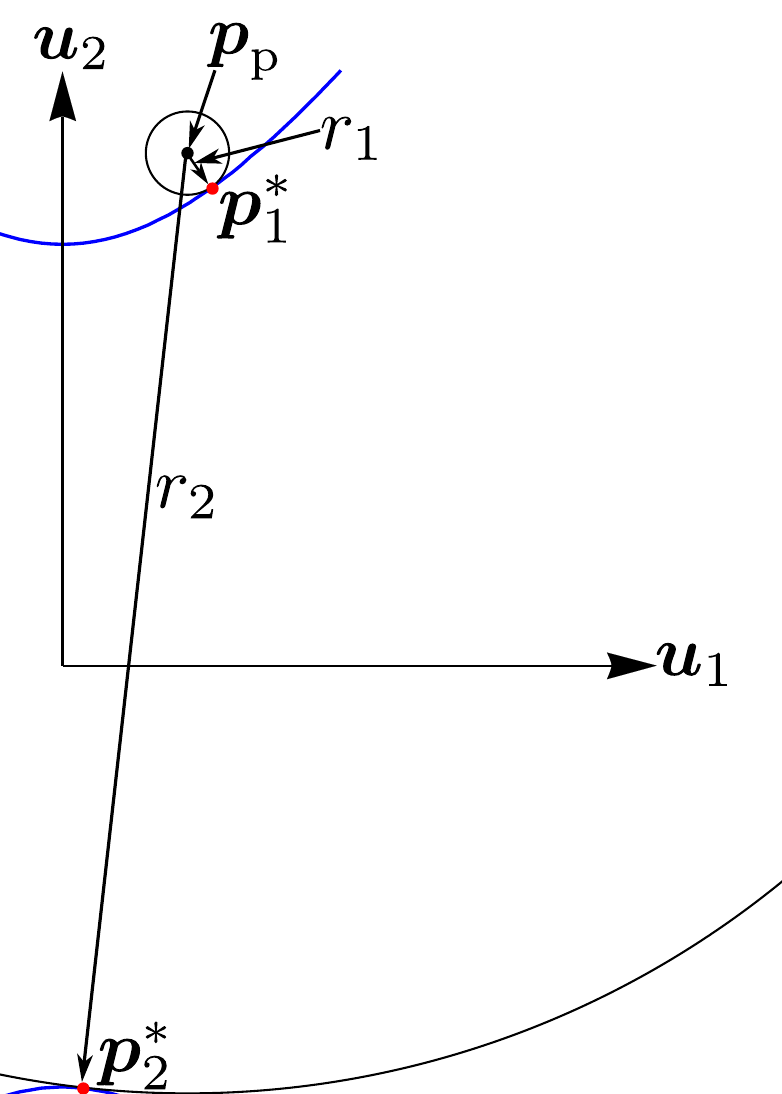}
				\subcaption{$\bpp$ lies outside the hyperbola; a case of two distinct real solutions leading to two distances~$r_1,r_2$}
			\end{subfigure}
			\caption{Points, $\bp^*_j,j=1,2$, corresponding to the two real distinct roots of Eq.~(\ref{eq:quarticTellipHyp}) for an ellipse and a hyperbola, when~\bpp is not on the axes}
			\label{fg:twoDistRootsQuartEq}
		\end{figure}
	\end{enumerate}
	After computing $\bp^*_j$, the distance between the point $\bpp$ to an ellipse or a hyperbola using Eq.~(\ref{eq:shrtDistParabola}) with $j = 1,\dots,4$ or $j = 1,2$ depending upon the number of real roots. The minimum among $r_j$ is the proximity (\rmin) of $\bpp$ to the curve, as shown in Figs.~\ref{fg:shortDistanceEllipse},~\ref{fg:shortDistancehyperbola}, which is the same as the distance from the point $\bp_0$ to the spheroid or hyperboloid of one/two sheets, as shown in Figs.~\ref{fg:proSProidTan}-\ref{fg:HypTwoTan}.
\end{enumerate}

% % % % % % % % % % % % % % % % % % % % % % % % % % % % % % % % % % % % % % %
\subsubsection{Proximity of a point to a pair of intersecting lines $(e=\infty)$}
\begin{figure}[!t]
	\centering
	\begin{minipage}{0.45\textwidth}
		\centering
		\includegraphics[scale=0.15]{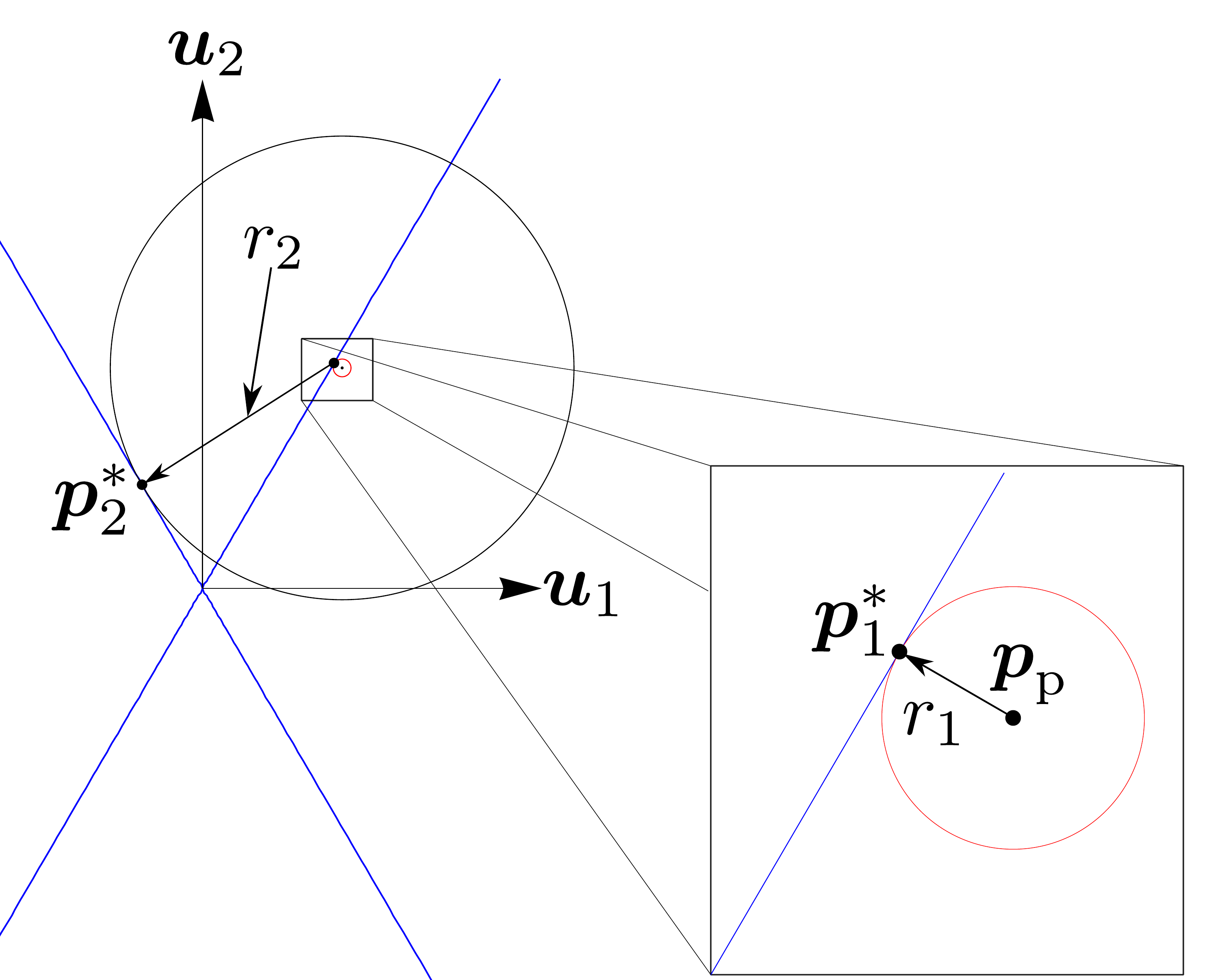}
		\caption{Proximity between the point $\bp_\text{p}$ and the pair of intersecting straight lines; refer to Table~\ref{tb:numresultAQ} for the distance~$r_1,r_2$.}
		\label{fg:shortpairIntLines}
	\end{minipage}
	\hfill
	\begin{minipage}{0.45\textwidth}
		\centering
		\includegraphics[scale=0.5]{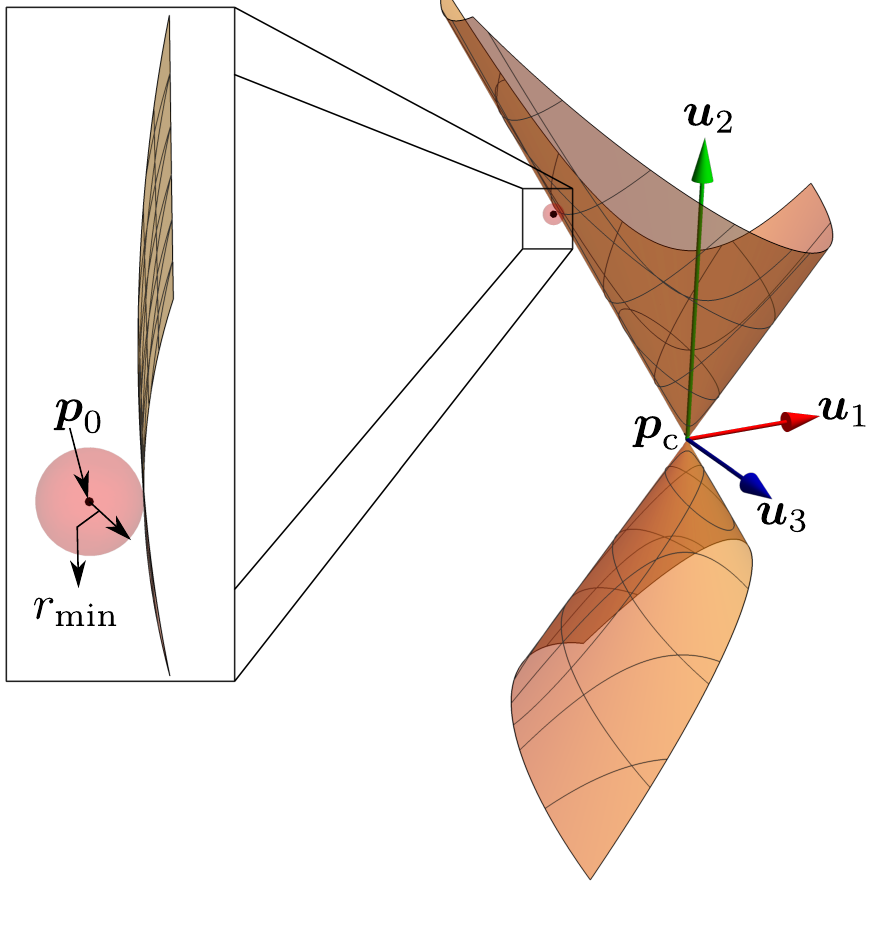}
		\caption{Proximity between the point $\bp_0$ and the cone; the radius of the sphere centred at~$\bp_0$ of radius~\rmin (see Table~\ref{tb:numresultAQ}) is tangent to the cone.}
		\label{fg:shortCone}
	\end{minipage}
\end{figure}
The proximity of a point to a straight line in a plane is the projection of the point onto the line. From Eq.~(\ref{eq:stdFrmIntLines}), two equations for the lines are
$\sqrt{m_1} u_1-u_2=0$, and $\sqrt{m_1} u_1+u_2=0$, where~$m_1=\left|\lambda_1/\lambda_3\right|$ and the shortest distance from the point to the lines are as follows:
\begin{align}
	&r_1 = \frac{\left|\sqrt{m_1}\upone-\uptwo\right|}{\sqrt{1+m_1}},~\text{and}\\
	&r_2 = \frac{\left|\sqrt{m_1}\upone+\uptwo\right|}{\sqrt{1+m_1}}.
	\label{eq:shortDistLines}
\end{align}
The minimum between~$r_1$, and~$r_2$ is the proximity of~$\bpp$ to the pair of the intersecting lines, as shown Fig.~\ref{fg:shortpairIntLines}, which is the same as the distance between the point $\bp_0$ and the cone, as shown in Fig.~\ref{fg:shortCone}.
% % % % % % % % % % % % % % % % % % % % % % % % % % % % % % % % %
\subsubsection{Proximity of a point to a pair of non-coincident vertical parallel lines $(e=\infty)$}
\begin{figure}[!t]
	\begin{minipage}{0.45\textwidth}
		\centering
		\includegraphics[scale=0.35]{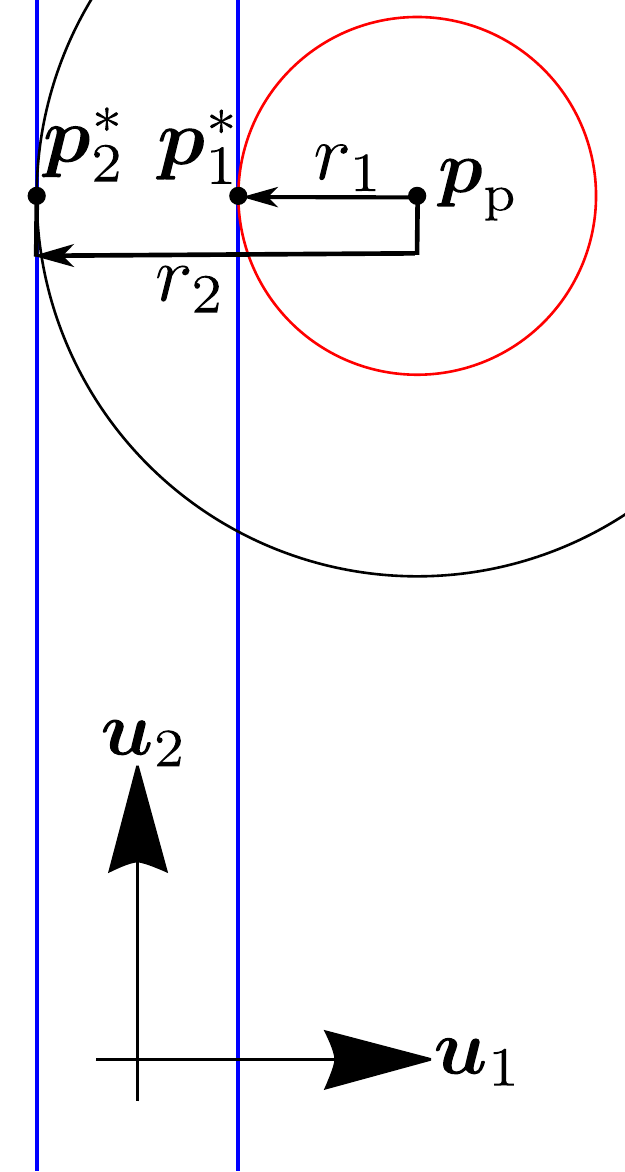}
		\caption{Proximity of~$\bpp$ to the pair of non-coincident parallel lines; refer to Table~\ref{tb:numresultAQ} for the distance~$r_1,r_2$.}
		\label{fg:proximitPtparallelLines}
	\end{minipage}
	\hfill
	\begin{minipage}{0.45\textwidth}
		\centering
		\includegraphics[scale=0.45]{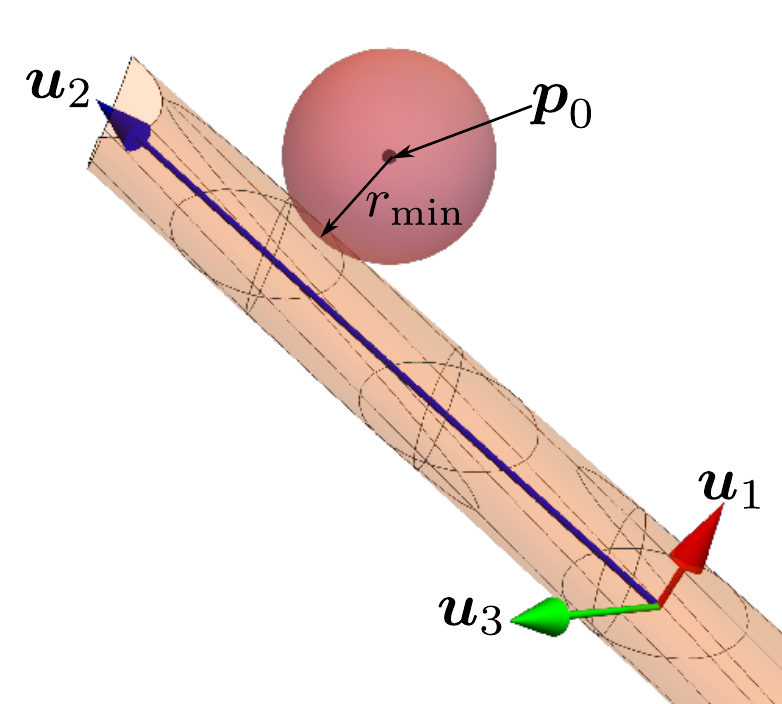}
		\caption{Proximity of~$\bp_0$ to cylinder; the radius of the sphere centred at~$\bp_0$ of radius~\rmin (see Table~\ref{tb:numresultAQ}) is tangent to the cylinder.}
		\label{fg:proximitPtCylinder}
	\end{minipage}
\end{figure}
The equations of a pair of non-coincident vertical parallel lines are as follows:~$u_1-m = 0$, and $u_1 + m = 0$. The proximity of~$\bpp$ to the pair of parallel lines is the perpendicular distance between point and the lines. The foot of perpendicular on to the parallel lines from~$\bpp$,~$\bp^*_1$ and~$\bp^*_2$, as shown in Fig.~\ref{fg:proximitPtparallelLines}, are defined as~$[m , \uptwo ]^\top$ and~$[-m , \uptwo ]^\top$, respectively. Hence, the distances between these two points and $\bpp$ are defined as follows:
\begin{align}
	&r_1 = |\upone - m|,~\text{and} \\
	&r_2 = |\upone + m|.  
\end{align}
The minimum value among~$r_1$ and~$r_2$ is the proximity~\rmin of the point~$\bp_0$ to the cylinder as shown in Fig.~\ref{fg:proximitPtCylinder}.  
%%%%%%%%%%%%%%%%%%%%%%%%%
\section{Numerical example}
\label{sc:examples}
\begin{figure}[!t]
	\begin{subfigure}{0.45\textwidth}
		\centering
		\includegraphics[scale=0.25]{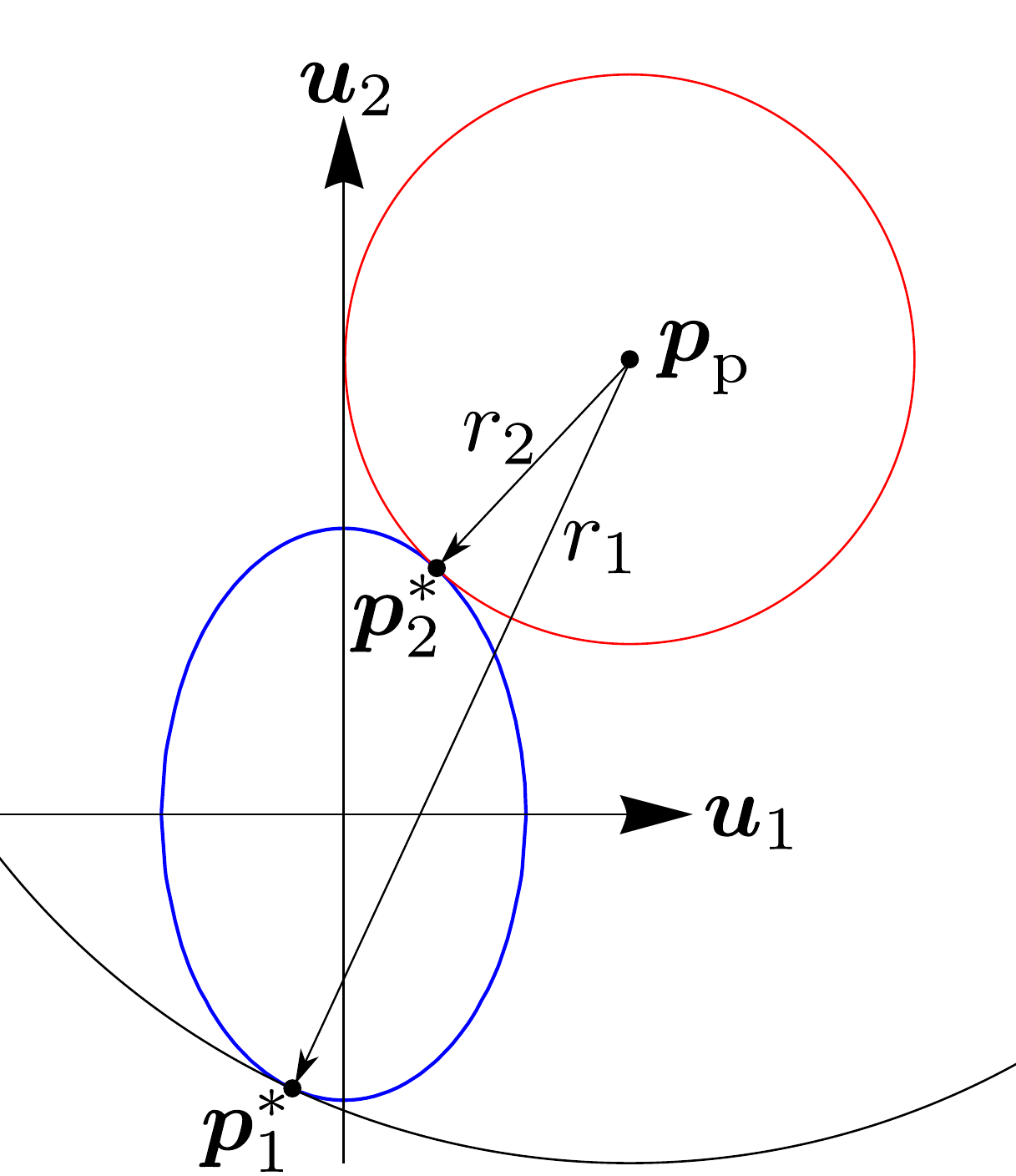}
		\subcaption{Proximity of~$\bpp$ to an ellipse associated with the prolate spheroid}
		\label{fg:ptOutVertEllipse}
	\end{subfigure}
	\hfill
	\begin{subfigure}{0.45\textwidth}
		\centering
		\includegraphics[scale=0.2]{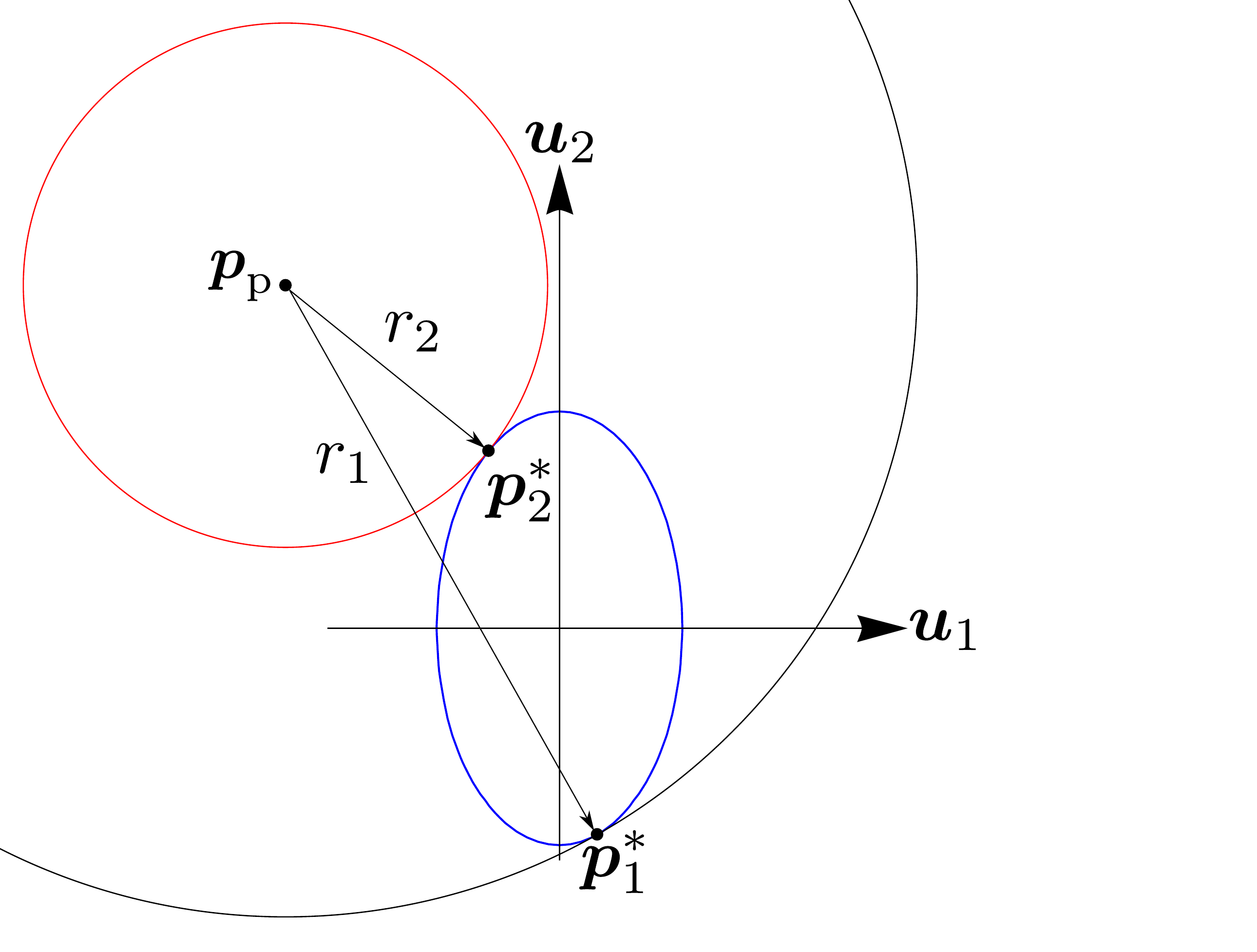}
		\subcaption{Proximity of~$\bpp$ to an ellipse associated with the oblate spheroid; showing the rotated ellipse by an angle~$\pi/2$ about the out of the plane axis in the CW manner}
		\label{fg:ptOutHorinEllipse}
	\end{subfigure}
	\hfill
	\begin{subfigure}{0.45\textwidth}
		\centering
		\includegraphics[scale=0.15]{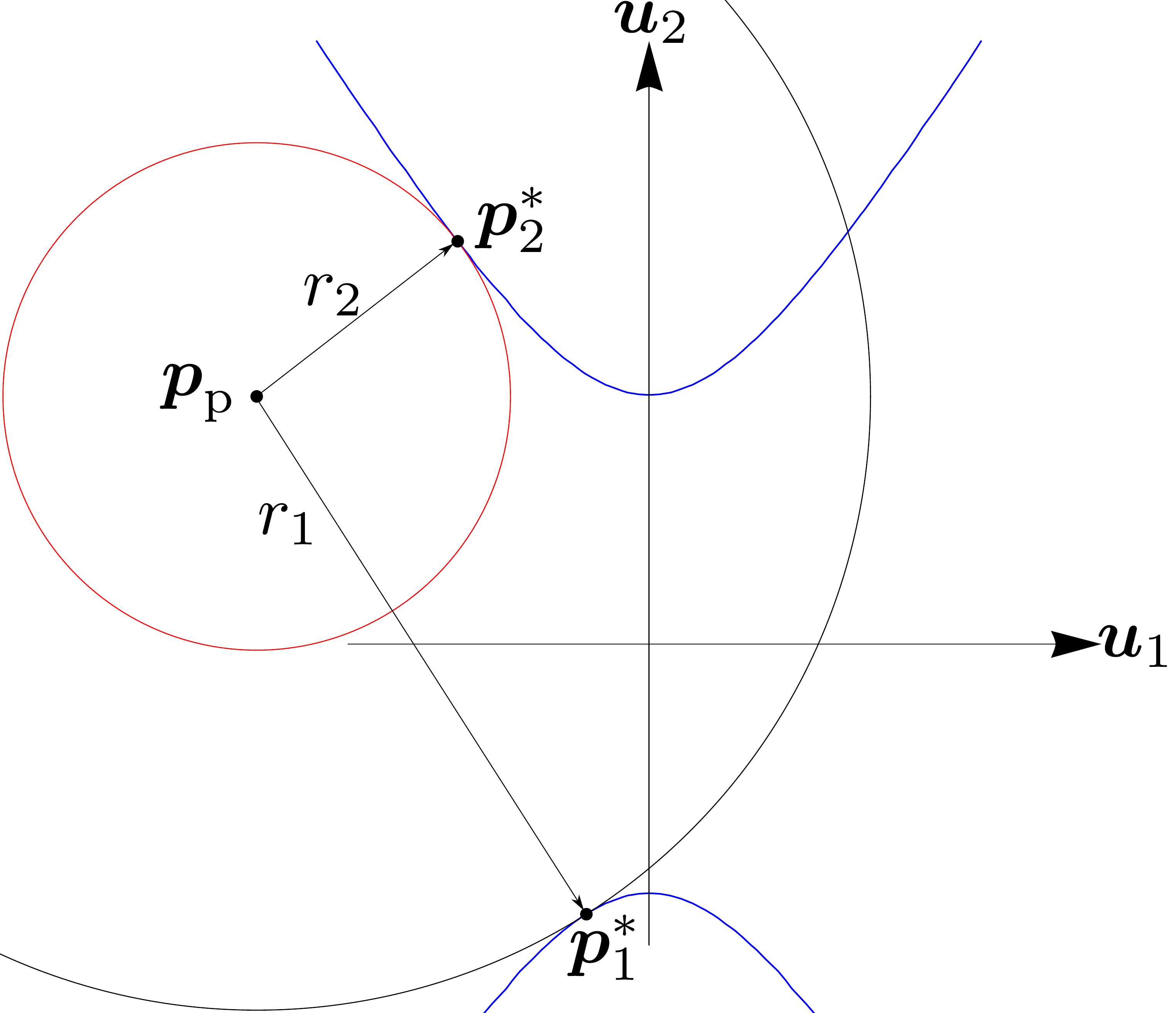}
		\subcaption{Proximity of~$\bpp$ to a hyperbola associated with the hyperboloid of one sheet}
		\label{fg:ptOutParabU1int}
	\end{subfigure}
	\hfill
	\begin{subfigure}{0.45\textwidth}
		\centering
		\includegraphics[scale=0.35]{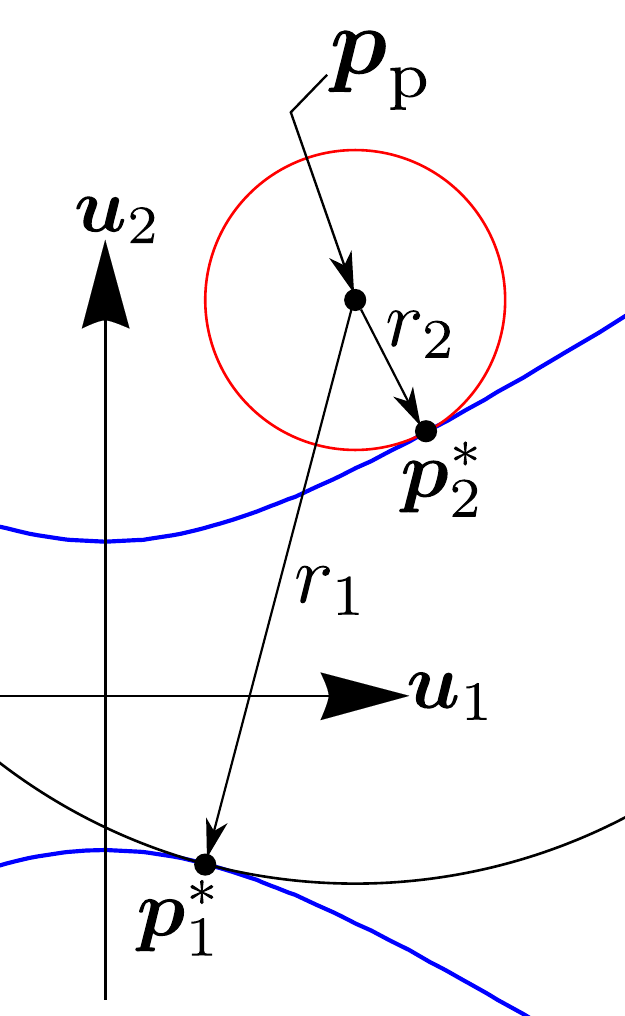}
		\subcaption{Proximity of~$\bpp$ to a hyperbola associated with hyperboloid of two sheets; showing the rotated hyperbola by an angle~$\pi/2$ about the out of the plane axis in the CW manner}
		\label{fg:ptInParabU2int}
	\end{subfigure}
	\caption{Proximity of~$\bpp$ to the ellipses and hyperbol\ae, which is equivalent to the circle tangent to these conic sections; refer to Table~\ref{tb:numresultAQ} for the details of the distances~$r_1,r_2$.}
	\label{fg:shortDistEllipHyp}
\end{figure}
\begin{figure}[!t]
	\centering
	\begin{subfigure}{0.45\textwidth}
		\centering
		\includegraphics[scale=0.6]{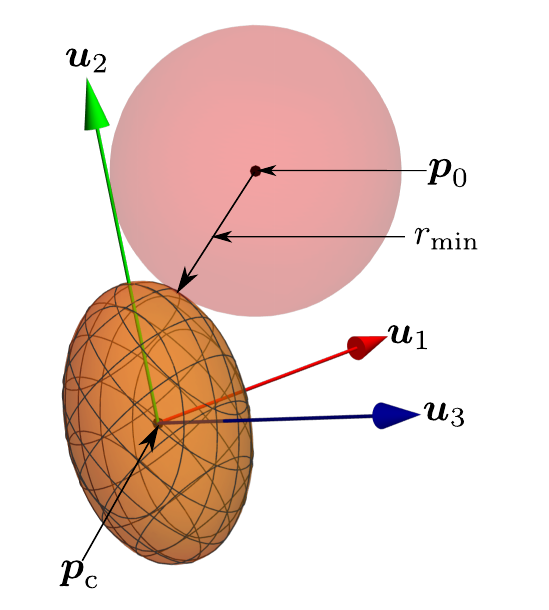}
		\caption{Proximity of~$\bp_0$ to prolate spheroid}
		\label{fg:proSProidTan}
	\end{subfigure}
	\hfill
	\begin{subfigure}{0.45\textwidth}
	%	\vspace{1.3 cm}
		\centering
		\includegraphics[scale=0.4]{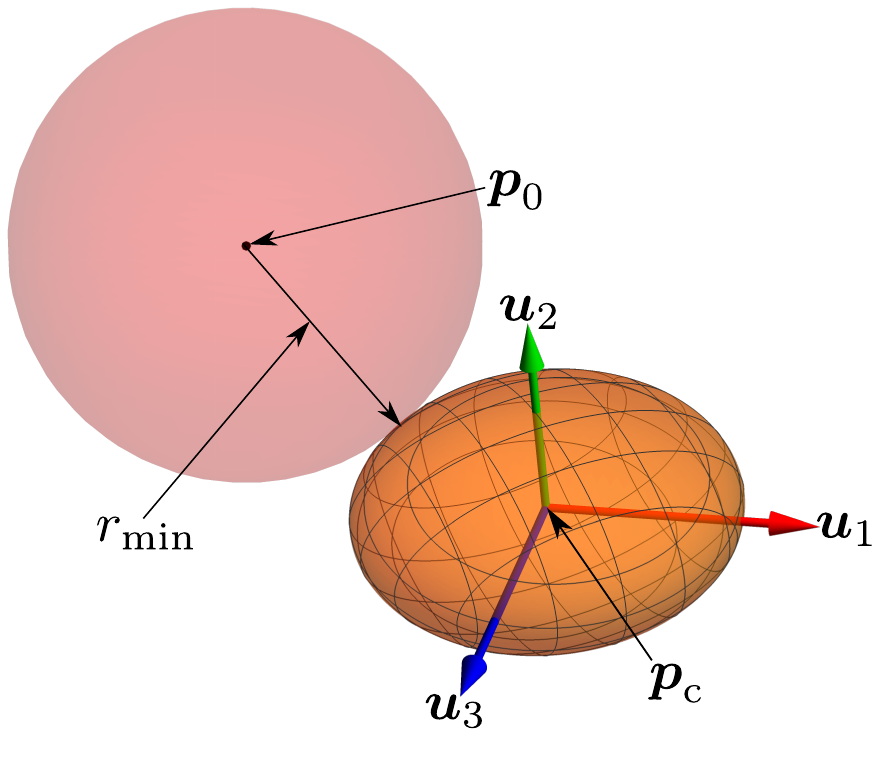}
		\caption{Proximity of~$\bp_0$ to oblate spheroid}
		\label{fg:OblSProidTan}
	\end{subfigure}
	\hfill
	\begin{subfigure}{0.45\textwidth}
		%\vspace{1.5 cm}
		\centering
		\includegraphics[scale=0.6]{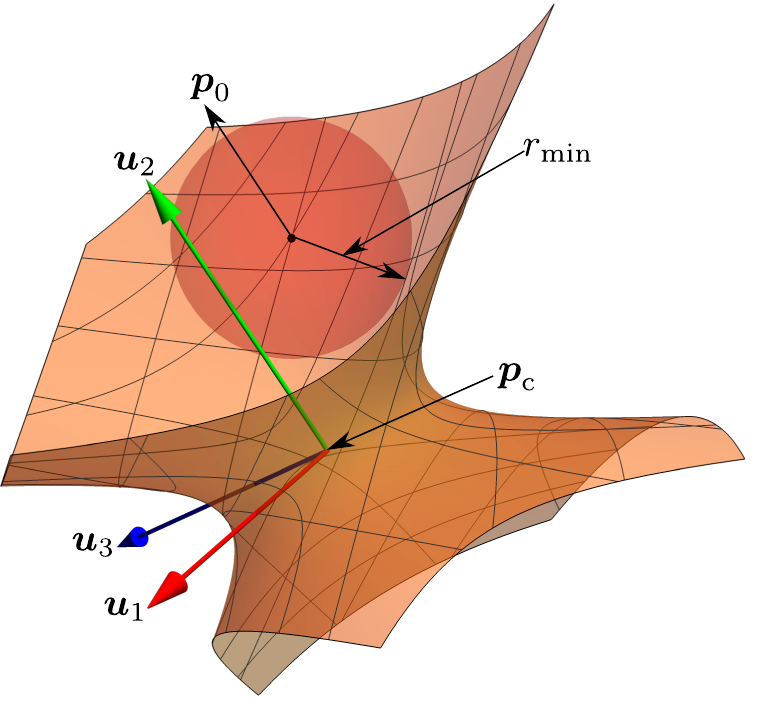}
		\caption{Proximity of~$\bp_0$ to hyperboloid of one sheet}
		\label{fg:HypOneTan}
	\end{subfigure}
	\hspace{0.5 cm}
	\begin{subfigure}{0.45\textwidth}
		\centering
		\includegraphics[scale=0.6]{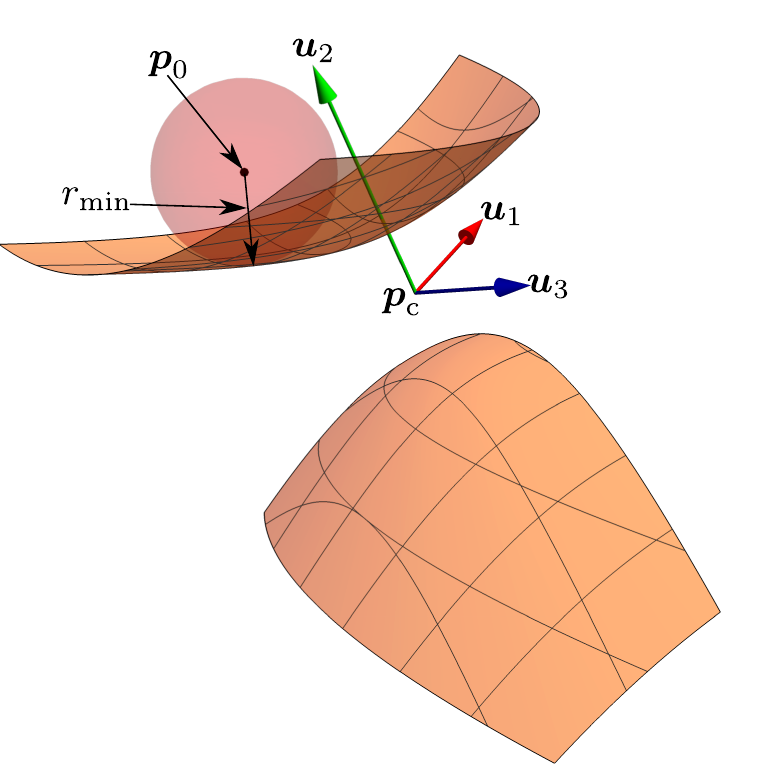}
		\caption{Proximity of~$\bp_0$ to hyperboloid of two sheets}
		\label{fg:HypTwoTan}
	\end{subfigure}
	\caption{Proximity of~$\bp_0$ to the hyperboloid and spheroids in the form of a sphere with radius \rmin (see Table~\ref{tb:numresultAQ}) that is tangent to the hyperboloid and spheroid.}
	\label{fg:shortSphHyp}
\end{figure}

The results obtained from the formulation for identifying an AQ and computing the proximity of~$\bp_0$ to the AQ is mentioned in this section. All the computations are performed using the programming language~\verb|C| on a single thread of an AMD\textsuperscript{\textregistered} Ryzen 9 7950x processor running at~$4.50$ GHz of frequency. The \verb|g++| compiler of version~$9.4.0$ with \verb|-O2| flag option in the \verb|Linux| environment is used for implementation. The point~$\bp_0$ is considered to be the same for all the cases, i.e.,~$\bp_0 = [-0.7230, 0.8655, 0.5549]^\top$, but for the paraboloid and cylinder,~$\bp_0$ is chosen to be~$[6.1658, 1.1438, -0.6710]^\top$. 
The proximity of~$\bp_0$ to~$\surf$ is represented as a sphere centred at~$\bp_0$ with radius~\rmin, i.e., the sphere is tangent to the~$\surf$, as shown in Figs.~\ref{fg:sphrShortDist},~\ref{fg:parabolaShortDist},~\ref{fg:shortCone},~\ref{fg:proximitPtCylinder}, and~\ref{fg:shortSphHyp}.
% % % % % % % % % % % % % % % % % % % % % % % % % % % % % %
\begin{table}[!h]
	\centering
	\caption{Values of the invariants~$I_4, J_3$,~$\Delta$, and~$\lambda_j,j=1,2,3$, for the AQs.}
	\label{tb:classifAQ}
	\begin{tabular}{|c|r|r|r|r|r|l|}
		\hline
		\makecell{Sl.  no.} & \makecell{$I_4$} & \makecell{$J_3$} & $\Delta$ & $\lambda_1 = \lambda_2$ & \makecell{$\lambda_3$} & \makecell{$\surf$} \\
		\hline
		1 & \makecell[r]{$-0.1204$} & \makecell[r]{$0.4315$} & \makecell[r]{$0$} & \makecell[r]{$1.0186$} & \makecell[r]{$0.4159$} & \makecell[l]{Prolate spheroid (refer Eq.~(\ref{eq:prolateEq}))} \\
		\hline
		2 & \makecell[r]{$-2.6952$} & \makecell[r]{$3.0871$} & \makecell[r]{$0$} & \makecell[r]{$0.9994$} & \makecell[r]{$3.0910$} &  \makecell[l]{Oblate spheroid (refer Eq.~(\ref{eq:oblateEqNum}))}\\
		\hline
		3 & \makecell[r]{$3.0356$} & \makecell[r]{$-3.9310$} & \makecell[r]{$0$} & \makecell[r]{$1.1299$} & \makecell[r]{$-3.0794$} &  \makecell[l]{Hyperboloid of one sheet (refer Eq.~(\ref{eq:hypOneSheetEq}))} \\
		\hline
		4 & \makecell[r]{$-1.9864$} & \makecell[r]{$-3.0858$} & \makecell[r]{$0$} & \makecell[r]{$1.1114$} & \makecell[r]{$-2.4984$} &  \makecell[l]{Hyperboloid of two sheets (refer Eq.~(\ref{eq:hypTwoSheetEq}))}\\
		\hline
		5 & \makecell[r]{$-10.3493$} & \makecell[r]{$0$} & \makecell[r]{$0$} & \makecell[r]{$1.0063$} & \makecell[r]{$0$} &  \makecell[l]{Paraboloid (refer Eq.~(\ref{eq:paraboloid}))} \\
		\hline
		6 & \makecell[r]{$0$} & \makecell[r]{$0$} & \makecell[r]{$0$} & \makecell[r]{$1.0044$} & \makecell[r]{$0$} & \makecell[l]{Cylinder (refer Eq.~(\ref{eq:cylinderEqNum}))}
		\\
		\hline
		7 & \makecell[r]{$0$} & $-0.3879$ & $0$ & $1.0432$ & $-0.3564$ &  \makecell[l]{Right circular cone (refer Eq.~(\ref{eq:coneEqNum}))}\\
		\hline
		8 & \makecell[r]{$-0.8710$} & $1$ & $0$ & $1$ & $1$ &  \makecell[l]{Sphere (refer Eq.~(\ref{eq:sphereEq}))}\\
		\hline
	\end{tabular}
\end{table}
\begin{landscape}
	\begin{table}
		\centering
		\caption{Values of $\bpc,n,e, \bpp,t_j~\text{(real roots)},r_j$, and~\rmin are presented up to four decimal points for all the AQs}
		\label{tb:numresultAQ}
		\begin{tabular}{|l|r|r|r|r|r|r|r|}
			\hline
			\makecell{$\surf$} & \makecell{$\bpc$} & \makecell{$n$} & \makecell{$e$} & \makecell{$\bpp$} & \makecell{$t_j$} & \makecell{$r_j$} & \makecell{\rmin}\\
			\hline
			\makecell[l]{Prolate spheroid} & $[-0.8546,0.2070,-0.8311]^\top$ & $0.8192$ & $0.7692$ & $[0.8220,1.3024]^\top$ &  $\{0.4646, -0.4169\}$ & $\{2.3025, 0.8157\}$ & \makecell[r]{$0.8157$\\ (See Fig.~\ref{fg:proSProidTan})}\\
			\hline
			\makecell[l]{Oblate spheroid} & $[0.2475, -0.5002,-0.3252]^\top$ & $0.9347$ & $0.8226$ & $[1.1823,1.4778]^\top$ & $\{-0.6018,0.5180\}$ & $\{2.7229, 1.1300\}$& \makecell[r]{$1.1300$ \\ (See Fig.~\ref{fg:OblSProidTan})}\\
			\hline
			\makecell[l]{Hyperboloid of\\one sheet} & $[0.8546,0.2070,-0.8311]^\top$ & $0.8267$ & $1.1692$ & $[1.3024,0.8220]^\top$ & $\{-1.2247,1.8263\}$ & $\{2.0363,0.8418\}$ & \makecell[r]{$0.8418$ \\(See Fig.~\ref{fg:HypOneTan})}\\
			\hline
			\makecell[l]{Hyperboloid of \\ two sheets} & $[-0.8546,0.2070,-0.8311]^\top$ & $0.5076$ & $1.8022$ & $[0.8220,-1.3024]^\top$ & $\{1.7968,-2.8212\}$ & $\{1.19198,0.4933\}$ & \makecell[r]{$0.4933$\\(See Fig.~\ref{fg:HypTwoTan})}\\
			\hline
			\makecell[l]{Paraboloid} & $[0.4950,0.2826,-0.1122]^\top$ & $-$ & $1$ & $[5.7589,-0.2196]^\top$ & $5.0642$ & $3.1161$ & \makecell[r]{$3.1161$\\(See Fig.~\ref{fg:parabolaShortDist})}\\
			\hline
			\makecell[l]{Cylinder} & $[1.0000, -13.6017, -15.6763]^\top$ & $-$ & $\infty$ & $[6.6747,	20.6088]^\top$ & $-$ & $4.2691$ &  \makecell[r]{$4.2691$\\(See Fig.~\ref{fg:proximitPtCylinder})}\\
			\hline
			\makecell[l]{Cone} & $[-0.8546,0.2070,-0.8311]^\top$& $-$ & $\infty$& $[0.8220,-1.3024]^\top$& $-$ & $\{0.0524,1.3669\}$ & \makecell[r]{$0.0524$ \\(See Fig.~\ref{fg:shortCone})}\\
			\hline
			\makecell[l]{Sphere} & $[0.2475,-0.5002,-0.3252]^\top$ & $-$ & $0$ & $[1.6754,0.8801]^\top$ &$-$& $\{0.9592,2.8258\}$ & \makecell[r]{$0.9592$\\(See Fig.~\ref{fg:sphrShortDist})}\\
			\hline
		\end{tabular}
	\end{table}
\end{landscape}
% % % % % % % % % % % % % % % % % % % % % % % % % % % % % % % % % % % %

For a given equation for the AQs, which are listed below, the classification of the surface is performed based on the values\footnote{Here, the absolute values which are less than~$10^{-6}$ are treated as~$0$.} of~$\det(\bA), J_3,\Delta,a_0,a,~\text{and}, d$, as shown in Table~\ref{tb:classifAQ}. The distances~$r_j$ and minimum among them~\rmin are listed in Table~\ref{tb:numresultAQ}.
\begin{enumerate}
	\item As per Table~\ref{tb:classifAQ}, Eq.~(\ref{eq:prolateEq}) is classified to be a prolate spheroid.
	\begin{align}
	S(\bx):=&x^2 + 
	0.6356 y^2 + 0.8175 z^2 + 0.1688 x y - 0.5550 y z + 0.1223 x z + 1.7758 x \nonumber\\
	&- 0.5803 y   + 1.5783 z    + 1.1956 = 0.
	\label{eq:prolateEq}
	\end{align}
	Using the coefficients of~$\surf$ and the centre~$\bpc$ mentioned in Table~\ref{tb:numresultAQ}, the value of~$\gamma$ is obtained as~$-0.2791$. The equation of the ellipse whose major axis aligned along the~$\bu_2$-axis, as shown in Fig.~\ref{fg:ptOutVertEllipse}, is as follows:
	\begin{align}
	1.0186 u_1^2 + 0.4159 u_2^2 -0.2791  = 0.
	\label{eq:ellipseVertEq}
	\end{align} 
	Using Eq.~(\ref{eq:ellipseVertEq}), the corresponding quartic equation in~$t$ is obtained as:
	\begin{equation}
	t^4 + 2.6048 t^3 + 1.7373 t^2 - 0.6119 t -0.3985 = 0.
	\label{eq:vertEllipseQuartic}
	\end{equation}
	Two real roots to Eq.~(\ref{eq:vertEllipseQuartic}) and the corresponding distances in~$\Re^2$ are listed in Table~\ref{tb:numresultAQ} and plotted in Fig.~\ref{fg:ptOutVertEllipse}.
	The radius of the sphere centred at $\bp_0$ of radius~\rmin, mentioned in the first row of Table~\ref{tb:numresultAQ}, is tangent to the prolate spheroid, as shown in Fig.~\ref{fg:proSProidTan}.
	\item From Table~\ref{tb:classifAQ}, the following equation is an oblate spheroid. 
	\begin{align}
	S(\bx):=&x^2 + 
	1.1353 y^2 + 2.9544 z^2 + 0.0184 x y + 1.0310 y z + 0.0698 x z - 0.4631 x \nonumber \\ & +  1.4665 y + 2.4198 z -0.0556 = 0.
	\label{eq:oblateEqNum}
	\end{align} 
	Equation~(\ref{eq:horizontalEllipse}) is used to represent the ellipse for which the major axis aligned with the~$\bu_2$-axis, as shown in Fig.~\ref{fg:ptOutHorinEllipse}, and the equation of the ellipse with the numerical coefficients is as follows:
	\begin{equation}
	3.0910 u_1^2 +  0.9994u_2^2 -0.8731 = 0.
	\end{equation}   
	 Upon substituting~$e,n$, and~$\bpp$ from Table~\ref{tb:numresultAQ}, the quartic equation in Eq.~(\ref{eq:quarticTellipHyp}) becomes:
	\begin{equation}
	t^4- 2.9555 t^3 + 2.2357 t^2 + 1.1822 t -0.8735 =0.
	\label{eq:HorinEllipseQuartic}
	\end{equation}  
	The solutions to Eq.~(\ref{eq:HorinEllipseQuartic})	and the corresponding values of the distances are tabulated in Table~\ref{tb:numresultAQ}. These can be seen in Fig.~\ref{fg:ptOutHorinEllipse}.
	The radius of the sphere centred at $\bp_0$ of radius~\rmin, mentioned in the second row of Table~\ref{tb:numresultAQ}, is tangent to the oblate spheroid, as shown in Fig.~\ref{fg:OblSProidTan}.
	\item The following equation is associated with a hyperboloid of one sheet.
	\begin{align}
	S(\bx):=&x^2 - 
	1.5451 y^2 - 0.2746 z^2 + 1.1787 x y - 3.8765 y z + 0.8541 x z + 2.1749 x \nonumber \\
	&- 1.5747 y  + 1.0761 z    + 0.7673 =0.
	\label{eq:hypOneSheetEq}
	\end{align}
	To represent the hyperbola for which the major axis is along the~$\bu_2$-axis, as shown in Fig.~\ref{fg:ptOutParabU1int}, Eq.~(\ref{eq:stdFrmHypu2int}) is used. Therefore, the equation of the hyperbola is as follows:
	\begin{equation}
	1.1299 u_2^2 - 3.0794 u_1^2 - 0.7722 = 0. 
	\end{equation} 
	Upon substituting~$e,n$, and~$\bpp$ from Table~\ref{tb:numresultAQ}, the quartic equation in Eq.~(\ref{eq:quarticTellipHyp}) becomes:
	\begin{equation}
	t^4 - 1.6440 t^3 - 1.2233 t^2 + 2.0991 t -0.8627 = 0.
	\label{eq:HypOneQuartic}
	\end{equation}
	Two real solutions to Eq.~(\ref{eq:HypOneQuartic}) and the corresponding distances are presented in Table~$\ref{tb:numresultAQ}$ and depicted in Fig.~\ref{fg:ptOutParabU1int}. The radius of the sphere centred at $\bp_0$ of radius~\rmin, mentioned in the third row of Table~\ref{tb:numresultAQ}, is tangent to the hyperboloid of one sheet, as shown in Fig.~\ref{fg:HypOneTan}.
	\item From Table~\ref{tb:classifAQ}, the following equation represents a hyperboloid of two sheets.
	\begin{align}
	S(\bx):=&x^2 - 1.1826 y^2 - 0.0930 z^2 + 1.0109 x y - 3.3244 y z + 0.7325 x z + 2.1086 x \nonumber \\ 
	&- 1.4094 y  + 1.1596 z	 + 2.1724 = 0. 
	\label{eq:hypTwoSheetEq}  
	\end{align} 
	The equation of the hyperbola for which the major axis aligned along the~$\bu_2$-axis, as shown in Fig.~\ref{fg:ptInParabU2int}, is as follows:
	\begin{equation}
	2.4984 u_2^2 - 1.1114 u_1^2 - 0.6437 = 0.
	\end{equation}
	Upon substituting~$e,n$, and~$\bpp$ from Table~\ref{tb:numresultAQ}, the quartic equation for the proximity in $\Re^2$ shown in Eq.~(\ref{eq:quarticTellipHyp}) becomes:
	\begin{equation}
	t^4 + 2.6048 t^3 - 2.5407 t^2 - 7.0802 t - 4.6106 = 0.
	\label{eq:HypTwoQuartic}
	\end{equation}
	The real solutions to Eq.~(\ref{eq:HypTwoQuartic}) and the corresponding distances are presented in Table~$\ref{tb:numresultAQ}$ and shown in Fig.~\ref{fg:ptInParabU2int}. The radius of the sphere centred at $\bp_0$ of radius~\rmin, mentioned in the third row of Table~\ref{tb:numresultAQ}, is tangent to the hyperboloid of one sheet. The radius of the sphere centred at $\bp_0$ of radius~\rmin, mentioned in the fourth row of Table~\ref{tb:numresultAQ}, is tangent to the hyperboloid of two sheets, as shown in Fig.~\ref{fg:HypTwoTan}.
	\item From Table~\ref{tb:classifAQ}, Eq.~(\ref{eq:paraboloid}) represents paraboloid.
	\begin{align}
	S(\bx):=&x^2 + 0.9884 y^2 +  0.02411 z^2 + 0.0211 x y + 0.2648 y z - 0.1568 x z - 1.5177 x \nonumber \\
	&+ 0.3121 y   - 6.3089 z  -0.3723  = 0. 
	\label{eq:paraboloid}
	\end{align}
	From Eq.~(\ref{eq:parabola}), by using the coefficients of~$\surf$, the vertex~$\bpc$, and~the axis of the symmetry, i.e.,~$\bu_2 = [0.0789, -0.1332, 0.9880]^\top$, the equation of the parabola, as shown in Fig.~\ref{fg:shortDistParabola}, is obtained as:
	\begin{equation}
	1.0063 u_1^2 - 6.3941 u_2 = 0.
	\end{equation}
	From Eq.~(\ref{eq:stdFrmParab}), the $u_2$-coordinate of the focus,~$\gamma$, is obtained as~$1.5886$. Substituting~$\gamma$, and~$\bpp$ in Eq.~(\ref{eq:cubicPolyT}), the cubic equation with numerical coefficients of~$t$ is obtained as:
	\begin{equation}
	t^3 - 2.7379 t^2 - 1.3474 t -  52.8378 =0. 
	\label{eq:parbolaEqNum}
	\end{equation}
	There is only one real solution to Eq.~(\ref{eq:parbolaEqNum}). The corresponding distance is presented in Table~\ref{tb:numresultAQ} and shown in Fig.~\ref{fg:shortDistParabola}. The radius of the sphere centred at $\bp_0$ of radius~\rmin, mentioned in the fifth row of Table~\ref{tb:numresultAQ}, is tangent to the paraboloid, as shown in Fig.~\ref{fg:parabolaShortDist}. 
	\item According to Table~\ref{tb:classifAQ}, Eq.~(\ref{eq:cylinderEqNum}) represents a cylinder.   
	\begin{align}
		S(\bx) := &x^2 + 
		0.5766 y^2 + 
		0.4321 z^2 + 0.0864 x y - 0.9895 y z + 0.0999 x z + 0.7423 x \nonumber\\ 
		&+  0.0880 y - 0.0113 z -5.6730 = 0.
		\label{eq:cylinderEqNum}
	\end{align}
	The corresponding equation of the pair of non-coincident vertical parallel lines is obtained as follows:
	\begin{equation}
		 1.0044 u_1^2 -5.8121 = 0.
	\end{equation}
	The two perpendicular distances from the~$\bpp$ to the parallel lines are mentioned in Table~\ref{tb:numresultAQ}. These two distances are depicted in Fig.~\ref{fg:proximitPtparallelLines}. The radius of the sphere centred at $\bp_0$ of radius~\rmin, mentioned in the sixth row of Table~\ref{tb:numresultAQ}, is tangent to the cylinder, as shown in Fig.~\ref{fg:proximitPtCylinder}.
	\item As per Table~\ref{tb:classifAQ}, Eq.~(\ref{eq:coneEqNum}) represents a right circular cone.
	\begin{align}
	S(\bx):=&x^2 + 
	0.1537 y^2 + 0.5762 z^2 + 0.3920 x y - 1.2890 y z + 0.2840 x z + 1.8640 x \nonumber \\
	&- 0.8000 y + 1.4673 z + 1.4891 = 0.  
	\label{eq:coneEqNum}  
	\end{align}
	 The equation of the corresponding pair of intersecting lines in~$\Re^2$, as shown in Fig.~\ref{fg:shortpairIntLines}, is defined as follows:
	\begin{equation}
	1.0432 u_1^2 - 0.3564 u_2^2 = 0.
	\end{equation}
	The two perpendicular distances from~$\bpp$ are mentioned in Table~\ref{tb:numresultAQ} and shown in Fig.~\ref{fg:shortpairIntLines}. The radius of the sphere centred at $\bp_0$ of radius~\rmin, mentioned in the seventh row of Table~\ref{tb:numresultAQ}, is tangent to the right circular cone, as shown in Fig.~\ref{fg:shortCone}. 
	\item The equation of sphere is defined as:
	\begin{equation}
	S(\bx):=x^2 + y^2 + z^2- 0.4950 x + 1.0004 y  + 0.6503 z -0.4538  =0 	
	\label{eq:sphereEq}
	\end{equation}
	The equation of the circle obtained from the intersection of the sphere and the plane is as follows:
	\begin{equation}
	u_1^2 + u_2^2 -0.8710 = 0.
	\end{equation}
	The two distances,~$r_1,r_2$ are mentioned in Table~\ref{tb:numresultAQ} and presented in Figs.~\ref{fg:circleTanCcircle}. The radius of the sphere centred at $\bp_0$ of radius~\rmin, mentioned in the eighth row of Table~\ref{tb:numresultAQ}, is tangent to the sphere, as shown in Fig.~\ref{fg:sphrShortDist}.
\end{enumerate}
% % % % % % % % % % % % % % % % % % % % % % % % % % % % % % % % % % % % % % %

The above results are compiled into a table along with the execution time in each case. The time mentioned in Table~\ref{tb:execTimeTable} is the time required to identify a surface as an AQ and compute the shortest proximity of~$\bp_0$ to AQ. The reported time is averaged over 1 billion cycles of execution.  
\begin{table}[!h]
	\caption{Amount of time required to identify an AQ and compute the proximity}
	\label{tb:execTimeTable}
	\centering
	\begin{tabular}{|c|c|}
		\hline
		$\surf$ &  Execution time (ns)\\
		\hline
		\makecell[l]{Sphere}  & $0$ \\
		\hline
		\makecell[l]{Cone}  & $8$\\
		\hline
		\makecell[l]{Cylinder}  & $21$\\
		\hline
		\makecell[l]{Paraboloid}  & $77$ \\
		\hline
		\makecell[l]{Hyperboloid of two sheets} & $79$ \\
		\hline
		\makecell[l]{Prolate spheroid} & $82$ \\
		\hline
		\makecell[l]{Oblate spheroid} & $82$ \\
		\hline
		\makecell[l]{Hyperboloid of one sheet} & $88$ \\
		\hline
			
	\end{tabular}
\end{table}
\section{Discussions}
\label{sc:discussion}
The key points in this paper are noted in the following.

The classifications of the general quadric into an AQ is shown in the flowchart, as shown in Fig.~\ref{fg:flowCharSurface}. Even though there is literature that mentions the classification of a quadric, no article has been found that discusses characterising a quadric as AQ. Furthermore, instead of using {\em rank} of~$\bB$, and~$\bA$ for the classification, a different set of parameters such as~$\det(\bA),$and the invariant of~\bB, i.e.,~$J_3$, are used which is advantageous in terms of computational efficiency.
 
 The proximity of a given point to an ellipsoid and hyperboloids in~$\Re^3$ are well studied either analytically or numerically. However, the proximity of a point to the remaining AQs are not found to be in literature. Hence, an extensive studies are performed to compute the proximity to a given point to any AQ. 
	
The problem of computing the proximity of a given point to an AQ in~$\Re^3$ is reduced to the same in~$\Re^2$ due to two reasons. First, the problem becomes more geometrically intuitive, i.e., the proximity of a given point and a conic in~$\Re^2$. Hence, the geometrical properties of the conics such as sub-normal, length of the semi-major axis and eccentricity are used in the derivation of the univariate in the case of parabola, ellipse, and hyperbola. Similarly, for the case of pair of intersecting lines and pair of non-coincident vertical parallel lines, the proximity is computed based on the projection of the point~$\bpp$ onto the lines. In addition to that, solving the problem in~$\Re^2$ is computationally less expensive than solving it in~$\Re^3$, which involves solving three quadratic in equations in~$\bx$, due to the size\footnote{Here, the ``size" means the amount of storage required to store a symbolic expression in the internal format of a computer algebra software, namely,~\texttt{Mathematica}~\cite{Mathematica13}.} of the univariate. For example, the maximum size of the univariate for the latter case is 1.5 MBs~\cite{Aditya2023} whereas the size of the univariate in the former case is in 1.6 KBs. 

The problem of proximity in the case of parabola, ellipse and hyperbola is further classified based on the location of the point~$\bpp$. The classification described in Sections~\ref{sc:pointToParabola} and~\ref{sc:pointToEllipHyp} are summarised in Table~\ref{tb:classfConics}. Such classification is presented for the first time through this paper.
	\begin{table}[t]
		\centering
		\caption{The classification of the problem of proximity between~$\bpp$ and parabola, ellipse, and hyperbola}
		\label{tb:classfConics}
		\begin{tabular}{|c|c|c|}
			\hline
			Conic & Cases & Sub-cases \\
			\hline
			\makecell[l]{\multirow{6}{*}{Parabola}} & \makecell[l]{\multirow{2}{*}{(a)~$\bpp$ lies on the axis~$\bu_2$}} & \makecell[l]{(a.1)~$\bpp$ lies outside the conics (see Fig.~\ref{fg:ptOnU2OutsideParab})}\\
			\cline{3-3}
			& & \makecell[l]{(a.1)~$\bpp$ lies inside the conics (see Fig.~\ref{fg:ptOnU2InsideParab})}\\
			\cline{2-3}
			& \makecell[l]{\multirow{4}{*}{(b)~$\bpp$ does not lie on the axis~$\bu_2$}} & \makecell[l]{(b.1)~$b_1<0 \land \delta <0$: three distinct real roots \\(see Fig.~\ref{fg:threePointsParabola})}\\
			\cline{3-3}
			& & \makecell[l]{(b.2)~$b_1<0 \land \delta =0$: one simple and pair of \\double roots (see Fig.~\ref{fg:threePointParabolaDel0})} \\
			\cline{3-3}
			& & \makecell[l]{(b.2)~$b_1<0 \land \delta>0$: One real root (see Fig.~\ref{fg:onePointParabolaDel0})}\\
			\cline{3-3}
			& & \makecell[l]{(b.2)~$b_1=0 \land \delta>0$: One real root (see Fig.~\ref{fg:onePointParabolaDel01})}\\
			\cline{1-3}
			\makecell[l]{\multirow{7}{*}{\makecell[l]{Ellipse and\\ hyperbola}}} & \makecell[l]{\multirow{2}{*}{(a)~$\bpp$ lies on the major axis}}& \makecell[l]{(a.1)~$\bpp$ lies inside the conics (see Fig.~\ref{fg:ptInsideEllipOnU2axis})}\\
			\cline{3-3}
			& & \makecell[l]{(a.2)~$\bpp$ lies outside the conics (see Fig.~\ref{fg:ptOutsideEllipOnU2axis})}\\
			\cline{2-3}
			& \makecell[l]{\multirow{2}{*}{(b)~$\bpp$ lies on the minor axis}} &\makecell[l]{(b.1)~$\bpp$ lies either outside or inside the ellipse \\(see Fig.~\ref{fg:ptOnXaxisOutEllip})}\\
			\cline{3-3}
			& & \makecell[l]{(b.1)~$\bpp$ lies either outside the hyperbola \\(see Fig.~\ref{fg:ptOnXaxisOutHyp})}\\
			\cline{2-3}
			& \makecell[l]{\multirow{3}{*}{(c)~$\bpp$ does not lie on the axes}} & \makecell[l]{(c.1)~$\delta_1>0:$ four distinct real roots (see Fig.~\ref{fg:ellipse4Pt})}\\
			\cline{3-3}
			& & \makecell[l]{(c.2)~$\delta_1=0:$ two distinct and two \\repeated real roots (see Fig.~\ref{fg:twoReaptTwoDistReRoots})}\\
			\cline{3-3}
			& & \makecell[l]{(c.3)~$\delta_1<0:$ two distinct real roots (see~\ref{fg:twoDistRootsQuartEq})}\\
			\cline{1-3}
		\end{tabular}
	\end{table}
	
The computational complexity for computing the proximity for two cases, i.e., (i) of a point to a right circular cone~($C_1$) and (ii) of a point to a sphere~($S_1$), are compared with a commercial library, namely,~\verb|C++ Bullet|~\cite{Bullet2015}. The library is capable of computing the proximity of~$C_1$ to a sphere as well as between two spheres with suitable modifications. In both the cases, the sphere centred around~$\bp_0$ should not intersect with the~$C_1$ and~$S_1$. The functions, namely,~\verb|getClosestPoints()| and~\verb|normDist()| are used to compute the normal distances from~$\bp_0$ to~$C_1$, and from~$\bp_0$ to~$S_1$. It is found that the~\verb|Bullet| library~$19$ and $10^6$ times slower for the cases (i) and (ii), respectively, than the proposed method. 

\section{Conclusions}
\label{sc:conclusion}
A new and effective way of classifying a general quadric is introduced by using the invariants of the coefficient matrices, instead of computing their ranks. The novel approach of solving the proximity problem of a point to an AQ by reducing from~$\Re^3$ to~$\Re^2$ and categorising the problem in~$\Re^2$ into different sub-cases for parabola and ellipse/hyperbola depending on the location of the point~$\bpp$. The geometrical properties of the conics are used to derive the univariate for which the size of the symbolic expression is lesser than the size of the univariate if the problem is solved in~$\Re^3$. 
The numerical examples are set-up for all the AQs to validate the geometrical method and implemented in~\verb|C| to show the computational efficiencies. It is observed that the classifying a quadric into an AQ and computing the proximity do not take more than 90 ns on a PC. The~\verb|Bullet| library with suitable modifications is used for comparing the execution time to compute the proximity of a point to the right circular cone, and of a point to a sphere in~$\Re^3$ and it is found that the proposed geometrical method is faster than the commercial library. Hence, such method is highly efficient for the applications like design of a robot, collision avoidance, etc., which would be a future scope of this work.
\begin{appendices}
	\section{Computation of $\mathcal{N}(B_1)$}
	\label{sc:nullspace}
	The nullspace of~$\bB_1$,~$\mathcal{N}(\bB_1)$, mentioned in Section~\ref{sc:intPlane}, may be computed as follows:
	\begin{itemize}
		\item The row vector, whose first element is not equal to zero and the $L_2$-norm, is highest among all the three rows is identified and considered to be the first row of~$\bB_1$, as mentioned below:
		\begin{equation}
		\bB_1 = \begin{bmatrix}
		\zeta_{11} & \zeta_{12} & \zeta_{13} \\
		\zeta_{21} & \zeta_{22} & \zeta_{23} \\
		\zeta_{31} & \zeta_{32} & \zeta_{33}
		\end{bmatrix}, \quad \text{where}~\zeta_{11}\neq0,~\text{and}~\norm{[\zeta_{11}, \zeta_{12}, \zeta_{13}]^\top}\neq 0.
		\end{equation}
		\item The second and third row of ~$\bB_1$ are modified as follows:
		\begin{align}
		\bw_2 &=[0, \zeta_{11}\zeta_{22}-\zeta_{12}\zeta_{21}, \zeta_{11}\zeta_{23}-\zeta_{13}\zeta_{21}]^\top,\nonumber\\
		\bw_3&=[0,\zeta_{11}\zeta_{32}-\zeta_{12}\zeta_{31},\zeta_{11}\zeta_{33}-\zeta_{31}\zeta_{13}]^\top.
		\end{align}
		As any of the two rows are linearly dependent, either~$\bw_2$ or~$\bw_3$ vanishes, say,~$\bw_3=\boldsymbol{0}$, in this case.
		\item The vector in~$\nullsp{(\bB_1)}$ is orthogonal to the vectors in the rowspace of~$\bB_1$. Hence, the nullspace of~$\bB_1$ is defined as
		\begin{equation}
		\nullsp{(\bB_1)} :=\bw_1 \times \bw_2 = \bv_3,\quad \text{where}~\bw_1=[\zeta_{11},\zeta_{12},\zeta_{13}]^\top.
		\end{equation} 
		\item If~$\bw_2=\bw_3=\boldsymbol{0}$, the nullspace of~$\bB_1$ is computed as follows:
		\begin{equation}
		\nullsp{(\bB_1)} := \left[-\frac{\zeta_{12}+\zeta_{13}}{\zeta_{11}},1,1\right]^\top=\bv_3.
		\end{equation}
	\end{itemize}

\section{Real roots of a cubic equation}
\label{ap:cubeRoot}
This section provides the expressions for the real roots~\cite{Holmes2002} of Eq.~(\ref{eq:depCubicParabola}) mentioned in Section~\ref{sc:pointToParabola}. Based on the values of $\delta$, and $b_0$, the expressions of the real roots required for this paper are mentioned below.

\begin{enumerate}
	\item \textbf{If $\delta < 0$ and $b_0 <0$}: The solutions of Eq.~(\ref{eq:depCubicParabola}) are real and distinct which are as follows:
	\begin{equation}
	\rho_i = 2\sqrt{-\frac{b_0}{3}}\cos\left(\frac{1}{3}\arccos\left(\frac{3b_1}{2b_0}\sqrt{\frac{-3}{b_0}}-\frac{2\pi j}{3}\right)\right),\quad j=0,\dots,2.
	\label{eq:threeDistRootCube}
	\end{equation}
	\item \textbf{If $\delta > 0$ and $b_0 <0$}: There is one real solution and two complex solutions to Eq.~(\ref{eq:depCubicParabola}) and the expression for the root is as follows:
	\begin{equation}
	\rho_1 = -2\frac{|b_1|}{b_1}\sqrt{-\frac{b_0}{2}}\cosh\left(\frac{1}{3}\text{arcosh}\left(\frac{-3|b_1|}{2b_0}\sqrt{\frac{-3}{b_0}}\right)\right).
	\label{eq:oneRealRootCube}
	\end{equation}
	\item \textbf{If $\delta = 0$ and $b_0 \neq 0$}: In this case, there are two repeated and one distinct real root. The expressions of these roots are:
	\begin{equation}
	\rho_1 = \frac{3 b_1}{b_0},~\text{and}~\rho_2 = \rho_3 = -\frac{3b_1}{2b_0}.
	\label{eq:threeRealRootCube}
	\end{equation}
	\item \textbf{If $\delta > 0$ and $b_0 = 0$}: There exists only one real root, which is defined as follows:
	\begin{equation}
	\rho_1 = \sqrt[3]{-b_1}.
	\label{eq:oneRealRootCubic}
	\end{equation} 
\end{enumerate}

\section{Roots of a quartic equation}
\label{sc:rootquartic}
To determine the solutions of Eq.~(\ref{eq:quarticTellipHyp}), mentioned in Eq.~\ref{sc:pointToEllipHyp}, the quartic equation should be in a factorable form. The equation which makes the quartic equation factorable is called \textit{resolvent cubic}~\cite{Stegun1972} (see, pp.~17-18) and the equation is as follows:
\begin{equation}
\vartheta^3 - (\uptwo^2 - (e_2 +e_1 \upone^2))\vartheta^2  + 4 e_2 e_1 \upone^2 \uptwo^2 = 0.
\label{eq:resolvantCubicEq}
\end{equation}
One real root of Eq.~(\ref{eq:resolvantCubicEq}), say, $\vartheta_1$, is used to determine the four roots~\cite{Stegun1972} (see, pp.~17-18) of Eq.~(\ref{eq:quarticTellipHyp}) as follows:
\begin{align}
&t_1 = \frac{1}{2}(\uptwo + \tau + \omega),\nonumber\\
&t_2 = \frac{1}{2}(\uptwo + \tau - \omega),\nonumber\\
&t_3 = \frac{1}{2}(\uptwo - \tau + \xi),\nonumber\\
&t_4 = \frac{1}{2}(\uptwo - \tau - \xi), \quad \text{where} \label{eq:solnQuarticEq} \\
&\tau= \sqrt{e_2 + \upone^2 e_1 + \vartheta_1},\nonumber\\
&\varpi_1 = 2 e_2 - \tau^2 + 2 \upone^2 e_1 + \uptwo^2,  \nonumber\\
&\varpi_2 = \frac{2 \uptwo(e_1 \upone^2 - e_2)}{\tau},\nonumber\\
&\varpi_3 = \varpi_1 + \tau^2, \nonumber \\
&\varpi_4 = 2\sqrt{\vartheta_1^2+4 e_2 \uptwo^2},\nonumber\\
&\omega = \begin{cases}
\sqrt{\varpi_1 + \varpi_2}, \quad \text{for}~\tau\neq 0,\\
\sqrt{\varpi_3 + \varpi_4},\quad \text{for}~\tau = 0,
\end{cases}\nonumber\\
&\xi = \begin{cases}
\sqrt{\varpi_1 - \varpi_2}, \quad \text{for}~\tau\neq 0,\\
\sqrt{\varpi_3 - \varpi_4},\quad \text{for}~\tau = 0.
\end{cases}\nonumber
\end{align}

\end{appendices}

\bibliographystyle{ieeetr} 
\bibliography{Bibek}
\end{document}